\documentclass[runningheads]{llncs}

\usepackage[mobile]{eccv}

\usepackage{eccvabbrv}

\usepackage{graphicx}
\usepackage{booktabs}

\usepackage[accsupp]{axessibility}  

\usepackage{colortbl, pifont}

\definecolor{dt}{HTML}{ADCAD8}
\definecolor{dt2}{HTML}{cddfe7}

\newcommand{\otsmodel}[1]{\cellcolor{dt2}{#1}}

\definecolor{baselinecolor}{gray}{.9}

\definecolor{defaultcolor}{HTML}{E8E2F7}

\newcommand{\scolorbox}[2]{{\setlength{\fboxsep}{2pt}\colorbox{#1}{#2}}}

\usepackage{hyperref}

\usepackage{orcidlink}

\begin{document}

\title{Agglomerative Token Clustering} 

\titlerunning{Agglomerative Token Clustering}

\author{Joakim Bruslund Haurum\inst{1}\orcidlink{0000-0002-0544-0422} \and
Sergio Escalera\inst{2,1}\orcidlink{0000-0003-0617-8873} \and
Graham W.~Taylor\inst{3}\orcidlink{0000-0001-5867-3652} \and Thomas B.~Moeslund\inst{1}\orcidlink{0000-0001-7584-5209}}

\authorrunning{J.~B.~Haurum \etal}

\institute{Visual Analysis and Perception (VAP) Laboratory, Aalborg University \& Pioneer Centre for AI, Denmark \and
Universitat de Barcelona \& Computer Vision Center, Spain\and
University of Guelph \& Vector Institute for AI, Canada\\
\email{\{joha, tbm\}@create.aau.dk, sescalera@ub.edu, gwtaylor@uoguelph.ca}\\
\small{\url{https://vap.aau.dk/atc/}}}

\maketitle

\begin{abstract}
We present Agglomerative Token Clustering (ATC), a novel token merging method that consistently outperforms previous token merging and pruning methods across image classification, image synthesis, and object detection \& segmentation tasks. ATC merges clusters through bottom-up hierarchical clustering, without the introduction of extra learnable parameters. We find that ATC achieves state-of-the-art performance across all tasks, and can even perform on par with prior state-of-the-art when applied \textit{off-the-shelf}, \ie without fine-tuning. ATC is particularly effective when applied with low keep rates, where only a small fraction of tokens are kept and retaining task performance is especially difficult. 
\end{abstract}

\section{Introduction}
Since their introduction in 2020, Vision Transformers (ViTs) \cite{ViT_2021} have been utilized for a wide variety of computer vision tasks such as image classification, synthesis, segmentation and more with great success \cite{ViTDet, ViTAdapter, stableDiffusion}. Unlike Convolutional Neural Networks (CNNs), which require a structured representation throughout the network, ViTs can process variable length input sequences, even allowing for the sequences to be modified throughout the network. This gives ViTs stronger expressive powers than CNNs \cite{kornblithViTSee}, but comes at a computational cost due to the quadratic scaling of the self-attention computation. Therefore, there has been increasing research interest in making ViTs more efficient while retaining their task performance. \textit{Token reduction} has shown to be a promising subfield which directly decreases model complexity by reducing the input sequence through pruning or merging \cite{Haurum_2023_ICCV}, thereby decreasing the computation cost of the self-attention operation. Haurum \etal \cite{Haurum_2023_ICCV} conducted an in-depth study of 13 different methods, and found that the baseline Top-K pruning-based method, and its extension EViT \cite{EViT_2022}, outperformed the vast majority of more complex methods on four image classification tasks. However, token pruning has the disadvantage that it typically requires the backbone to be fine-tuned in order to maintain good performance even at high keep rates, which by design leads to information loss as tokens are removed, and performs poorly on image synthesis tasks \cite{ToMe_2023,ToMeSD}. Motivated by these observations, we focus on merging-based token reduction methods as they show the most versatility.

\begin{figure}[!t]
    \centering
    \includegraphics[width=0.9\linewidth]{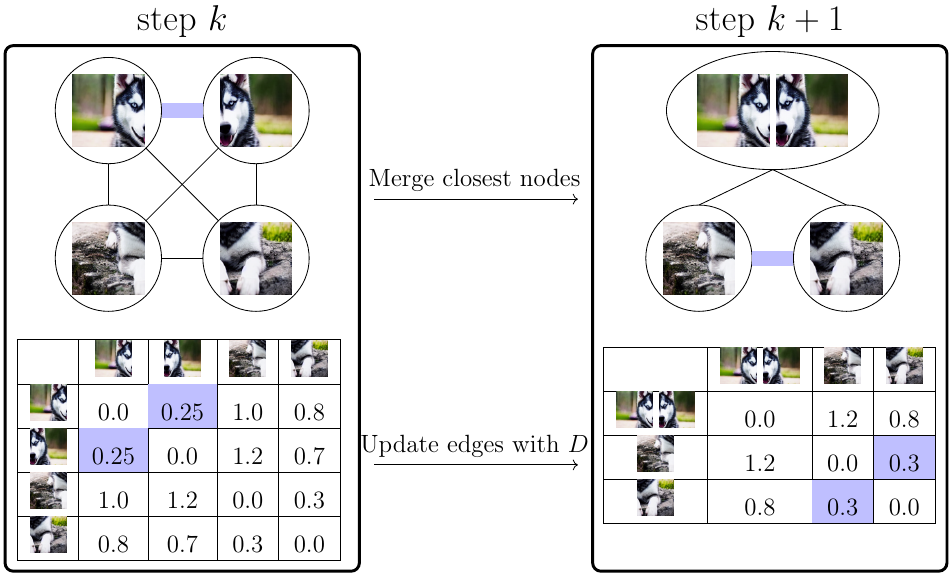}
    \caption{\textbf{Illustration of the Agglomerative Clustering Method.} Prior hard merging-based methodologies have focused on using either partition-based approaches (\eg DPC-KNN \cite{DPCKNN_2022} or K-Medoids \cite{Marin_2023}) or graph-based (\ie ToMe). All of these methods globally cluster the input tokens through the use of cluster centers. In contrast, our \textbf{Agglomerative Token Clustering (ATC)} method builds clusters locally, \ie by iteratively combining the most similar tokens, until the desired amount of tokens remain. A step of this process is shown here, where a graph of nodes (in this case, tokens) are connected with edges based on their similarity. The most similar pair of nodes are combined, and the edges are updated using linkage function $D$, in this case $D^{\text{complete}}$ (Eq.~\ref{eq:complete}).}
    \label{fig:mergeComp}
\end{figure}

We present a novel merging-based token reduction method, Agglomerative Token Clustering (ATC), which outperforms all prior merging-based and pruning-based token reduction methods on both classification tasks and dense computer vision tasks such as image synthesis and object detection \& segmentation, and achieve comparable performance without any fine-tuning, \ie \textit{off-the-shelf}. This is the most diverse set of experiments considered within token reduction, where previous methods have been evaluated just on classification \cite{Haurum_2023_ICCV, ToMe_2023, EViT_2022}, image synthesis \cite{ToMeSD}, or detection and segmentation \cite{SViT}. Through this suite of diverse tasks we demonstrate the efficacy of the proposed ATC method. 

ATC is motivated by the observation that prior merging-based methods such as K-Medoids \cite{Marin_2023}, DPC-KNN \cite{DPCKNN_2022}, and ToMe \cite{ToMe_2023} all perform merging globally, which may lead to redundant clusters. Our hypothesis is that hierarchically merging similar, and thus redundant, observations results in more informative groupings. A natural and robust methodology for including this notion into token reduction is via agglomerative clustering \cite{ClusterBook}, where tokens are iteratively clustered in a bottom-up hierarchical way, see Figure~\ref{fig:mergeComp}. 

Our contributions are the following:
\begin{itemize}
    \item We propose Agglomerative Token Clustering (ATC), a novel parameter-free hierarchical merging-based token reduction method.
    \item Using ATC we achieve state-of-the-art performance on image classification, image synthesis, and object detection \& segmentation tasks, outperforming all other token reduction methods, including both merging-based and pruning-based token reduction methods.
    \item We show ATC can reach comparable performance to the prior fine-tuned state-of-the-art in image classification and object detection \& segmentation when applied off-the-shelf, \ie without any fine-tuning.
\end{itemize}

\section{Related Work}
\noindent\textbf{Efficient Transformers.} As ViTs have become widely adapted by the computer vision community, there have been several attempts at making ViTs more efficient. These attempts range from model pruning \cite{SViTE_2021, UPViT_2021, ViTSlim_2022, AdaVIT_2022, CP-ViT, SparseViT_2023}, quantization \cite{QVIT_2022, FQViT_2022}, structured downsampling \cite{SWIN_2021, PiT_2021, LeViT_2021, LIT_2022}, sparsification of the attention module \cite{Sparsifinier, STViT_2023}, part selection modules \cite{TransFG_2022, MAWS_2021, HAS_2022, RAMSTrans_2021, R2Trans_2022}, and dynamically adjusting the size of input patches \cite{DVT_2021, ReViT_2021, FlexiViT_2023, PSViT_2021, MSViT_2022, SuperViT_2022, CFVIT_2022, NaViT}. Lastly, the field of token reduction has emerged, which is the topic of this paper.

\noindent\textbf{Token Reduction.} The goal of token reduction is to sparsify the sequence of patches, also referred to as tokens, processed by the ViT. This has been achieved through either token pruning or token merging \cite{Haurum_2023_ICCV}. Token pruning focuses on removing tokens either through keeping the tokens with the highest attention from the class (CLS) token \cite{EViT_2022, EVoViT_2022, Haurum_2023_ICCV, PPT, ATS_2022, DiversityEVIT_2022, TPS_2023}, introducing a gating mechanism based on the Gumbel-Softmax trick \cite{GumbelSoftmax, DyViT_2021, SViT, SPViT_2022, TPS_2023}, sampling based approaches \cite{AViT_2022, ATS_2022} modifying the training loop \cite{SaiT_2022, SViTE_2021} or reinforcement learning \cite{IARED2_2021}.

Token merging, on the other hand, combines tokens instead of explicitly pruning them. This can be done either through hard or soft merging of tokens. Hard merging includes techniques such as the partition-based K-Medoids \cite{Haurum_2023_ICCV, Marin_2023} and DPC-KNN \cite{DPCKNN_2022}, as well as the bipartite graph-based approach Token Merging (ToMe) \cite{ToMeSD, ToMe_2023}. Hard merging-based approaches are characterized by having the tokens assigned to clusters in a mutually exclusive manner. In contrast, soft merging techniques let tokens be assigned to multiple clusters resulting in cluster centers being a convex combination of tokens. This combination is either based on similarity between the spatial tokens \cite{Marin_2023}, or the similarity between the spatial tokens and a set of explicit queries which are optimized \cite{SiT_2022, PatchMerger_2022,Sinkhorn_2022,GroupViT_2022}.

The token reduction field has been moving at an immense speed, resulting in little to no comparisons between methods. This was rectified by Haurum \etal \cite{Haurum_2023_ICCV}, who compared 13 different token reduction methods over four image classification datasets. The study provided insights into the token reduction process through extensive experiments, showing that the Top-K and EViT pruning-based methods consistently outperformed all other token reduction methods on the considered classification datasets. However, Bolya and Hoffmann found that merging-based methods outperform pruning-based methods for image synthesis \cite{ToMeSD}, while Bolya \etal showed that the merging-based ToMe \cite{ToMe_2023} can perform well on several classification tasks without any fine-tuning.

Despite its impressive performance, a core component of ToMe is the bipartite matching algorithm, where tokens are split into two exclusive sets between which a bipartite graph is created. This inherently limits which tokens can be merged as there is no within-set comparison and limits the keep rate, $r$, such that it must be $50\%$ or higher. Therefore, we make a single, but important, modification to the ToMe method, replacing the bipartite matching algorithm with the classical agglomerative clustering method \cite{ClusterBook}.

\section{Agglomerative Token Clustering}
Similar to previous token merging methods, the objective of ATC is to merge redundant tokens, while preserving or enhancing the performance of the ViT model. We insert the token merging operation between the self-attention and Multi Layer Perceptron (MLP) modules in a ViT block. This is consistent with prior merging-based methods, such as ToMe \cite{ToMe_2023}. 

 We believe the agglomerative clustering algorithm is a more appropriate choice as it builds the clusters in a bottom-up manner, such that redundant features are clustered early on, while more diverse features are kept unmodified for as long as possible. 
 Prior partition-based (\eg DPC-KNN and K-Medoids) and graph-based (\ie ToMe) merging methods are limited by having to create clusters globally, necessitating the selection of cluster centers which may be redundant. In contrast, ATC creates clusters in a sequential manner, resulting in a local merging approach which leads to the most redundant feature to be clustered at any step in the process. We revisit the ToMe method in Section~\ref{sec:ToMe} and agglomerative clustering in Section~\ref{sec:AggC}.

\subsection{Token Merging}\label{sec:ToMe}
 ToMe was designed for seamless integration into ViTs, allowing for minimal performance loss without necessitating fine-tuning. Its key feature is a fast bipartite merging operation, placed between the self-attention and MLP modules in the ViT block. A bipartite graph is then constructed by setting the edge between nodes in token subsets $A$ and $B$ equal to their similarity, where the $t$ highest valued edges are kept while allowing only a single edge for each node in subset $A$.
Bolya \etal investigated how to construct $A$ and $B$, and found the best performance was achieved by assigning tokens in an alternating manner. This inherently limits which tokens can be merged, which can lead to redundant clusters as spatially co-located patches contain semantically similar information.

\subsection{Agglomerative Clustering}\label{sec:AggC}
Agglomerative Clustering is a classical method for bottom-up hierarchical clustering, where each element is initially its own cluster \cite{ClusterBook}. 
The elements are combined by iteratively comparing the clusters according to some \textit{linkage} function with distance metric $D(\cdot)$, combining the two closest clusters in each iteration. This is repeated until a certain stopping criteria is met, such as the number of desired clusters, leading to a static reduction method, or a minimum distance between clusters, leading to a dynamic reduction method. This is illustrated in Figure~\ref{fig:mergeComp}. In this paper we consider the static reduction scenario. Similar to Bolya \etal , we use the cosine distance as our distance metric $D(\cdot)$ and the keys from the self-attention module as token features. The choice of linkage function can have a large impact on how the elements are clustered. We consider the three most common ones: single (Eq.~\ref{eq:single}), complete (Eq.~\ref{eq:complete}), and average (Eq.~\ref{eq:average}) \cite{NNChainingClust}. 
\begin{equation}\label{eq:single}
    D(I,J)^{\text{single}} = \min_{i\in I,\ j\in J} D(i,j)
\end{equation}
\begin{equation}\label{eq:complete}
    D(I,J)^{\text{complete}} = \max_{i\in I,\ j\in J} D(i,j)
\end{equation}
\begin{equation}\label{eq:average}
    D(I,J)^{\text{average}} = \frac{1}{|I||J|}\sum_{i\in I}\sum_{j\in J}D(i,j)
\end{equation}
where $I$ and $J$ are clusters with elements $i\in I$ and $j\in J$. 

After the stopping criteria has been reached we average the tokens in each cluster to get an updated cluster representation. However, as tokens are merged they represent more than one input patch. In order to advantage tokens that capture a larger spatial extent, we use the weighted average for the cluster representation and proportional attention in the self-attention module as proposed by Bolya \etal.
\section{Experiments}
To evaluate the versatility and applicability of ATC, we conduct assessments across a diverse set of tasks (image classification, image synthesis, and object detection \& segmentation) and datasets. 

For the image classification task we follow the experimental protocol of Haurum \etal \cite{Haurum_2023_ICCV}, evaluating multi-class and multi-label classification performance across four classification datasets using three DeiT models and token reduction performed at three discrete stages. We also follow the MAE experiments of Bolya \etal \cite{ToMe_2023}, evaluating on the ImageNet-1K dataset with token reduction at all stages following a constant and linearly decreasing schedule. For the image synthesis task, we follow the proposed protocol of Bolya \& Hoffman \cite{ToMeSD}, incorporating the token reduction method into the Stable Diffusion image generation model \cite{stableDiffusion}.  Lastly, for the object detection and segmentation task we follow the experimental protocol of Liu \etal \cite{SViT}, evaluating on the COCO-2017 dataset \cite{COCO}. Two versions of the ViT-Adapter model are used with the token reduction method incorporated at three discrete stages.

All experiments are conducted using ATC with the single, complete, and average linkage functions, respectively. The different linkage functions are indicated using a superscript, such as ATC$^{\text{single}}$ for the single linkage function. For the image classification and object detection \& segmentation tasks we report both \textit{off-the-shelf} results, where ATC is inserted into the pre-trained model and evaluated without any fine-tuning, and results after fine-tuning. For the image synthesis task we report off-the-shelf results, similarly to Bolya \& Hoffman \cite{ToMeSD}.\\

\begin{figure}[!t]
\begin{minipage}[t]{0.47\linewidth}
    \centering
    \includegraphics[width=\linewidth]{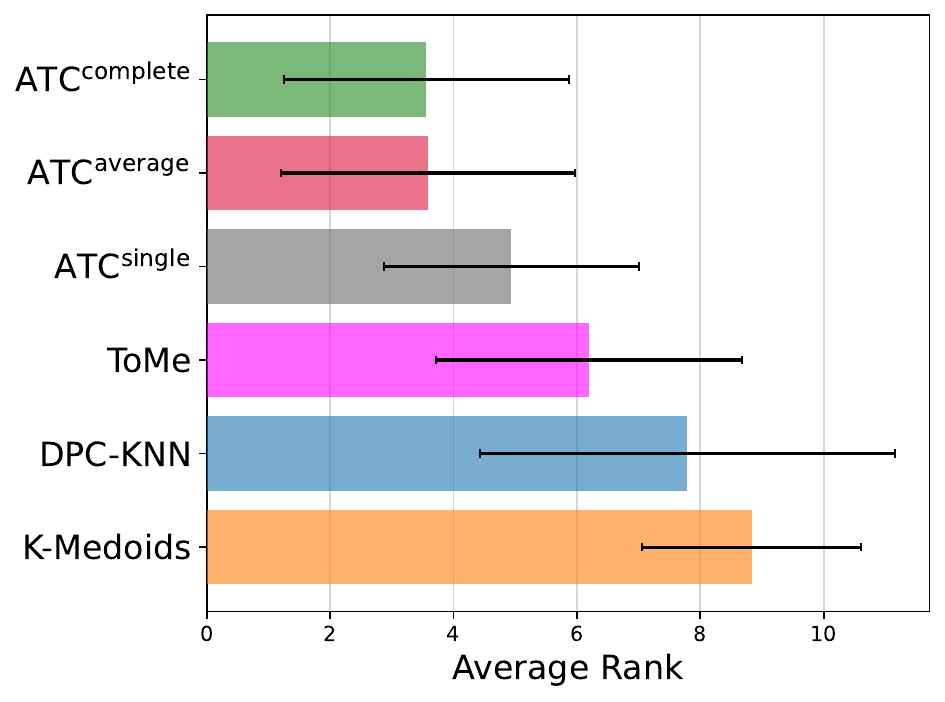}
    \captionof{figure}{\textbf{Average Token Reduction Rank (Lower is better).} We compare our proposed ATC method with the hard-merging based token reduction methods investigated by Haurum \etal \cite{Haurum_2023_ICCV}. We average across four keep rates, three model capacities, and four datasets, and plot with $\pm 1$ standard deviation similar to Haurum \etal. We test three versions of ATC, varying the linkage function, and find that the three variants all outperform the prior merging-based methods.}
    \label{fig:rankedICCV}
\end{minipage}%
\hfill
\begin{minipage}[t]{0.47\linewidth}
    \centering
    \includegraphics[width=\linewidth]{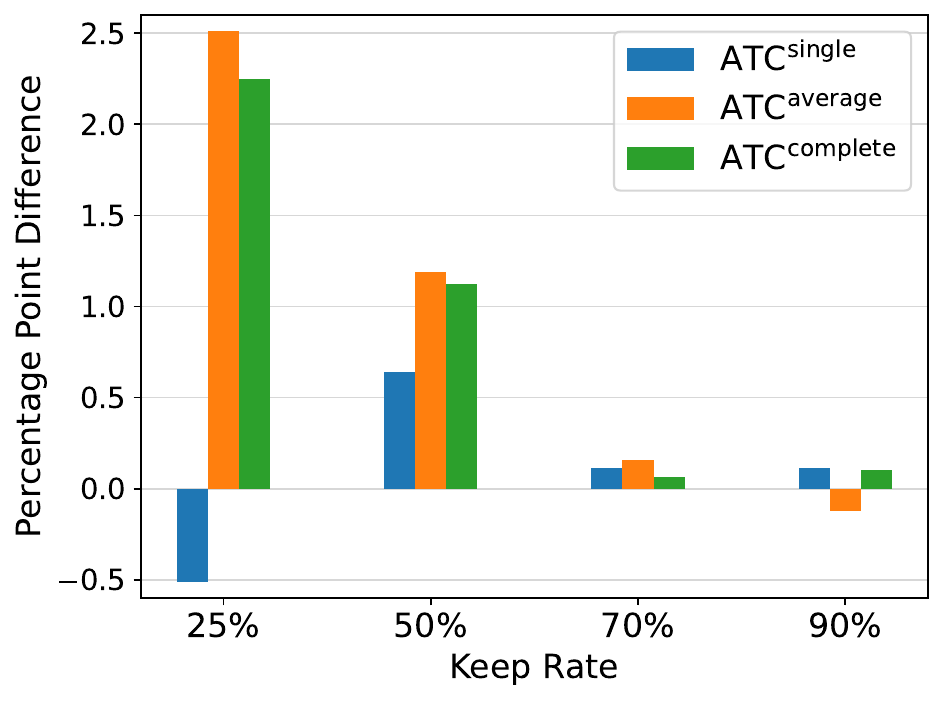}
    \captionof{figure}{\textbf{Percentage Point Difference per Keep Rate.} We compute the average difference between our proposed ATC method and the best prior merging-based methods investigated by Haurum \etal \cite{Haurum_2023_ICCV} for each keep rate, measured in percentage points. We average across the three model capacities and four datasets. We find that for high keep rates $r=\{70, 90\}\%$ ATC is comparable to the prior best merging method, while for $r=\{25,50\}\%$ our proposed ATC method leads to significant performance gains.}
    \label{fig:deltaICCV}
\end{minipage}%
\end{figure}

\subsection{Cross-Dataset Classification Performance}\label{sec:iccvDataset}

Following the experimental protocol proposed by Haurum \etal \cite{Haurum_2023_ICCV}, we evaluate the performance of ATC across four classification datasets covering multi-class and multi-label classification: ImageNet-1K \cite{ImageNet}, NABirds, \cite{NABirds} COCO 2014 \cite{COCO}, and NUS-WIDE \cite{NUS_WIDE} datasets. ImageNet-1K and NABirds are evaluated with the accuracy metric, while COCO and NUS-WIDE are evaluated with the mean Average Precision (mAP) metric. Token reduction is applied at the 4th, 7th, and 10th ViT block, where at each stage only $r\in \{25, 50, 70, 90 \}$\% of the available tokens are kept. The backbone model is a DeiT \cite{DeiT_2021} model trained without distillation, across three model capacities: DeiT-Tiny, DeiT-Small, and DeiT-Base. The models are denoted DeiT-T, DeiT-S, and DeiT-B, respectively. We consider both the off-the-shelf scenario, where ATC is inserted into the pre-trained DeiT models with no further training, as well as the fine-tuning scenario, where the model is fine-tuned for 30 epochs \cite{Haurum_2023_ICCV}. We conduct a hyperparameter sweep following the setup used by Haurum \etal \cite{Haurum_2023_ICCV}. The final hyperparameters can be found in the supplementary materials.\\

\begin{figure}[!t]
    \centering
    \includegraphics[width=\linewidth]{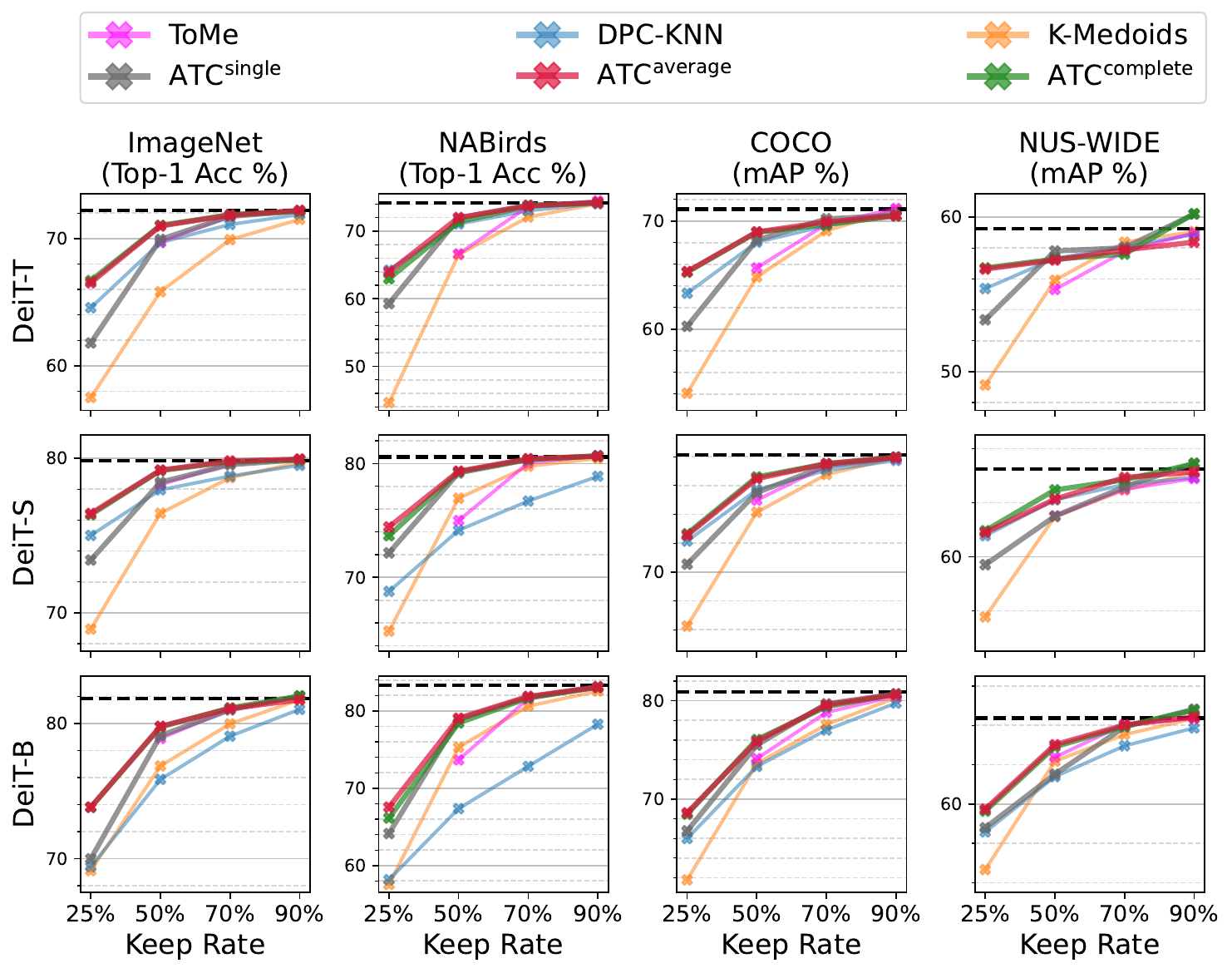}
    \caption{\textbf{Hard Token Merging Method Comparison with the DeiT Backbone.} We compare the hard-merging token reduction methods considered by Haurum \etal \cite{Haurum_2023_ICCV} with the proposed ATC method. All methods have been fine-tuned. Model performance is measured across keep rates, $r$, denoted in percentage of tokens kept at each reduction stage, and with the DeiT-\{Tiny, Small, Base\} models. Comparison with all 13 token reduction methods considered by Haurum \etal can be found in the supplementary materials. ImageNet and NABirds performance is measured with top-1 accuracy, whereas COCO and NUS-WIDE is measured with mAP. The baseline DeiT performance is noted with a dashed black line. Note that ToMe is limited to $r\geq 50\%$, and that ATC$^{\text{average}}$ and ATC$^{\text{complete}}$ often overlap.}
    \label{fig:iccvResults}
\end{figure}

We find that across all three backbone capacities, the fine-tuned ATC methods consistently outperform or match the performance of the previously proposed merging-based methods using all three proposed linkage functions, see Figure~\ref{fig:rankedICCV} and Figure~\ref{fig:iccvResults} for details. We also investigate the average improvement in performance between ATC and the prior best merging-based methods, see Figure~\ref{fig:deltaICCV}. We find that at keep rates of 70\% and 90\%, all three linkage functions achieve comparable results, while at lower keep rates the single linkage function drops in performance. We believe this is due to the \textit{chaining phenomenon} where two distinct clusters are merged due to outliers within the clusters \cite{ClusterBook}. 

\begin{figure}[!t]
\centering
\resizebox{0.49\linewidth}{!}{%
\begin{tabular}{ccc}
  
  \huge{DPC-KNN} & \huge{K-Medoids} & \huge{ATC$^{\text{average}}$} \\
\includegraphics[width=0.33\columnwidth]{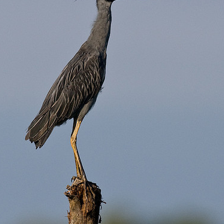}  & \includegraphics[width=0.33\columnwidth]{qualitive_eval/first_example/original.png} & \includegraphics[width=0.33\columnwidth]{qualitive_eval/first_example/original.png}\\
  \includegraphics[width=0.33\columnwidth]{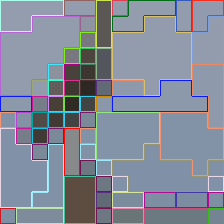} &   \includegraphics[width=0.33\columnwidth]{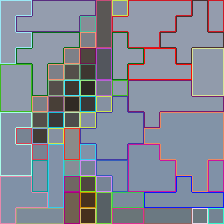} & \includegraphics[width=0.33\columnwidth]{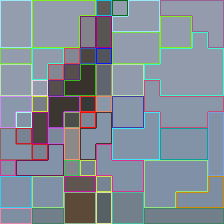} \\
    \includegraphics[width=0.33\columnwidth]{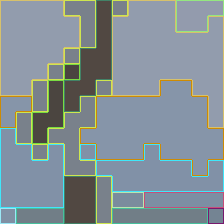} &   \includegraphics[width=0.33\columnwidth]{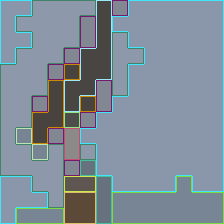} & \includegraphics[width=0.33\columnwidth]{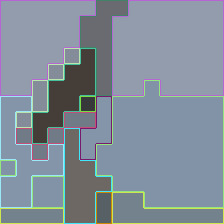} \\
      \includegraphics[width=0.33\columnwidth]{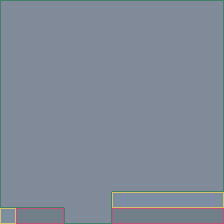} &   \includegraphics[width=0.33\columnwidth]{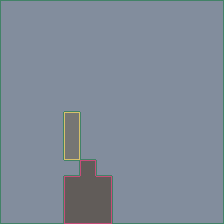} & \includegraphics[width=0.33\columnwidth]{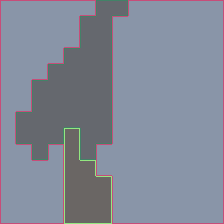} \\
      & \LARGE{\textbf{(a)}} &  \\
\end{tabular}
}\hfill
\resizebox{0.49\linewidth}{!}{%
\begin{tabular}{ccc}
  \huge{DPC-KNN} & \huge{K-Medoids} & \huge{ATC$^{\text{average}}$} \\
\includegraphics[width=0.33\columnwidth]{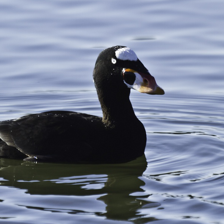}  & \includegraphics[width=0.33\columnwidth]{qualitive_eval/second_example//original.png} & \includegraphics[width=0.33\columnwidth]{qualitive_eval/second_example//original.png}\\
  \includegraphics[width=0.33\columnwidth]{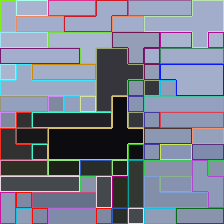} &   \includegraphics[width=0.33\columnwidth]{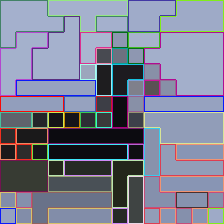} & \includegraphics[width=0.33\columnwidth]{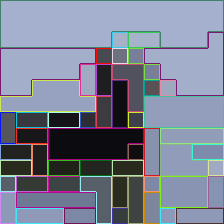} \\
    \includegraphics[width=0.33\columnwidth]{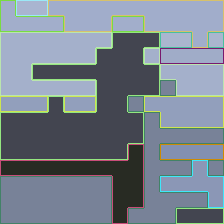} &   \includegraphics[width=0.33\columnwidth]{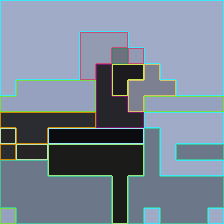} & \includegraphics[width=0.33\columnwidth]{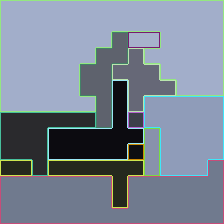} \\
      \includegraphics[width=0.33\columnwidth]{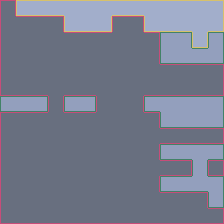} &   \includegraphics[width=0.33\columnwidth]{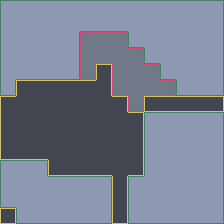} & \includegraphics[width=0.33\columnwidth]{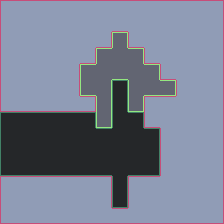} \\
      & \LARGE{\textbf{(b)}} &  \\
\end{tabular}

}
\caption{\textbf{Token Merging Visualization with $r=25$\%.} We visualize the three token merging steps for DPC-KNN \cite{DPCKNN_2022}, K-Medoids \cite{Marin_2023} and our ATC$^{\text{average}}$ on two examples from NABirds \cite{NABirds} with a DeiT-B backbone. The first row is the input image, and each subsequent row is the constructed clusters after the first, second, and third reduction stage. In subfigure \textbf{(a)} we find that there is a major difference in the final clustering of the data, where our ATC method creates separate clusters for the bird, wood pole, and background. In contrast, DPC-KNN and K-Medoids create mostly arbitrary clusters. Similarly in subfigure \textbf{(b)} we see that the DPC-KNN method creates very arbitrary clusters, while K-Medoids and ATC create more meaningful clusters. However, the ATC clusters still better contain the bird in the image, while the K-Medoids clusters have background patches in all clusters.  We find this to be a repeating occurrence and believe this is the reason for the large improvement by ATC on NABirds at $r=25$\%.}
\label{fig:iccvViz}
\end{figure}
However, at keep rates of 25\% and 50\% we find that both the average and complete lead to large performance improvements compared to the prior best merging methods (up to 2.5 percentage points as per Figure~\ref{fig:deltaICCV}). In some cases we even find that at keep rates of 25\% we can improve performance significantly, such as with the DeiT-S and DeiT-B backbones on NABirds, where ATC with the average linkage function results in a 5.7 and 9.6 percentage point increase in accuracy over the prior state-of-the-art, respectively. Through qualitative evaluation, as seen in Figure~\ref{fig:iccvViz}, we can provide intuition for why ATC outperforms prior merging-based methods. We find that DPC-KNN and K-Medoids create arbitrary clusters whereas ATC creates more meaningful clusters even at the third reduction stage with $r=25\%$.

Lastly, we find that applying ATC off-the-shelf on the DeiT backbone leads to good performance at high keep rates, matching or even outperforming the prior best performing merging methods. However, when the keep rate is lowered the performance drops dramatically, which can be rectified through fine-tuning. For full details on the off-the-shelf results, we refer to the supplementary materials.

\subsection{Classification with Self-Supervised Pretraining}
We follow the protocol of Bolya \etal \cite{ToMe_2023} and compare the performance on ImageNet-1K when applying token reduction on pretrained ViT models, specifically the MAE models trained initially with masked image modelling \cite{MAE} and fine-tuned on ImageNet \cite{ImageNet}. We consider the off-the-shelf performance using the publicly available checkpoints, as well as fine-tuning using the original fine-tuning protocol by He \etal \cite{MAE}.
We consider the ViT-Base, ViT-Large, and ViT-Huge models, where token reduction is applied at every block, following the constant and linear decreasing schedules proposed by Bolya \etal:\\
\begin{equation}\label{eq:constantSched}
    t^l = t
\end{equation}
\begin{equation}\label{eq:linearSched}
    t^l = \left\lfloor2t-\frac{2tl}{L-1}\right\rfloor,
\end{equation}
where $L$ is the total number of ViT blocks, $t^l$ is the number of tokens to be removed at ViT block $l=\{0,1,\ldots, L-1\}$, and $t$ is a hyperparameter controlling the aggressiveness of the reduction. Both schedules result in $tL$ tokens being removed in total, with the linear schedule removing more tokens at early layers.

We fine-tune the ViT block models using both schedules in the most aggressive settings considered by Bolya \etal: ViT-B with $t=16$, ViT-L with $t=8$, and ViT-H with $t=7$. Note that we also fine-tune ToMe, leading to a second set of values that are a small improvement compared to the original paper. We also replicate the larger sweep over $t$ values on off-the-shelf ViT models with weights from different training protocols \cite{SWAG, MAE, AugReg}, originally performed by Bolya \etal. These are found in the supplementary materials.
\begin{table}[!tp]
\centering\setlength{\tabcolsep}{2pt}
\caption{\textbf{MAE Pretrained Backbone Comparison.} We compare ToMe and ATC with a self-supervised MAE backbone on ImageNet-1K. We consider three backbones: ViT-Base, ViT-Large, and ViT-Huge. Tokens are removed at each ViT block, following the constant (Eq.~\ref{eq:constantSched}) or linear (Eq.~\ref{eq:linearSched}) token schedules. The best performing method per column is denoted in \textbf{bold}. A \scolorbox{dt2}{blue} background indicates that the token reduction method was applied off-the-shelf on the ViT backbone, whereas all other models have have been fine-tuned. For each ATC method we write the performance improvement over ToMe in parenthesis after the model accuracy.}
\label{tab:mae}

\resizebox{\linewidth}{!}{%
\begin{tabular}{lllccllccll}
\toprule
 & \multicolumn{2}{c}{ViT-B ($t=16$)} &&&  \multicolumn{2}{c}{ViT-L ($t=8$)} &&&  \multicolumn{2}{c}{ViT-H ($t=7$)} \\ \cmidrule(lr){2-3} \cmidrule(lr){6-7} \cmidrule(l){9-11}  
 
Schedule & \multicolumn{1}{c}{Constant} & \multicolumn{1}{c}{Linear} &&& \multicolumn{1}{c}{Constant} & \multicolumn{1}{c}{Linear} &&& \multicolumn{1}{c}{Constant} & \multicolumn{1}{c}{Linear} \\ \midrule

No Reduction &  \multicolumn{2}{c}{83.6}  &&& \multicolumn{2}{c}{85.9} &&& \multicolumn{2}{c}{86.9}\\ \midrule

\otsmodel{ToMe} & \otsmodel{78.5}& \otsmodel{}56.6 &\otsmodel{}&\otsmodel{}& \otsmodel{84.2} & \otsmodel{80.1}&\otsmodel{}&\otsmodel{}& \otsmodel{86.0} & \otsmodel{85.0}\\
\otsmodel{ATC$^{\text{single}}$ (ours)} & \otsmodel{79.7 (+1.2)} & \otsmodel{65.8 (+9.2)} &\otsmodel{}&\otsmodel{}& \otsmodel{84.8 (+0.6)} & \otsmodel{82.5 (+2.4)} &\otsmodel{}&\otsmodel{}  & \otsmodel{86.4 (+0.4)} &\otsmodel{85.6 (+0.6)}\\
\otsmodel{ATC$^{\text{average}}$ (ours)} & \otsmodel{80.1 (+1.6)} & \otsmodel{67.1 (+10.5)} &\otsmodel{}&\otsmodel{}& \otsmodel{84.8 (+0.6)} & \otsmodel{82.6 (+2.5)}&\otsmodel{}&\otsmodel{}& \otsmodel{86.4 (+0.4)} & \otsmodel{85.6 (+0.6)}\\
\otsmodel{ATC$^{\text{complete}}$ (ours)} & \otsmodel{80.2 (+1.7)} & \otsmodel{67.1 (+10.5)} &\otsmodel{}&\otsmodel{}& \otsmodel{84.9 (+0.7)} & \otsmodel{82.6 (+2.5)}  &\otsmodel{}&\otsmodel{}& \otsmodel{86.4 (+0.4)} & \otsmodel{85.8 (+0.8)}\\ \midrule

ToMe & 81.9 &78.6 &&&85.1 &83.8 &&&86.4 &86.1 \\ 
ATC$^{\text{single}}$ (ours)& 82.2 (+0.3) &80.0 (+1.4) &&&85.3 (+0.2) &\textbf{84.5} (+0.7) &&&86.6 (+0.2) &86.3 (+0.2) \\ 
ATC$^{\text{average}}$ (ours) & 82.3 (+0.4) &\textbf{80.2} (+1.6) &&&85.3 (+0.2) &\textbf{84.5} (+0.7) &&&\textbf{86.7} (+0.3) &86.3 (+0.2) \\ 
ATC$^{\text{complete}}$ (ours) & \textbf{82.5} (+0.6) &80.1 (+1.5)&&&\textbf{85.4} (+0.3) &84.3 (+0.5) &&&\textbf{86.7} (+0.3) &\textbf{86.4} (+0.3) \\ \bottomrule
\end{tabular}%
}
\end{table}
We find that ATC consistently outperforms ToMe across both token reduction schedules and when using off-the-shelf and fine-tuned models, see Table~\ref{tab:mae}. We find that ATC drastically outperforms ToMe when using the linear token scheduler, and when using models with lower backbone capacity. This is especially observable on the ViT-Base backbone where using a linear schedule off-the-shelf leads to a 10.5 percentage point improvement when using ATC instead of ToMe. When fine-tuning the backbone using the same token scheduler this gap is reduced to 1.6 percentage points when fine-tuning, with ATC still outperforming ToMe. This illustrates the clear general benefit of using ATC for adapting already trained ViT backbones, even when using the more aggressive linear token scheduler and without applying any fine-tuning.

\subsection{Image Synthesis with Stable Diffusion}

Bolya \& Hoffman \cite{ToMeSD} demonstrated how a modified ToMe algorithm can be incorporated into an image generation model, specifically Stable Diffusion \cite{stableDiffusion}, without any fine-tuning, resulting in improved quality of the generated images as well as generation speed. We follow the experimental setup of Bolya \& Hoffman, inserting the token reduction method only over the self-attention block. We generate two images with a resolution of $512\times512$ pixels for each class in the ImageNet-1K dataset, following the exact setting from Bolya \& Hoffman \cite{ToMeSD}.

\begin{table}[!tp]
    \caption{\textbf{Stable Diffusion FID Comparison.} We apply ToMe and ATC over the self-attention blocks to an off-the-shelf (\ie frozen) Stable Diffusion model, denoted with a \scolorbox{dt2}{blue} background. We compare the FID score across different keep rates (Note that a lower FID score is better). The best FID score per column is denoted in \textbf{bold}, and the best per row is \underline{underlined}.}
    \label{tab:sd}
    \centering
    \begin{tabular}{lccccccc}
    \toprule
        $r$ (\%) & 100 & 90 & 80 & 70 & 60 & 50 & 40 \\ \midrule
        \otsmodel{ToMe} & \otsmodel{\textbf{33.80}} &  \otsmodel{33.77} & \otsmodel{33.61} & \otsmodel{33.59} &  \otsmodel{\underline{33.57}}& \otsmodel{33.63} & \otsmodel{33.67} \\
        \otsmodel{ATC$^{\text{single}}$ (ours)}& \otsmodel{\textbf{33.80}} & \otsmodel{33.63} &\otsmodel{33.48} & \otsmodel{33.53} & \otsmodel{\underline{33.43}} & \otsmodel{34.01} & \otsmodel{36.95} \\
        \otsmodel{ATC$^{\text{average}}$ (ours)} & \otsmodel{\textbf{33.80}} &\otsmodel{\textbf{33.51}} & \otsmodel{\textbf{33.45}} & \otsmodel{\textbf{33.33}} & \otsmodel{\textbf{33.40}} & \otsmodel{\textbf{33.36}} &  \otsmodel{\underline{\textbf{33.22}}} \\
        \otsmodel{ATC$^{\text{complete}}$ (ours)} & \otsmodel{\textbf{33.80}} & \otsmodel{33.67} & \otsmodel{33.71} & \otsmodel{33.70} & \otsmodel{33.74} & \otsmodel{\underline{33.56}} & \otsmodel{33.65} \\ \bottomrule
    \end{tabular}
\end{table}
We implement the setup using the HuggingFace Diffusers framework \cite{Diffusers}, use the \textsc{stable-diffusion-v1-5} model, and use the exact same seed for each model. We investigate the effect when using keep rates of $r\in\{40,50,60,70,80,90\}$\%. In order to evaluate the quality of the generated images, we measure the Fréchet Inception Distance (FID) score \cite{FID} between the generated images and a reference dataset consisting of 5000 images, created by taking the first five validation images per ImageNet-1K class.\\

\begin{figure}[!tp]
\centering
\begin{tabular}{cccc}

ToMe  & ATC$^\text{single}$  & ATC$^\text{average}$  & ATC$^\text{complete}$\\
\includegraphics[width=0.24\linewidth]{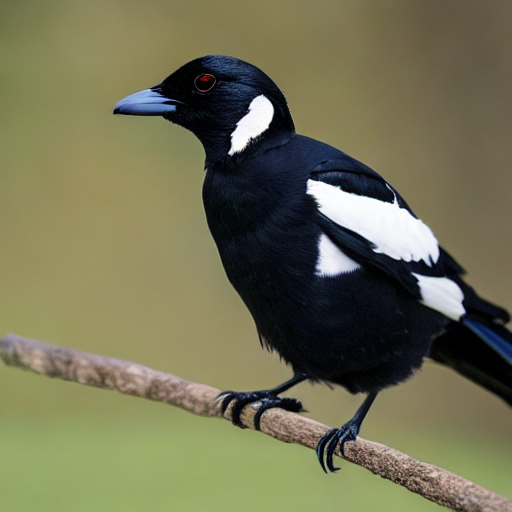} & \includegraphics[width=0.24\linewidth]{sd_example/sd.png} & \includegraphics[width=0.24\linewidth]{sd_example/sd.png} & \includegraphics[width=0.24\linewidth]{sd_example/sd.png} \\
\includegraphics[width=0.24\linewidth]{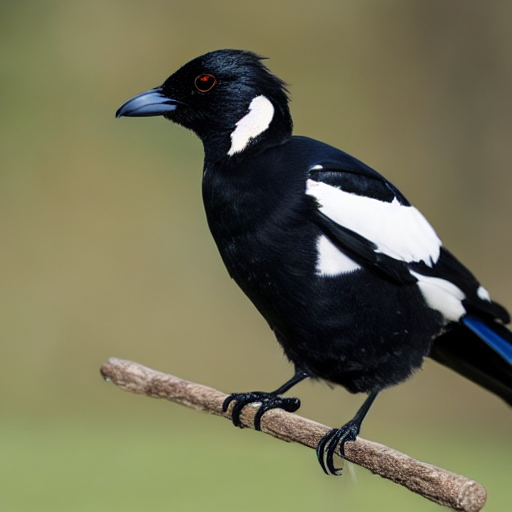} & \includegraphics[width=0.24\linewidth]{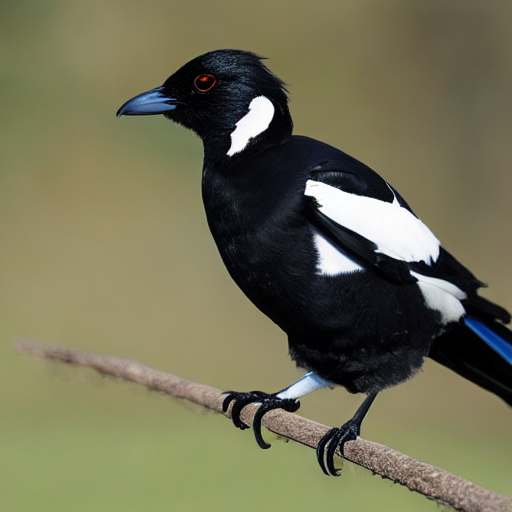} & \includegraphics[width=0.24\linewidth]{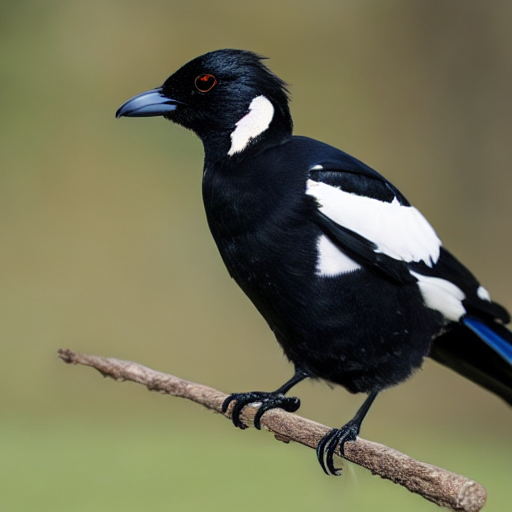} & \includegraphics[width=0.24\linewidth]{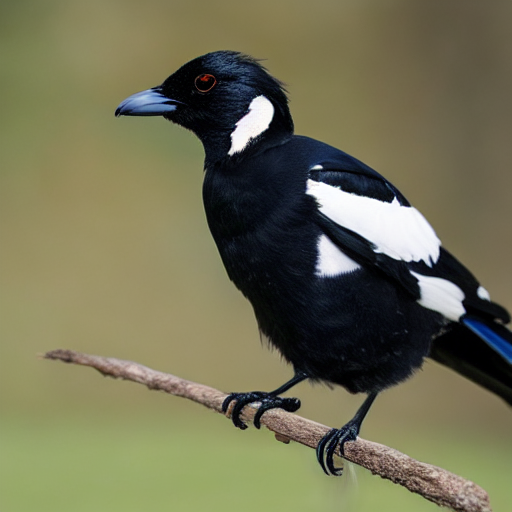} \\
\includegraphics[width=0.24\linewidth]{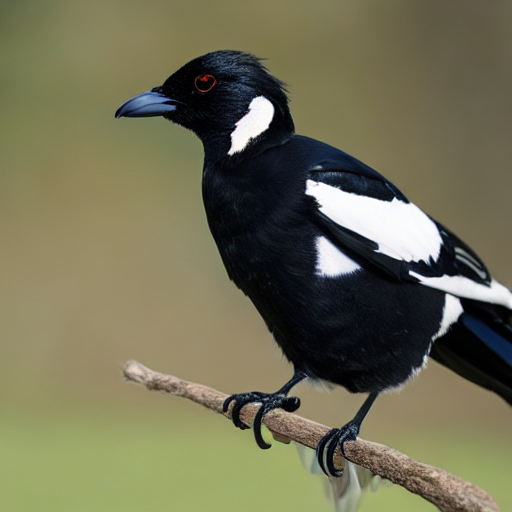} & \includegraphics[width=0.24\linewidth]{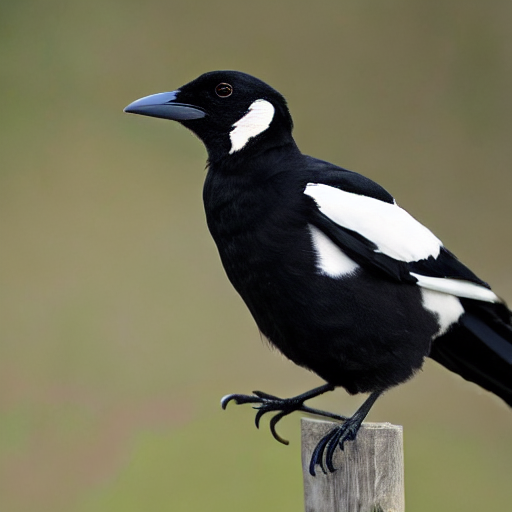} & \includegraphics[width=0.24\linewidth]{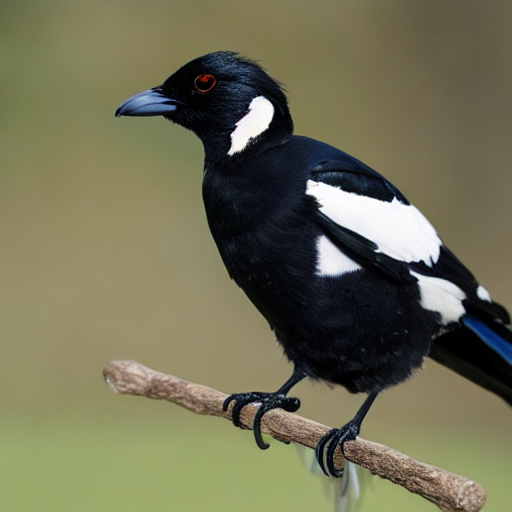} & \includegraphics[width=0.24\linewidth]{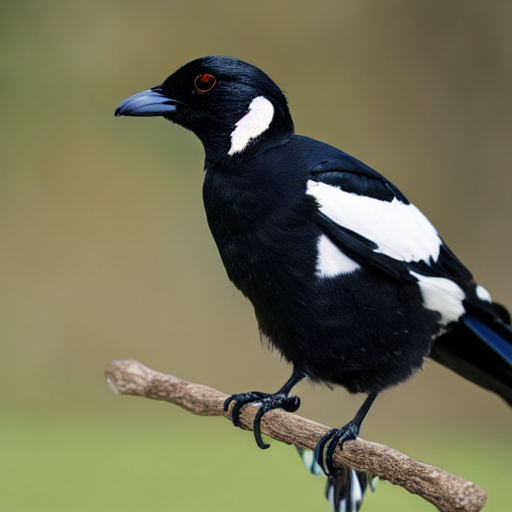}  \\
 \includegraphics[width=0.24\linewidth]{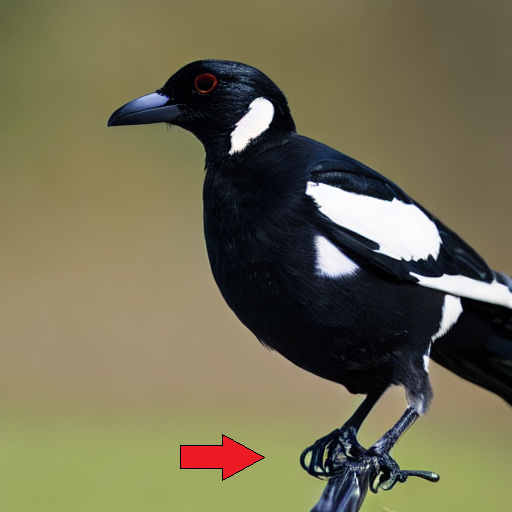} & \includegraphics[width=0.24\linewidth]{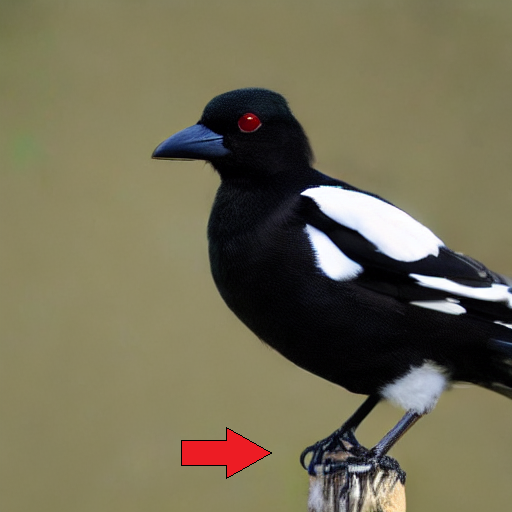} & \includegraphics[width=0.24\linewidth]{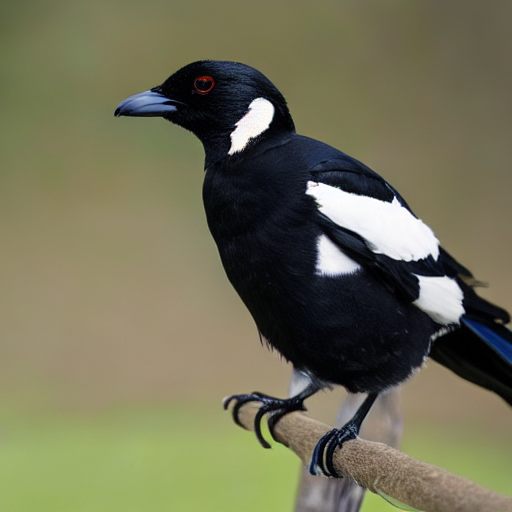} & \includegraphics[width=0.24\linewidth]{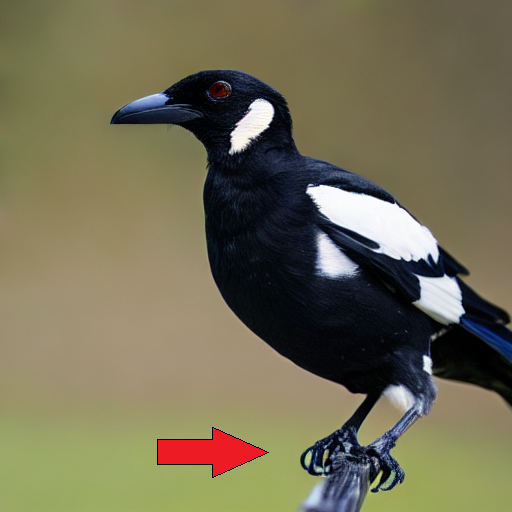} \\

\end{tabular}
\caption{\textbf{Image Synthesis Visualization.} We visualize image synthesis results for the ``magpie'' ImageNet class. The first row is the standard Stable Diffusion result, with each subsequent row having token merging applied with $r\in\{80,60,40\}\%$. Examples for all considered keep rates and more classes can be found in the supplementary materials. We find that as the keep rate is lowered the ATC$^\text{single}$ method drastically changes the image, specifically the head and patterns of the magpie, as well as having a more monochrome background. We see that even at $r=40\%$ ATC$^\text{average}$ manages to keep most of the details such as the branch the magpie is sitting on, whereas ToMe and ATC$^\text{complete}$ keep the general patterns but switches from a branch to a wooden pole (highlighted with a red arrow), leading to a change in pose of the generated magpie.}
\label{fig:sdViz}
\end{figure}

We compare the FID scores of the ToMe and ATC models in Table~\ref{tab:sd}. The ToMe results are computed using the same setup as the ATC models, and therefore differ from the original paper. Even though our generated images lead to a generally higher (thus worse) FID for ToMe, we find that the general trends observed in the original paper still hold. We find that across all linkage functions the ATC model outperforms or matches the ToMe model. While ToMe achieves a minimum FID of 33.57, this is outperformed by both the single and average linkage functions with FID scores of 33.43 and 33.22, respectively. The complete linkage function in general performs worse than both the single and average linkage functions, but we also find that the results with single linkage diverges when $r\leq 50\%$. We observe that the average linkage function leads to better results as the keep rate is reduced, achieving the best performance when $r=40\%$, whereas the other methods peak at $r=50\%$ and $r=60\%$. By looking at the generated images, see Figure~\ref{fig:sdViz}, we find that the average linkage function manages to keep a lot more of the distinctive patterns and contextual background. In comparison, the single linkage function loses the head pattern, and all methods except ATC$^\text{average}$ convert the tree branch to a tree pole resulting in a change in pose of the generated magpie.

\begin{table}[!t]
    \caption{\textbf{Object Detection \& Segmentation Results.} We compare the performance of ATC and ToMe on the dense object detection \& segmentation task using the COCO 2017 dataset, following Liu \etal \cite{SViT}. The ViT-Adapter method is used, with DeiT-T and DeiT-S backbones. The best performing method per column is denoted in \textbf{bold}. A \scolorbox{dt2}{blue} background indicates that ATC and ToMe were applied off-the-shelf on the frozen ViT-Adapter backbone, whereas all other models have been fine-tuned. }
    \label{tab:objdetseg}
    \begin{subtable}[t]{0.5\linewidth}
    \caption{\textbf{ViT-Adapter-T Backbone}}
    \label{tab:vitadaptT}
    \centering
    
    \resizebox{\linewidth}{!}{%
    \begin{tabular}{lccrcccrc}
    
    \toprule
     & \multicolumn{4}{c}{mAP$^{\text{box}}$}  & \multicolumn{4}{c}{mAP$^{\text{mask}}$}  \\ \cmidrule(lr){2-5} \cmidrule(lr){6-9}  
    ViT-Adapter-T & \multicolumn{4}{c}{45.8} & \multicolumn{4}{c}{40.9} \\ \midrule\midrule
    $r$  (\%) & 25 & 50 & \multicolumn{1}{c}{70} & 90 & 25 & 50 & \multicolumn{1}{c}{70} & 90 \\ \midrule
\otsmodel{ToMe} & \otsmodel{-} & \otsmodel{39.9} & \otsmodel{45.3} & \otsmodel{45.8} & \otsmodel{-} & \otsmodel{36.2} & \otsmodel{40.5} & \otsmodel{40.8}  \\ 
\otsmodel{ATC$^{\text{single}}$ (ours)} & \otsmodel{13.9} & \otsmodel{38.5} & \otsmodel{44.8} & \otsmodel{45.7} & \otsmodel{12.6} & \otsmodel{34.6} & \otsmodel{39.9} & \otsmodel{40.8}  \\ 
\otsmodel{ATC$^{\text{average}}$ (ours)} & \otsmodel{33.8} & \otsmodel{43.7} & \otsmodel{45.5} & \otsmodel{45.8} & \otsmodel{31.5} & \otsmodel{39.2} & \otsmodel{40.7} & \otsmodel{40.8}  \\ 
\otsmodel{ATC$^{\text{complete}}$ (ours)} & \otsmodel{34.9} & \otsmodel{43.9} & \otsmodel{45.6} & \otsmodel{45.8} & \otsmodel{32.3} & \otsmodel{39.3} & \otsmodel{40.6} & \otsmodel{40.8}   \\  \midrule
ToMe& -& 43.7& 45.7& \textbf{46.0}& -& 39.3& 40.9& 41.0  \\ 
ATC$^{\text{single}}$ (ours)& 36.9& 43.5& 45.5& 45.9& 33.6& 39.1& 40.7& 40.9  \\ 
ATC$^{\text{average}}$ (ours)& 42.4& 45.2& 45.8& 45.9& 38.5& \textbf{40.5} & 40.8& \textbf{41.1}  \\ 
ATC$^{\text{complete}}$ (ours)& \textbf{42.6}& \textbf{45.3}& \textbf{45.9}& \textbf{46.0}& \textbf{38.7}& \textbf{40.5}& \textbf{41.0}& \textbf{41.1} \\ \bottomrule
    \end{tabular}
        }
    \end{subtable} \hfill    
    \begin{subtable}[t]{0.5\linewidth}
    \caption{\textbf{ViT-Adapter-S Backbone}}
    \label{tab:vitadaptS}
    \centering
    
    \resizebox{\linewidth}{!}{%
    \begin{tabular}{lccrcccrc}
    \toprule
     & \multicolumn{4}{c}{mAP$^{\text{box}}$}  & \multicolumn{4}{c}{mAP$^{\text{mask}}$}  \\ \cmidrule(lr){2-5} \cmidrule(lr){6-9}  
    ViT-Adapter-S & \multicolumn{4}{c}{48.5} & \multicolumn{4}{c}{42.8} \\ \midrule\midrule
    $r$  (\%) & 25 & 50 & \multicolumn{1}{c}{70} & 90 & 25 & 50 & \multicolumn{1}{c}{70} & 90 \\  \midrule
\otsmodel{ToMe} & \otsmodel{-} & \otsmodel{42.4} & \otsmodel{47.8} & \otsmodel{48.4} & \otsmodel{-} & \otsmodel{38.2} & \otsmodel{42.3} & \otsmodel{42.7}  \\ 
\otsmodel{ATC$^{\text{single}}$ (ours)} & \otsmodel{16.9} & \otsmodel{41.9} & \otsmodel{47.4} & \otsmodel{48.4} & \otsmodel{14.4} & \otsmodel{36.8} & \otsmodel{41.8} & \otsmodel{42.7}  \\ 
\otsmodel{ATC$^{\text{average}}$ (ours)} & \otsmodel{37.2} & \otsmodel{46.5} & \otsmodel{48.0} & \otsmodel{48.4} & \otsmodel{33.7} & \otsmodel{41.2} & \otsmodel{42.4} & \otsmodel{42.7}  \\ 
\otsmodel{ATC$^{\text{complete}}$ (ours)} & \otsmodel{38.3} & \otsmodel{46.6} & \otsmodel{48.0} & \otsmodel{48.4} & \otsmodel{34.7} & \otsmodel{41.3} & \otsmodel{42.4} & \otsmodel{42.7} \\\midrule
ToMe& -& 46.4& 48.2& 48.3& -& 41.4& \textbf{42.6}& 42.7  \\ 
ATC$^{\text{single}}$ (ours)& 40.3& 46.4& 47.9& 48.3& 36.2& 41.1& 42.3& 42.6  \\ 
ATC$^{\text{average}}$ (ours)& 45.1& \textbf{47.8}& \textbf{48.3}& 48.5& 40.2& \textbf{42.4}& \textbf{42.6}& \textbf{42.9}  \\ 
ATC$^{\text{complete}}$ (ours)& \textbf{45.3}& \textbf{47.8}& 48.2& \textbf{48.6}& \textbf{40.5}& 42.3& \textbf{42.6}& 42.7 \\ \bottomrule
    \end{tabular}
    }
    \end{subtable}
\end{table}

\subsection{Object Detection and Segmentation}\label{sec:objdetseg}
Liu \etal \cite{SViT} conducted a systematic comparison of several token pruning methods for object detection and segmentation on the COCO 2017 dataset \cite{COCO}. The Mask-RCNN model \cite{maskrcnn} is used for predicting bounding boxes and segmentation masks with a ViT-Adapter backbone model \cite{ViTAdapter}, building upon an ImageNet pretrained DeiT model \cite{DeiT_2021}. Instead of the typical windowed self-attention used in ViT-Adapter, Liu \etal use global self-attention and train using the original ViT-Adapter settings for 36 epochs. Tokens are reduced at the 4th, 7th, and 10th ViT block, and a DeiT-Tiny and DeiT-Small backbone are used. When applying the token reduction method, the ViT-Adapter-Tiny and ViT-Adapter-Small models are fine-tuned for 6 and 4 epochs, respectively, using the MMDetection framework \cite{MMDetection} and following the training protocol of Liu \etal.
Unlike the original protocol by Liu \etal which only considered a keep rate of $r=70\%$, we extend the considered keep rates to $r\in\{25, 50, 70, 90\}$\%, similar to the protocol used by Haurum \etal \cite{Haurum_2023_ICCV}. We evaluate both ToMe and ATC in the off-the-shelf and fine-tuned scenarios. As the Injector and Extractor modules in the ViT-Adapter expect the original number of tokens, we back-project through the clustering when relevant, leading to the original number of patches.\\

As is apparent in Table~\ref{tab:objdetseg}, we find that both ToMe and ATC are very strong token reduction methods for detection and segmentation.
When applying ToMe and ATC off-the-shelf (\ie without fine-tuning the ViT-Adapter backbone) with keep rates $r\in\{70, 90\}\%$, both methods can match the detection and segmentation performance of the baseline ViT-Adapter backbones. When the keep rate is lowered to $r=50\%$ we find that our ATC method with the average and complete linkage function outperforms ToMe by 4 percentage points in detection mAP and 3 percentage points in segmentation mAP.

When fine-tuning the Tiny and Small backbones we see major improvements at keep rates of 25\% and 50\%, such as a 26 and 22 percentage points improvement in bounding box mAP for ATC$^{\text{single}}$ with the Tiny and Small backbones and $r=25\%$, respectively. In comparison, fine-tuning has less of an effect on ATC$^{\text{average}}$ and ATC$^{\text{complete}}$ as the off-the-shelf performance is already high. We also observe that after fine-tuning, our ATC method still outperforms ToMe, though with a smaller margin, and in several cases outperforms the baseline ViT-Adapter performance. Lastly, we see that by fine-tuning, we can get comparable performance to the baseline method at keep rates of 50\%, whereas ToMe is several percentage points worse than the baseline ViT-Adapter.

\section{Discussion}

Through our experiments we have demonstrated how our proposed ATC method consistently outperforms prior merging-based token reduction methods across a diverse set of tasks. This includes the prior best merging-based method ToMe, which we confidently outperform across all considered tasks. While our analysis have been focused on comparing ATC with prior merging-based approaches, we also find that ATC outperforms pruning-based methods on the image classification task (Following the setup from Sec.~\ref{sec:iccvDataset}) and object detection and segmentation, where ATC outperforms the prior state-of-the-art pruning-based SViT method from Liu \etal \cite{SViT}. Detailed experimental results including the pruning-based methods are available in the supplementary materials, and establishes that our ATC method is the current best token reduction method.
\section{Limitations}
We show that ATC is a very adaptable method, achieving state-of-the-art performance across the different classification, image synthesis, and object detection \& segmentation tasks considered. However, ATC is not without its limitations. Firstly, ATC requires the selection of a linkage function, adding an extra hyperparameter. While there is not a specific linkage function that is the best at all tasks, we find that the average or complete linkage functions are good general choices, whereas the single linkage function often underperforms when working with more aggressive keep rates, matching prior linkage function recommendations \cite{ClusterBook}. Secondly, we find that the inference throughput of the current implementation of ATC is hampered by the available implementations of the agglomerative clustering functions, which are all non-batched and often CPU-bounded. This is discussed and analyzed at lengths in the supplementary materials. However, there are also clear indications that these limitations can be lifted if the current frameworks are slightly modified. 
\section{Conclusion}
In this work, we introduce Agglomerative Token Clustering (ATC), a novel hard-merging token reduction approach grounded in the principles of classical bottom-up hierarchical clustering. ATC distinguishes itself by efficiently merging redundant observations early on, thus preserving the semantic richness of more diverse observations. We evaluate our method across the most diverse sets of tasks considered in the literature, covering both classification, synthesis, detection, and segmentation tasks.  In image classification, image synthesis, and object detection \& segmentation, ATC sets a new state-of-the-art, highlighting its significant contribution to the field. We also demonstrate that ATC can achieve comprable performance to the prior state-of-the-art without any fine-tuning, \ie when applied \textit{off-the-shelf}. We are optimistic that ATC will inspire subsequent advancements, leveraging classical clustering methods to enhance modern neural architectures.

\section*{Acknowledgements} This work was supported by the Pioneer Centre for AI (DNRF grant number P1), partially supported by the Spanish project PID2022-136436NB-I00 and by ICREA under the ICREA Academia programme, and the Canada CIFAR AI Chairs.

\appendix
\section{Overview of Supplementary Materials}
In these supplementary materials we describe in further detail aspects of the training process, performance of the tested models, in-depth inference analysis across the considered tasks, and visual examples from the image classification and synthesis tasks. We refer to both the relevant sections in the main manuscript as well as sections in the supplementary materials. Specifically, the following will be described:

\begin{itemize}
    \itemsep0em
    \item Hyperparameters for the four classification datasets (Section~4.1 / \ref{sec:hypers}). 
    \item Performance of Agglomerative Token Clustering (ATC) compared to all considered token reduction methods on different image classification datasets (Section~4.1 / \ref{sec:methodComp}).
    \item Off-the-shelf performance of ATC on different image classification datasets (Section~4.1 / \ref{sec:methodCompOTS}).
    \item Off-the-shelf comparison with different pretrained ViTs (Section~4.2 / \ref{sec:weights}).  
    \item Performance of ATC compared to all considered token reduction methods for object detection and segmentation (Section~4.4 / \ref{sec:methodCompObj}).  
    \item In-depth inference throughput analysis of ATC (Section~6 / \ref{sec:inference}). 
    \item Per-dataset example visualization of the hard-merging reduction patterns (Section~4.1 / \ref{sec:reducViz}).
    \item Examples of images generated in the image synthesis task (Section~4.3 / \ref{sec:synthesisViz}).
\end{itemize}

\section{Hyperparameters}\label{sec:hypers}

We follow the hyperparameter search originally used by Haurum \etal \cite{Haurum_2023_ICCV}, when training the DeiT models with ATC on ImageNet, NABirds, COCO, and NUS-Wide. Specifically, we compare the number of warmup epochs (W-E), the number of epochs the backbone is frozen (B-FE), as well as the learning rate of the backbone (B-LR). We determine the best setting on ImageNet using a DeiT-S backbone, Table~\ref{tab:imagenetHyper}, and consider these hyperparameter settings for the other three dataset. We compare the ImageNet hyperparameters with the hyperparameters used to train the per-dataset baseline DeiT-S model, originally determined by Haurum \etal, and use the settings of the best hyperparameter set. We denote the ImageNet set as $\mathcal{I}$ and dataset specific set as $\mathcal{D}$, see Table~\ref{tab:hyperAssign}.

\begin{table*}[!htp]
\caption{\textbf{Selected token reduction method hyperparameters - ImageNet.} We present the selected hyperparameters when searching on ImageNet for ATC with the three considered linkage functions.}
\label{tab:imagenetHyper}
\centering
\begin{tabular}{@{}lccclccclccclccc@{}}
\toprule
 $r$ (\%) & \multicolumn{3}{c}{25} &  & \multicolumn{3}{c}{50} &  & \multicolumn{3}{c}{70} &  & \multicolumn{3}{c}{90} \\\cmidrule(lr){2-4} \cmidrule(lr){6-8} \cmidrule(lr){10-12} \cmidrule(lr){14-16}
 & W-E & B-LR & B-FE & & W-E & B-LR & B-FE & & W-E & B-LR & B-FE & & W-E & B-LR & B-FE \\ \midrule
ATC$^\text{single}$ & 5 & 0.01 & 5 & & 5 & 0.01 & 5 & & 5 & 0.01 & 5 & & 20 & 1 & 0 \\
ATC$^\text{average}$ & 5 & 0.01 & 5 & & 20 & 0.01 & 5 & & 5 & 0.01 & 5 & & 5 & 0.01 & 5 \\
ATC$^\text{complete}$ & 5 & 0.01 & 5 & & 20 & 0.01 & 5 & & 5 & 0.01 & 5 & & 20 & 1 & 0 \\ \bottomrule
\end{tabular}
\end{table*}

\begin{table*}[!htp]
\caption{\textbf{Hyperparameter indicator matrix.} We illustrate below for each method and keep rate $r$ whether the hyperparameter settings from the ImageNet dataset ($\mathcal{I}$) or the dataset specific DeiT-S baseline ($\mathcal{D}$) are used.}
\label{tab:hyperAssign}
\centering
\begin{tabular}{@{}lcccclcccclcccc@{}}
\toprule
 & \multicolumn{4}{c}{NABirds} &  & \multicolumn{4}{c}{COCO} &  & \multicolumn{4}{c}{NUS-WIDE} \\\cmidrule(lr){2-5} \cmidrule(lr){7-10} \cmidrule(lr){12-15}
$r$  (\%) & 25 & 50 & 70 & 90 &  & 25 & 50 & 70 & 90 &  & 25 & 50 & 70 & 90 \\ \midrule
ATC$^\text{single}$ &$\mathcal{D}$  & $\mathcal{D}$ & $\mathcal{D}$ & $\mathcal{D}$ & & $\mathcal{D}$ & $\mathcal{D}$ & $\mathcal{D}$ & $\mathcal{D}$& & $\mathcal{D}$ & $\mathcal{D}$ & $\mathcal{D}$ & $\mathcal{I}$  \\
ATC$^\text{average}$ &$\mathcal{D}$  & $\mathcal{D}$ & $\mathcal{D}$ & $\mathcal{D}$ & & $\mathcal{D}$ & $\mathcal{D}$ & $\mathcal{D}$ & $\mathcal{D}$ & & $\mathcal{D}$ & $\mathcal{D}$ & $\mathcal{D}$ & $\mathcal{D}$ \\
ATC$^\text{complete}$ & $\mathcal{D}$ & $\mathcal{D}$ & $\mathcal{D}$ & $\mathcal{D}$ & & $\mathcal{D}$ & $\mathcal{D}$ & $\mathcal{D}$ & $\mathcal{D}$ &  & $\mathcal{D}$ & $\mathcal{D}$ & $\mathcal{D}$ & $\mathcal{I}$  \\ \bottomrule
\end{tabular}
\end{table*}

\section{Comparison of Token Reduction Methods on Image Classification Datasets}\label{sec:methodComp}

\begin{figure*}[!tp]
     \centering
     \begin{subfigure}[b]{0.495\textwidth}
         \centering
         \includegraphics[width=\textwidth]{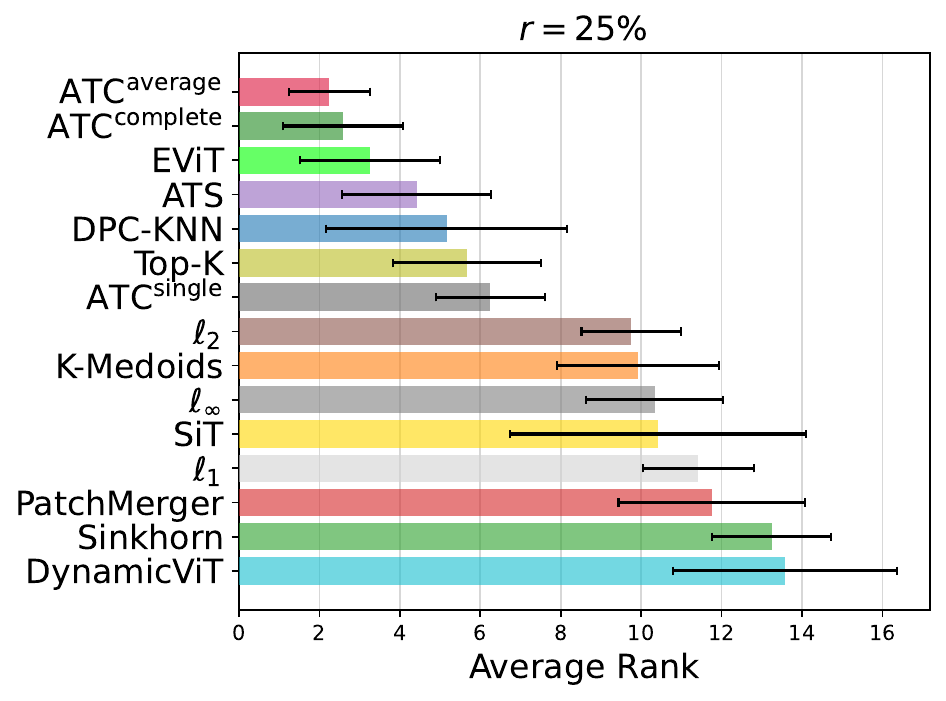}
         \caption{$r=25\%$}
         \label{fig:r025}
     \end{subfigure}
     \hfill
     \begin{subfigure}[b]{0.495\textwidth}
         \centering
         \includegraphics[width=\textwidth]{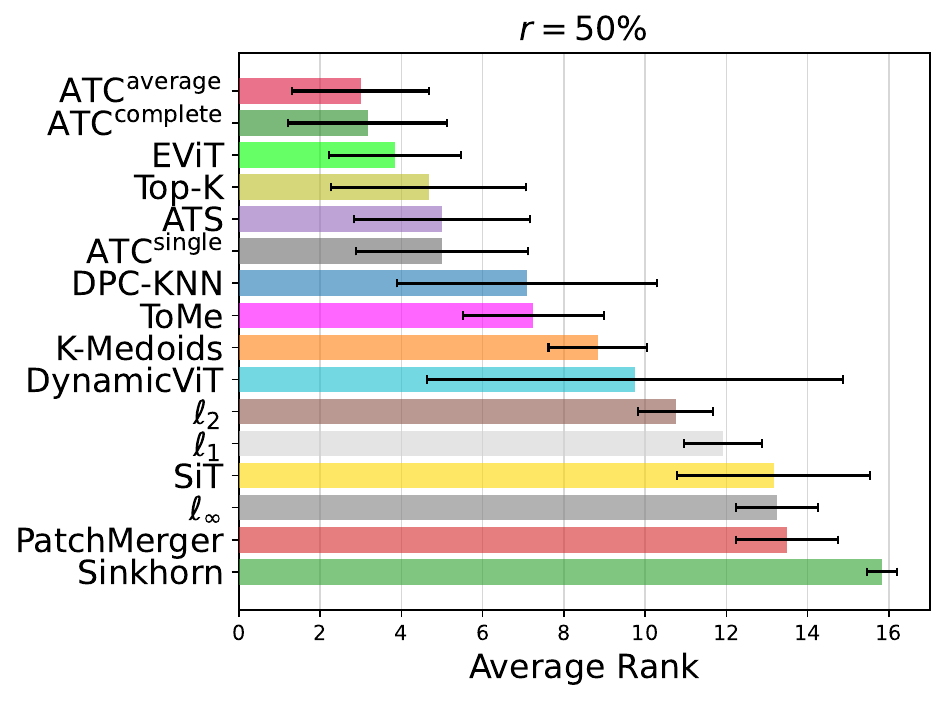}
         \caption{$r=50\%$}
         \label{fig:r50}
     \end{subfigure}
     \hfill
     \begin{subfigure}[b]{0.495\textwidth}
         \centering
         \includegraphics[width=\textwidth]{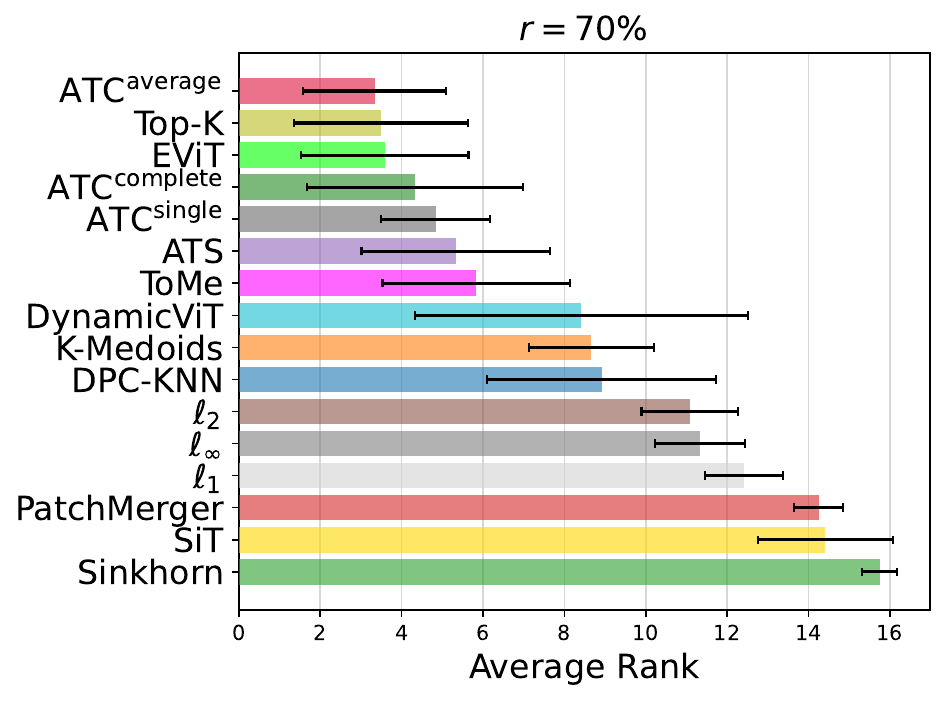}
         \caption{$r=70\%$}
         \label{fig:r70}
     \end{subfigure}
     \hfill
     \begin{subfigure}[b]{0.495\textwidth}
         \centering
         \includegraphics[width=\textwidth]{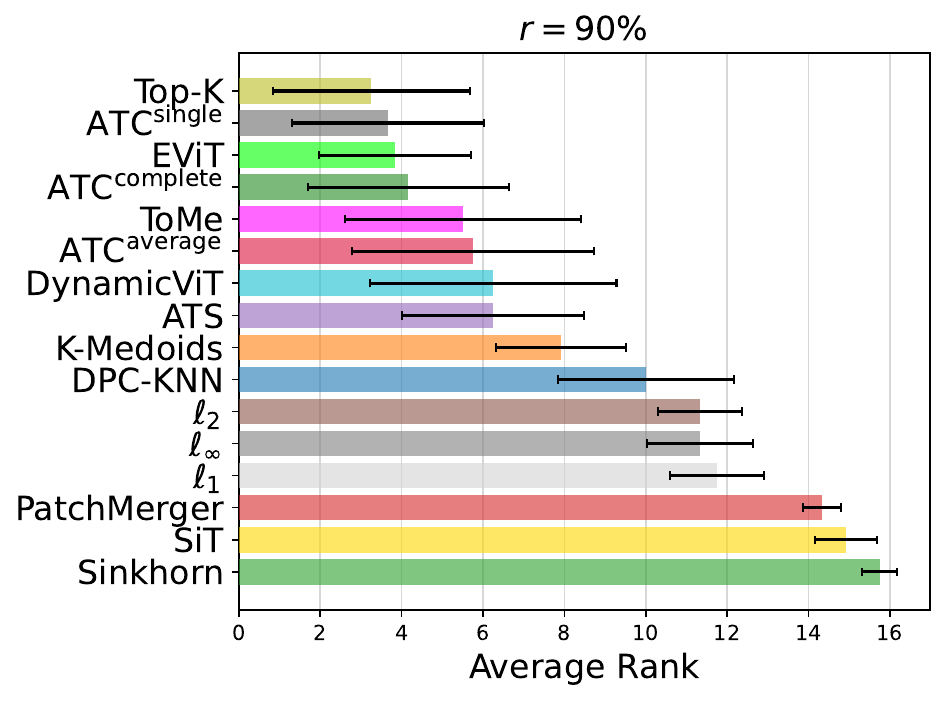}
         \caption{$r=90\%$}
         \label{fig:r90}
     \end{subfigure}
        \caption{\textbf{Average Rankings per Keep Rates (Lower is Better).} We compare ATC with the methods investigated by Haurum et al. \cite{Haurum_2023_ICCV}, reporting the average rank per keep rate. We average across the three considered model capacities and four datasets, and plot with ±1 standard deviation similar to Haurum et al. We test three versions of ATC, varying the linkage function, and find that ATC$^\text{average}$ is the best method for $r\leq70\%$, while dropping to the sixth place for $r=90\%$. }
        \label{fig:avgRanking}
\end{figure*}

In this section we present a full comparison between the fine-tuned ATC methods and all methods considered by Haurum \etal, see Table~\ref{tab:perfICCV}. We also determine the average rank of all considered methods per keep rate, see Figure~\ref{fig:avgRanking}. We find that for all keep rates ATC is the best performing merging-based method, and for all but r=90\% ATC is the best overall token reduction method.

\begin{table*}[!htp]
\caption{\textbf{Full Comparison of Token Reduction methods.} Full results of ATC and the token reduction methods investigated by Haurum \etal \cite{Haurum_2023_ICCV}. This table is adapted from the relevant tables in \cite{Haurum_2023_ICCV}. Performance is measured across keep rates and backbone capacities. Scores exceeding the DeiT baseline are noted in \textbf{bold}, measured as Top-1 accuracy for ImageNet \& NABirds and mean Average Precision for COCO \& NUS-WIDE. The three best performing methods per keep rate are denoted in descending order with \colorbox{red!25}{red}, \colorbox{orange!25}{orange}, and \colorbox{yellow!25}{yellow}, respectively. Similarly, the three worst performing methods are denoted in descending order with \colorbox{blue!10}{light blue}, \colorbox{blue!20}{blue}, and \colorbox{blue!30}{dark blue}.}
\label{tab:perfICCV}
\centering
\resizebox{0.85\textwidth}{!}{%
\begin{tabular}{lcccclcccclcccclcccc}
\toprule
 & \multicolumn{4}{c}{ImageNet} &  & \multicolumn{4}{c}{NABirds} &  & \multicolumn{4}{c}{COCO} &  & \multicolumn{4}{c}{NUS-WIDE} \\ \cmidrule(lr){2-5} \cmidrule(lr){7-10} \cmidrule(lr){12-15} \cmidrule(lr){17-20}  
 
DeiT-T & \multicolumn{4}{c}{72.20} & & \multicolumn{4}{c}{74.16} & & \multicolumn{4}{c}{71.09} & & \multicolumn{4}{c}{59.27} \\ \midrule\midrule
$r$  (\%) & 25 & 50 & 70 & 90 &  & 25 & 50 & 70 & 90 &  & 25 & 50 & 70 & 90 &  & 25 & 50 & 70 & 90 \\ \midrule
$\ell_1$ & 58.58 & \cellcolor{blue!10}62.27 & 67.91 & 71.06 &  & 51.82 & 59.25 & 68.36 & 73.47 &  & \cellcolor{blue!10}49.09 & \cellcolor{blue!10}58.27 & 67.03 & 70.24 &  & \cellcolor{blue!10}44.81 & 52.64 & 57.30 & 58.73 \\
$\ell_2$ & 58.85 & 62.91 & 67.91 & 71.13 &  & 53.10 & 60.87 & 69.20 & 73.42 &  & 50.46 & 60.00 & 67.33 & 69.98 &  & 45.73 & 53.45 & 57.34 & 58.54 \\
$\ell_\infty$ & 59.08 & \cellcolor{blue!20}61.92 & \cellcolor{blue!10}67.79 & 71.38 &  & 52.60 & 57.70 & 69.25 & 73.34 &  & 50.08 & 57.30 & 67.22 & 69.89 &  & 45.32 & 51.91 & 57.41 & 58.30 \\ \midrule
Top-K & 62.19 & 68.55 & 70.96 & 71.85 &  & 62.14 & \cellcolor{orange!25}73.19 & \cellcolor{red!25}\textbf{74.57} & \cellcolor{red!25}\textbf{74.64} &  & 60.31 & 67.47 & \cellcolor{orange!25}70.20 & \cellcolor{red!25}\textbf{71.65} &  & 52.20 & 57.02 & \cellcolor{orange!25}58.60 & \cellcolor{yellow!25}\textbf{59.50} \\
EViT & 64.11 & 68.69 & 71.06 & 71.83 &  & \cellcolor{yellow!25}64.13 & \cellcolor{red!25}73.24 & \cellcolor{orange!25}\textbf{74.49} & \cellcolor{orange!25}\textbf{74.53} &  & 61.44 & 67.62 & \cellcolor{red!25}70.25 & \cellcolor{orange!25}\textbf{71.63} &  & 53.09 & 57.26 & \cellcolor{red!25}58.64 & \textbf{59.49} \\
DynamicViT & \cellcolor{blue!30}36.93 & 67.40 & 70.94 & 72.14 &  & 57.38 & \cellcolor{yellow!25}72.54 & \cellcolor{yellow!25}73.97 & \textbf{74.30} &  & \cellcolor{blue!30}24.67 & 61.70 & 68.83 & \cellcolor{yellow!25}\textbf{71.30} &  & \cellcolor{blue!30}28.09 & \cellcolor{blue!10}49.36 & 56.79 & 58.95 \\
ATS & 62.63 & 68.61 & 70.77 & 71.71 &  & \cellcolor{red!25}64.53 & 71.07 & 73.71 & \textbf{74.43} &  & 60.97 &67.37 & 69.88 & \textbf{71.10} &  & 52.85 & \cellcolor{yellow!25}57.30 & \cellcolor{yellow!25}58.55 & 59.20 \\ \midrule

ToMe & - & 69.72 & \cellcolor{yellow!25}71.74 & \cellcolor{orange!25}72.16 &  & - & 66.61 & 73.65 & \cellcolor{yellow!25}\textbf{74.50} &  & - & 65.66 & 69.70 & \textbf{71.16} &  & - & 55.32 & 57.78 & 58.98 \\
K-Medoids & \cellcolor{blue!10}57.50 & 65.82 & 69.90 & 71.50 &  & 44.62 & 66.52 & 72.09 & 74.04 &  & 54.08 & 64.83 & 69.09 & 71.05 &  & 49.13 & 55.92 & 58.38 & 59.07 \\
DPC-KNN & \cellcolor{yellow!25}64.56 & 69.68 & 71.10 & 71.88 &  & \cellcolor{orange!25}64.23 & 71.05 & 73.02 & 74.05 &  & \cellcolor{yellow!25}63.32 & 68.03 & 69.55 & 70.84 &  & \cellcolor{yellow!25}55.37 & \cellcolor{orange!25}57.33 & 58.08 & 58.88 \\ \midrule
SiT & 63.43 & 67.98 & 68.99 & \cellcolor{blue!10}68.90 &  & \cellcolor{blue!30}36.35 & \cellcolor{blue!20}36.65 & \cellcolor{blue!30}34.00 & \cellcolor{blue!30}35.07 &  & \cellcolor{blue!20}48.01 & \cellcolor{blue!30}47.50 & \cellcolor{blue!30}46.98 & \cellcolor{blue!30}46.48 &  & \cellcolor{blue!20}36.67 & \cellcolor{blue!30}38.15 & \cellcolor{blue!30}36.98 & \cellcolor{blue!30}37.70 \\
PatchMerger & 60.38 & 64.80 & \cellcolor{blue!20}66.81 & \cellcolor{blue!20}68.09 &  & \cellcolor{blue!10}38.83 & \cellcolor{blue!10}54.20 & \cellcolor{blue!10}59.94 & \cellcolor{blue!10}62.60 &  & 52.49 & 59.69 & \cellcolor{blue!10}62.63 & \cellcolor{blue!10}64.30 &  & 47.56 & 52.69 & \cellcolor{blue!10}54.33 & \cellcolor{blue!10}55.37 \\
Sinkhorn & \cellcolor{blue!20}53.61 & \cellcolor{blue!30}53.49 & \cellcolor{blue!30}53.19 & \cellcolor{blue!30}53.51 &  & \cellcolor{blue!20}36.94 & \cellcolor{blue!30}35.98 & \cellcolor{blue!20}37.29 & \cellcolor{blue!20}36.19 &  & 50.47 & \cellcolor{blue!20}49.52 & \cellcolor{blue!20}49.12 & 4\cellcolor{blue!20}9.01 &  & 45.77 & \cellcolor{blue!20}44.81 & \cellcolor{blue!20}44.52 & \cellcolor{blue!20}44.20 \\ \midrule
ATC$^\text{single}$ (ours) & 61.79 & \cellcolor{yellow!25}69.93 & \cellcolor{yellow!25}71.74 & \cellcolor{yellow!25}72.15 & & 59.27 & 71.90 & 73.86 & \textbf{74.18} & & 60.26 & \cellcolor{yellow!25}68.22 & \cellcolor{yellow!25}70.19 & 70.52 & & 53.36 & \cellcolor{red!25}57.81 & 58.01 & \cellcolor{orange!25}\textbf{60.19}\\
ATC$^\text{average}$ (ours) & \cellcolor{orange!25}66.52 & \cellcolor{orange!25}70.98 & \cellcolor{orange!25}71.83 & \cellcolor{red!25}\textbf{72.21} & & 63.96 & 72.00 & 73.72 & \textbf{74.23} & & \cellcolor{red!25}65.32 & \cellcolor{red!25}68.99 & 69.91 & 70.48 & & \cellcolor{orange!25}56.64 & 57.22 & 57.88 & 58.38\\
ATC$^\text{complete}$ (ours) & \cellcolor{red!25}66.69 & \cellcolor{red!25}71.03 & \cellcolor{red!25}71.87 & \cellcolor{yellow!25}72.15 & & 62.96 & 71.39 & 73.69 & \textbf{74.20} & & \cellcolor{orange!25}65.27 & \cellcolor{orange!25}68.96 & 69.65 & 70.47 & & \cellcolor{red!25}56.71 & 57.28 & 57.61 & \cellcolor{red!25}\textbf{60.22}\\
 \midrule

&&&&&&&&&&&&&&&&&&\\ \midrule

DeiT-S & \multicolumn{4}{c}{79.85} & & \multicolumn{4}{c}{80.57} & & \multicolumn{4}{c}{78.11} & & \multicolumn{4}{c}{63.23} \\ \midrule\midrule
$r$  (\%) & 25 & 50 & 70 & 90 &  & 25 & 50 & 70 & 90 &  & 25 & 50 & 70 & 90 &  & 25 & 50 & 70 & 90 \\ \midrule

$\ell_1$ & 70.05 & 74.47 & \cellcolor{blue!10}77.25 & 79.17 &  & 62.90 & 70.52 & 77.10 & 80.08 &  & 61.28 & 69.49 & 74.31 & 77.09 &  & \cellcolor{blue!10}54.44 & 59.60 & 61.98 & 62.91 \\
$\ell_2$ & 70.54 & 74.86 & 77.41 & 79.27 &  & 64.28 & 72.09 & 77.53 & 80.11 &  & 62.23 & 70.30 & 74.66 & 77.19 &  & 55.31 & 60.22 & 62.07 & 62.71 \\
$\ell_\infty$ & 70.58 & \cellcolor{blue!20}74.03 & 77.48 & 79.23 &  & 63.36 & 70.19 & 77.23 & 79.96 &  & 61.50 & 69.11 & 74.73 & 77.27 &  & 55.10 & 59.34 & 62.11 & 62.77 \\ \midrule
Top-K & 72.91 & 77.82 & 79.22 & \textbf{79.87} &  & \cellcolor{orange!25}76.28 & \cellcolor{orange!25}80.38 & \cellcolor{yellow!25}\textbf{80.70} & \textbf{80.60} &  & 70.14 & \cellcolor{yellow!25}75.84 & \cellcolor{yellow!25}77.50 & \cellcolor{red!25}78.09 &  & 59.32 & 61.98 & \cellcolor{yellow!25}62.69 & \cellcolor{yellow!25}\textbf{63.26} \\
EViT & 74.17 & 78.08 & 79.30 & \textbf{79.87} &  & \cellcolor{red!25}76.74 & \cellcolor{yellow!25}80.28 & \cellcolor{orange!25}\textbf{80.73} & \textbf{80.64} &  & 71.28 & 75.78 & \cellcolor{yellow!25}77.50 & \cellcolor{orange!25}78.07 &  & 59.69 & 61.89 & 62.67 & \textbf{63.25} \\
DynamicViT & \cellcolor{blue!30}60.32 & 77.84 & 79.17 & 79.79 &  & 70.60 & \cellcolor{red!25}\textbf{80.62} & \cellcolor{red!25}\textbf{80.77} & \cellcolor{red!25}\textbf{80.84} &  & \cellcolor{blue!30}39.18 & 69.02 & 75.43 & 77.69 &  & \cellcolor{blue!30}39.20 & \cellcolor{blue!20}57.83 & 61.96 & 63.16 \\
ATS & 72.95 & 77.86 & 79.09 & 79.63 &  & 73.46 & 78.89 & 80.36 & 80.55 &  & 70.13 & 75.66 & 77.23 & 77.83 &  & 60.20 & \cellcolor{orange!25}62.35 & \cellcolor{red!25}62.93 & 63.18 \\ \midrule

ToMe & - & 78.29 & \cellcolor{yellow!25}79.63 & \cellcolor{orange!25}\textbf{79.92} &  & - & 74.99 & 80.05 & \cellcolor{orange!25}\textbf{80.68} &  & - & 74.99 & 77.36 & 77.88 &  & - & 61.51 & 62.50 & 62.89 \\
K-Medoids & 68.94 & 76.44 & 78.74 & 79.73 &  & 65.28 & 76.95 & 79.75 & 80.46 &  & 66.26 & 74.15 & 76.76 & 77.94 &  & 57.78 & 61.48 & 62.47 & 63.12 \\
DPC-KNN & \cellcolor{yellow!25}75.01 & 77.95 & 78.85 & 79.54 &  & 68.77 & 74.14 & 76.70 & 78.88 &  & \cellcolor{yellow!25}72.15 & 75.70 & 77.06 & 77.74 &  & \cellcolor{yellow!25}60.78 & 62.11 & 62.67 & 62.93 \\ \midrule
SiT & 74.65 & 77.16 & 77.52 & \cellcolor{blue!10}77.71 &  & \cellcolor{blue!10}62.82 & \cellcolor{blue!10}62.02 & \cellcolor{blue!20}60.72 & \cellcolor{blue!20}58.50 &  & \cellcolor{blue!10}57.65 & \cellcolor{blue!20}57.33 & \cellcolor{blue!20}57.11 & \cellcolor{blue!20}57.13 &  & 57.95 & \cellcolor{blue!10}58.84 & \cellcolor{blue!20}59.29 & \cellcolor{blue!20}59.59 \\
PatchMerger & \cellcolor{blue!10}69.44 & \cellcolor{blue!10}74.17 & \cellcolor{blue!20}75.80 & \cellcolor{blue!20}76.75 &  & \cellcolor{blue!30}47.26 & \cellcolor{blue!20}61.34 & \cellcolor{blue!10}65.45 & \cellcolor{blue!10}68.24 &  & 62.24 & \cellcolor{blue!10}68.09 & \cellcolor{blue!10}70.75 & \cellcolor{blue!10}72.12 &  & 55.82 & 59.27 & \cellcolor{blue!10}60.46 & \cellcolor{blue!10}61.20 \\
Sinkhorn & \cellcolor{blue!20}64.26 & \cellcolor{blue!30}64.07 & \cellcolor{blue!30}64.02 & \cellcolor{blue!30}64.09 &  & \cellcolor{blue!20}48.89 & \cellcolor{blue!30}50.19 & \cellcolor{blue!30}51.46 & \cellcolor{blue!30}51.22 &  & \cellcolor{blue!20}56.93 & \cellcolor{blue!30}56.68 & \cellcolor{blue!30}56.85 & \cellcolor{blue!30}56.65 &  & \cellcolor{blue!20}50.59 & \cellcolor{blue!30}50.67 & \cellcolor{blue!30}50.63 & \cellcolor{blue!30}50.21 \\ \midrule

ATC$^\text{single}$ (ours) & 73.41 & \cellcolor{yellow!25}78.44 & 79.58 & \cellcolor{yellow!25}\textbf{79.90} & & 72.16 & 79.18 & 80.34 & \cellcolor{orange!25}\textbf{80.68} & & 70.55 & 75.51 & 77.36 & 77.94 & & 59.71 & 61.50 & 62.58 & \cellcolor{red!25}\textbf{63.46}\\
ATC$^\text{average}$ (ours) & \cellcolor{red!25}76.42 & \cellcolor{red!25}79.23 & \cellcolor{red!25}79.82 & \cellcolor{red!25}\textbf{79.95} & & \cellcolor{yellow!25}74.43 & 79.32 & 80.43 & \textbf{80.64} & & \cellcolor{orange!25}72.55 & \cellcolor{orange!25}76.47 & \cellcolor{orange!25}77.51 & 77.95 & & \cellcolor{orange!25}60.90 & \cellcolor{yellow!25}62.13 & \cellcolor{red!25}62.93 & 63.14\\
ATC$^\text{complete}$ (ours) & \cellcolor{orange!25}76.34 & \cellcolor{orange!25}79.20 & \cellcolor{orange!25}79.77 & \cellcolor{yellow!25}\textbf{79.90} & & 73.67 & 79.27 & 80.37 & \cellcolor{yellow!25}\textbf{80.66} & & \cellcolor{red!25}72.65 & \cellcolor{red!25}76.60 & \cellcolor{red!25}77.52 & \cellcolor{yellow!25}77.98 & & \cellcolor{red!25}60.96 & \cellcolor{red!25}62.47 & \cellcolor{orange!25}62.85 & \cellcolor{orange!25}\textbf{63.44}\\

\midrule

&&&&&&&&&&&&&&&&&&\\ \midrule

DeiT-B & \multicolumn{4}{c}{81.85} &  & \multicolumn{4}{c}{83.32} &  & \multicolumn{4}{c}{80.93} &  & \multicolumn{4}{c}{64.37} \\ \midrule\midrule
$r$  (\%) & 25 & 50 & 70 & 90 &  & 25 & 50 & 70 & 90 &  & 25 & 50 & 70 & 90 &  & 25 & 50 & 70 & 90 \\ \midrule

$\ell_1$ & 71.23 & 74.96 & 78.94 & 81.04 &  & 59.79 & 71.57 & 78.92 & 82.42 &  & 58.28 & 69.27 & 76.23 & 79.65 &  & 53.01 & 60.10 & 63.25 & 64.14 \\
$\ell_2$ & 71.41 & 75.40 & 79.07 & 81.18 &  & 61.55 & 73.24 & 79.52 & 82.55 &  & 59.69 & 70.33 & 76.56 & 79.75 &  & 54.00 & 60.37 & 63.29 & 64.28 \\
$\ell_\infty$ & 71.67 & \cellcolor{blue!10}74.40 & 78.95 & 81.20 &  & 59.96 & 70.51 & 79.73 & 82.59 &  & 58.48 & 68.50 & 76.54 & 79.89 &  & 53.00 & 59.59 & 63.12 & 64.25 \\ \midrule
Top-K & 73.63 & 78.97 & 80.91 & \cellcolor{orange!25}\textbf{82.03} &  & \cellcolor{orange!25}74.71 & \cellcolor{orange!25}82.22 & \cellcolor{red!25}83.20 & \cellcolor{red!25}\textbf{83.40} &  & 67.63 & \cellcolor{orange!25}76.91 & \cellcolor{red!25}79.95 & \cellcolor{red!25}\textbf{80.97} &  & 58.51 & 62.78 & 63.92 &\textbf{64.40} \\
EViT & \cellcolor{red!25}75.26 & \cellcolor{yellow!25}79.22 & 80.99 & \cellcolor{yellow!25}\textbf{82.00} &  & \cellcolor{red!25}74.73 & \cellcolor{yellow!25}82.00 & \cellcolor{orange!25}83.19 & \cellcolor{orange!25}\textbf{83.33} &  & \cellcolor{red!25}68.93 & \cellcolor{red!25}76.92 & \cellcolor{orange!25}79.87 & \cellcolor{orange!25}80.92 &  & 59.00 & 62.88 & 63.90 & \textbf{64.43} \\
DynamicViT & \cellcolor{blue!30}27.94 & 74.58 & 80.68 & 81.76 &  & \cellcolor{blue!10}49.23 & \cellcolor{red!25}82.30 & \cellcolor{yellow!25}83.16 & 83.23 &  & \cellcolor{blue!30}24.88 & \cellcolor{blue!10}62.79 & 76.54 & 80.64 &  & \cellcolor{blue!30}28.56 & \cellcolor{blue!20}55.51 & \cellcolor{blue!10}60.73 & 63.83 \\
ATS & \cellcolor{orange!25}73.89 & 78.94 & 80.78 & 81.57 &  & \cellcolor{yellow!25}71.00 & 80.10 & 82.58 & \cellcolor{yellow!25}83.26 &  & 68.17 & \cellcolor{yellow!25}76.38 & 79.35 & 80.50 &  & \cellcolor{yellow!25}59.49 & \cellcolor{red!25}63.17 & \cellcolor{red!25}64.21 & \cellcolor{yellow!25}\textbf{64.48} \\ \midrule

ToMe & - & 78.89 & \cellcolor{yellow!25}81.05 & \cellcolor{yellow!25}\textbf{82.00} &  & - & 73.67 & 81.59 & 82.98 &  & - & 74.11 & 78.82 & 80.48 &  & - & 62.38 & \cellcolor{orange!25}64.06 & 64.35 \\
K-Medoids & 69.12 & 76.86 & 79.98 & 81.76 &  & 57.54 & 75.29 & 80.62 & 82.57 &  & 61.79 & 73.60 & 77.58 & 80.32 &  & 56.67 & 62.18 & 63.53 & 64.35 \\
DPC-KNN & 69.40 & 75.87 & 79.06 & 81.05 &  & 58.16 & \cellcolor{blue!10}67.36 & 72.83 & 78.29 &  & 65.99 & 73.32 & 77.03 & 79.76 &  & 58.58 & 61.39 & 62.96 & 63.87 \\ \midrule
SiT & 68.39 & 75.53 & \cellcolor{blue!10}76.63 & \cellcolor{blue!10}77.26 &  & 65.09 & 70.75 & \cellcolor{blue!10}70.36 & \cellcolor{blue!10}68.96 &  & 54.86 & \cellcolor{blue!20}53.27 & \cellcolor{blue!20}53.16 & \cellcolor{blue!20}52.73 &  & 56.12 & 59.76 & \cellcolor{blue!20}60.64 & \cellcolor{blue!20}61.08 \\
PatchMerger & \cellcolor{blue!20}58.78 & \cellcolor{blue!20}70.63 & \cellcolor{blue!20}74.52 & \cellcolor{blue!20}76.76 &  & \cellcolor{blue!30}40.38 & \cellcolor{blue!20}57.21 & \cellcolor{blue!20}62.20 & \cellcolor{blue!20}67.06 &  & \cellcolor{blue!10}54.25 & 66.22 & \cellcolor{blue!10}70.97 & \cellcolor{blue!10}73.72 &  & \cellcolor{blue!10}51.80 & \cellcolor{blue!10}58.83 & 60.79 & \cellcolor{blue!10}62.09 \\
Sinkhorn & \cellcolor{blue!10}63.37 & \cellcolor{blue!30}63.33 &\cellcolor{blue!30}63.36 & \cellcolor{blue!30}63.50 &  & \cellcolor{blue!20}42.89 &\cellcolor{blue!30}42.33 & \cellcolor{blue!30}41.72 & \cellcolor{blue!30}42.86 &  & \cellcolor{blue!20}52.57 & \cellcolor{blue!30}52.33 & \cellcolor{blue!30}52.21 & \cellcolor{blue!30}52.12 &  & \cellcolor{blue!20}47.55 & \cellcolor{blue!30}47.41 & \cellcolor{blue!30}47.26 & \cellcolor{blue!30}47.48 \\ \midrule
ATC$^\text{single}$ (ours) & 70.00 & 79.10 & 81.01 & \cellcolor{red!25}\textbf{82.04} & & 64.14 & 78.91 & 81.79 & 83.12 & & 66.76 & 75.49 & \cellcolor{yellow!25}79.69 & \cellcolor{yellow!25}80.70 & & 58.79 & 61.51 & 63.93 & \cellcolor{red!25}\textbf{64.82}\\
ATC$^\text{average}$ (ours) & \cellcolor{yellow!25}73.81 & \cellcolor{red!25}79.79 & \cellcolor{orange!25}81.10 & 81.76 & & 67.55 & 79.07 & 81.88 & 83.10 & & \cellcolor{orange!25}68.61 & 75.87 & 79.57 & 80.65 & & \cellcolor{red!25}59.74 & \cellcolor{orange!25}63.02 & \cellcolor{yellow!25}64.02 & \textbf{64.42}\\
ATC$^\text{complete}$ (ours) & 73.80 & \cellcolor{orange!25}79.75 & \cellcolor{red!25}81.15 & \cellcolor{orange!25}\textbf{82.03} & & 66.14 & 78.44 & 81.67 & 83.08 & & \cellcolor{yellow!25}68.46 & 76.05 & 79.40 & 80.67 & & \cellcolor{orange!25}59.64 & \cellcolor{yellow!25}62.91 & 63.95 & \cellcolor{orange!25}\textbf{64.77}\\ \bottomrule

\end{tabular}%
}
\end{table*}

\begin{table*}[!tp]
\caption{\textbf{Off-the-Shelf ATC Performance with DeiT-Tiny, DeiT-Small, and DeiT-Base backbones.} We report the off-the-shelf, \ie without fine-tuning, performance of the proposed ATC method. Model performance is measured across keep rates, $r$, denoted in percentage of tokens kept at each reduction stage, and with the DeiT-Tiny, DeiT-Small, and DeiT-Base models. ImageNet and NABirds performance is measured with top-1 accuracy, whereas COCO and NUS-WIDE is measured with mAP. The best performing method per column is denoted in \textbf{bold}. A \scolorbox{dt2}{blue} background indicates that ATC was applied off-the-shelf on the DeiT backbone.}
\label{tab:perfICCVPartialOTS}

\begin{subtable}[t]{\linewidth}
\caption{\textbf{DeiT-T Backbone}}
\label{tab:DeiTTICCVOTS}
\centering

\resizebox{\linewidth}{!}{%
\begin{tabular}{lcccclcccclcccclcccc}
\toprule
 & \multicolumn{4}{c}{ImageNet} &  & \multicolumn{4}{c}{NABirds} &  & \multicolumn{4}{c}{COCO} &  & \multicolumn{4}{c}{NUS-WIDE} \\ \cmidrule(lr){2-5} \cmidrule(lr){7-10} \cmidrule(lr){12-15} \cmidrule(lr){17-20}  
 
DeiT-T & \multicolumn{4}{c}{72.20} & & \multicolumn{4}{c}{74.16} & & \multicolumn{4}{c}{71.09} & & \multicolumn{4}{c}{59.27} \\ \midrule\midrule
$r$  (\%) & 25 & 50 & 70 & 90 &  & 25 & 50 & 70 & 90 &  & 25 & 50 & 70 & 90 &  & 25 & 50 & 70 & 90 \\\midrule

\otsmodel{ATC$^{\text{single}}$ (ours)} & \otsmodel{22.97} & \otsmodel{58.62} & \otsmodel{67.86} & \otsmodel{71.38} & \otsmodel{}& \otsmodel{27.77} & \otsmodel{64.74} & \otsmodel{70.88} & \otsmodel{73.38} & \otsmodel{}& \otsmodel{33.97} & \otsmodel{62.14} & \otsmodel{68.86} & \otsmodel{70.89} & \otsmodel{}& \otsmodel{34.66} & \otsmodel{52.92} & \otsmodel{57.49} & \otsmodel{59.02}\\
\otsmodel{ATC$^{\text{average}}$ (ours)} & \otsmodel{15.67} & \otsmodel{57.55} & \otsmodel{68.27} & \otsmodel{71.63} & \otsmodel{}& \otsmodel{28.08} & \otsmodel{64.30} & \otsmodel{71.16} & \otsmodel{73.50} & \otsmodel{}& \otsmodel{24.21} & \otsmodel{61.72} & \otsmodel{68.95} & \otsmodel{70.79} & \otsmodel{}& \otsmodel{29.66} & \otsmodel{53.02} & \otsmodel{57.59} & \otsmodel{59.00}\\
\otsmodel{ATC$^{\text{complete}}$ (ours)} & \otsmodel{12.15} & \otsmodel{57.17} & \otsmodel{68.46} & \otsmodel{71.59} & \otsmodel{}& \otsmodel{23.30} & \otsmodel{63.59} & \otsmodel{71.30} & \otsmodel{73.72} & \otsmodel{}& \otsmodel{18.92} & \otsmodel{60.71} & \otsmodel{68.82} & \otsmodel{70.74} & \otsmodel{}& \otsmodel{26.66} & \otsmodel{52.82} & \otsmodel{57.71} & \otsmodel{59.04}\\  \bottomrule

\end{tabular}%
}
\end{subtable}
\vspace{0.25mm}

\begin{subtable}[t]{\linewidth}
\caption{\textbf{DeiT-S Backbone}}
\label{tab:DeiTSICCVOTS}
\centering

\resizebox{\linewidth}{!}{%
\begin{tabular}{lcccclcccclcccclcccc}
\toprule
 & \multicolumn{4}{c}{ImageNet} &  & \multicolumn{4}{c}{NABirds} &  & \multicolumn{4}{c}{COCO} &  & \multicolumn{4}{c}{NUS-WIDE} \\ \cmidrule(lr){2-5} \cmidrule(lr){7-10} \cmidrule(lr){12-15} \cmidrule(lr){17-20}  
DeiT-S & \multicolumn{4}{c}{79.85} & & \multicolumn{4}{c}{80.57} & & \multicolumn{4}{c}{78.11} & & \multicolumn{4}{c}{63.23} \\ \midrule\midrule
$r$  (\%) & 25 & 50 & 70 & 90 &  & 25 & 50 & 70 & 90 &  & 25 & 50 & 70 & 90 &  & 25 & 50 & 70 & 90 \\\midrule
\otsmodel{ATC$^{\text{single}}$ (ours)} & \otsmodel{26.40} & \otsmodel{68.85} & \otsmodel{76.89} & \otsmodel{79.32} & \otsmodel{}& \otsmodel{38.32} & \otsmodel{72.87} & \otsmodel{78.30} & \otsmodel{80.21} & \otsmodel{}& \otsmodel{34.60} & \otsmodel{68.46} & \otsmodel{75.58} & \otsmodel{77.76} & \otsmodel{}& \otsmodel{31.15} & \otsmodel{56.11} & \otsmodel{61.59} & \otsmodel{63.08}\\
\otsmodel{ATC$^{\text{average}}$ (ours)} & \otsmodel{24.27} & \otsmodel{69.33} & \otsmodel{77.35} & \otsmodel{79.47} & \otsmodel{}& \otsmodel{38.79} & \otsmodel{72.73} & \otsmodel{78.46} & \otsmodel{80.38} & \otsmodel{}& \otsmodel{27.37} & \otsmodel{68.08} & \otsmodel{75.68} & \otsmodel{77.80} & \otsmodel{}& \otsmodel{33.41} & \otsmodel{57.90} & \otsmodel{62.12} & \otsmodel{63.07}\\
\otsmodel{ATC$^{\text{complete}}$ (ours)} & \otsmodel{20.12} & \otsmodel{69.44} & \otsmodel{77.38} & \otsmodel{79.47} & \otsmodel{}& \otsmodel{31.94} & \otsmodel{72.24} & \otsmodel{78.38} & \otsmodel{80.39} & \otsmodel{}& \otsmodel{22.12} & \otsmodel{67.09} & \otsmodel{75.63} & \otsmodel{77.81} & \otsmodel{}& \otsmodel{31.71} & \otsmodel{57.98} & \otsmodel{62.20} & \otsmodel{63.11}\\ \bottomrule

\end{tabular}%
}
\end{subtable}
\vspace{0.25mm}

\begin{subtable}[t]{\linewidth}
\caption{\textbf{DeiT-B Backbone}}
\label{tab:DeiTBICCVOTS}
\centering

\resizebox{\linewidth}{!}{%
\begin{tabular}{lcccclcccclcccclcccc}
\toprule
 & \multicolumn{4}{c}{ImageNet} &  & \multicolumn{4}{c}{NABirds} &  & \multicolumn{4}{c}{COCO} &  & \multicolumn{4}{c}{NUS-WIDE} \\ \cmidrule(lr){2-5} \cmidrule(lr){7-10} \cmidrule(lr){12-15} \cmidrule(lr){17-20}  
DeiT-B & \multicolumn{4}{c}{81.85} &  & \multicolumn{4}{c}{83.32} &  & \multicolumn{4}{c}{80.93} &  & \multicolumn{4}{c}{64.37} \\ \midrule\midrule
$r$  (\%) & 25 & 50 & 70 & 90 &  & 25 & 50 & 70 & 90 &  & 25 & 50 & 70 & 90 &  & 25 & 50 & 70 & 90 \\ \midrule

\otsmodel{ATC$^{\text{single}}$ (ours)} & \otsmodel{18.99} & \otsmodel{71.30} & \otsmodel{79.19} & \otsmodel{81.39} & \otsmodel{}& \otsmodel{26.23} & \otsmodel{73.56} & \otsmodel{80.32} & \otsmodel{82.64} & \otsmodel{}& \otsmodel{19.15} & \otsmodel{67.08} & \otsmodel{77.69} & \otsmodel{80.58} & \otsmodel{}& \otsmodel{26.27} & \otsmodel{57.22} & \otsmodel{62.63} & \otsmodel{64.16}\\
\otsmodel{ATC$^{\text{average}}$ (ours)} & \otsmodel{13.39} & \otsmodel{70.37} & \otsmodel{79.29} & \otsmodel{81.45} & \otsmodel{}& \otsmodel{25.56} & \otsmodel{73.22} & \otsmodel{80.52} & \otsmodel{82.63} & \otsmodel{}& \otsmodel{14.65} & \otsmodel{66.65} & \otsmodel{77.77} & \otsmodel{80.57} & \otsmodel{}& \otsmodel{23.66} & \otsmodel{57.68} & \otsmodel{62.78} & \otsmodel{64.12}\\
\otsmodel{ATC$^{\text{complete}}$ (ours)} & \otsmodel{11.41} & \otsmodel{69.65} & \otsmodel{79.33} & \otsmodel{81.43} & \otsmodel{}& \otsmodel{22.55} & \otsmodel{72.48} & \otsmodel{80.58} & \otsmodel{82.71} & \otsmodel{}& \otsmodel{12.85} & \otsmodel{65.53} & \otsmodel{77.61} & \otsmodel{80.55} & \otsmodel{}& \otsmodel{22.07} & \otsmodel{57.58} & \otsmodel{62.77} & \otsmodel{64.14}\\ \bottomrule

\end{tabular}%
}
\end{subtable}
\end{table*}

\section{Off-the-Shelf Performance of ATC on Image Classification Datasets}\label{sec:methodCompOTS}

As a supplement to the fine-tuning based results in Section~\ref{sec:methodComp}, we also present the ATC results when applied off-the-shelf, \ie without any fine-tuning, see Table~\ref{tab:perfICCVPartialOTS}. We find that for all datasets the off-the-shelf models achieve good performance when applied at high keep rates of 70\% and 90\%, even matching and at time outperforming prior fine-tuned hard-merging based methods. However, as the keep rate is lowered to 50\% and 25\% the performance drops significantly, introducing a gap between off-the-shelf ATC and prior fine-tuned hard-merging based methods. However, we observe that for $r=50\%$ off-the-shelf ATC still outperforms the fine-tuned soft-merging based methods.

\section{Off-the-Shelf Performance with AugReg, SWAG, and MAE weights}\label{sec:weights}
\begin{figure*}[!tp]
    \centering
    \includegraphics[width=\linewidth]{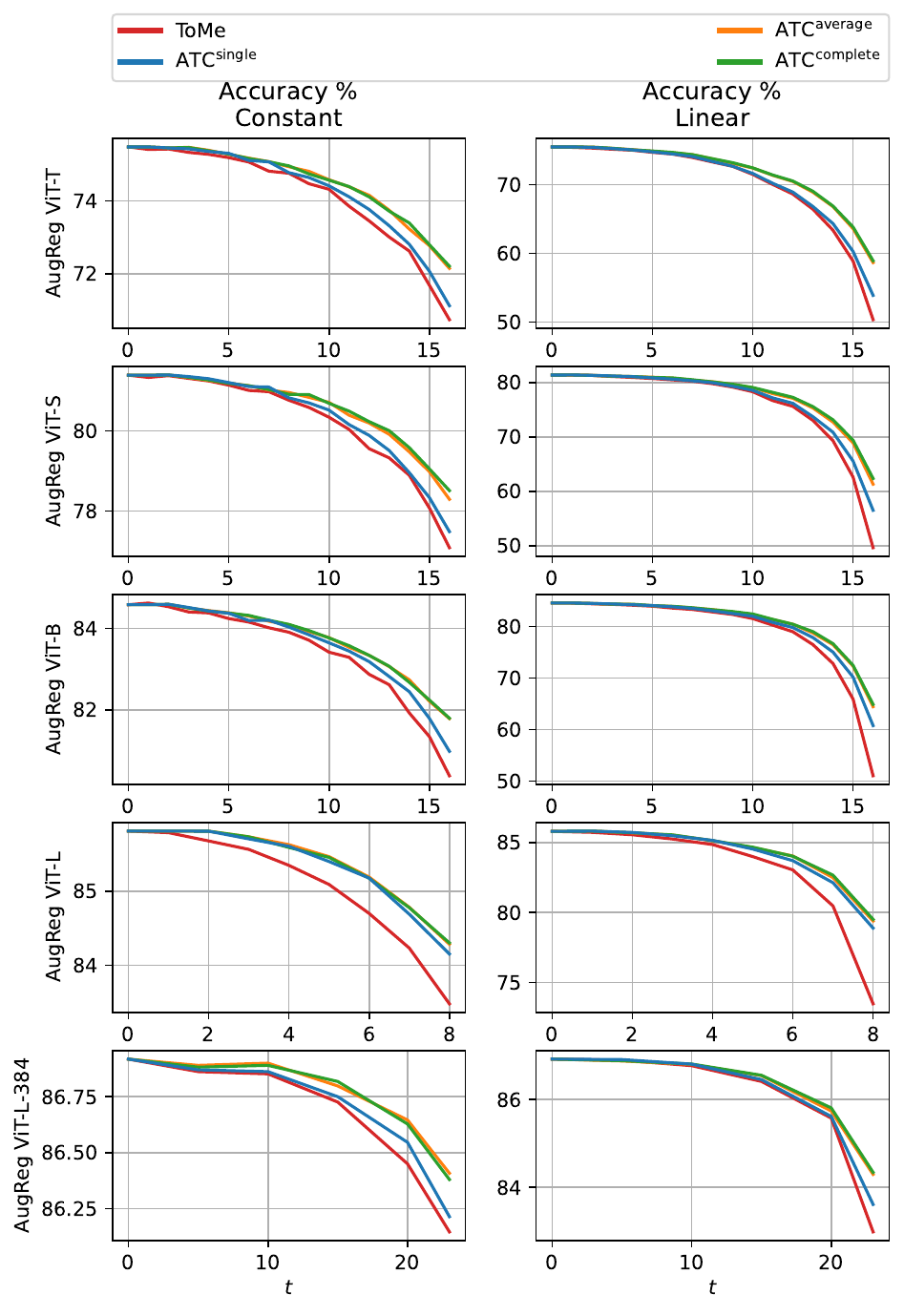}
    \caption{\textbf{Off-the-Shelf Results -- AugReg.}}
    \label{fig:ots_augreg}
\end{figure*}

\begin{figure*}[!tp]
    \centering\includegraphics[width=\linewidth]{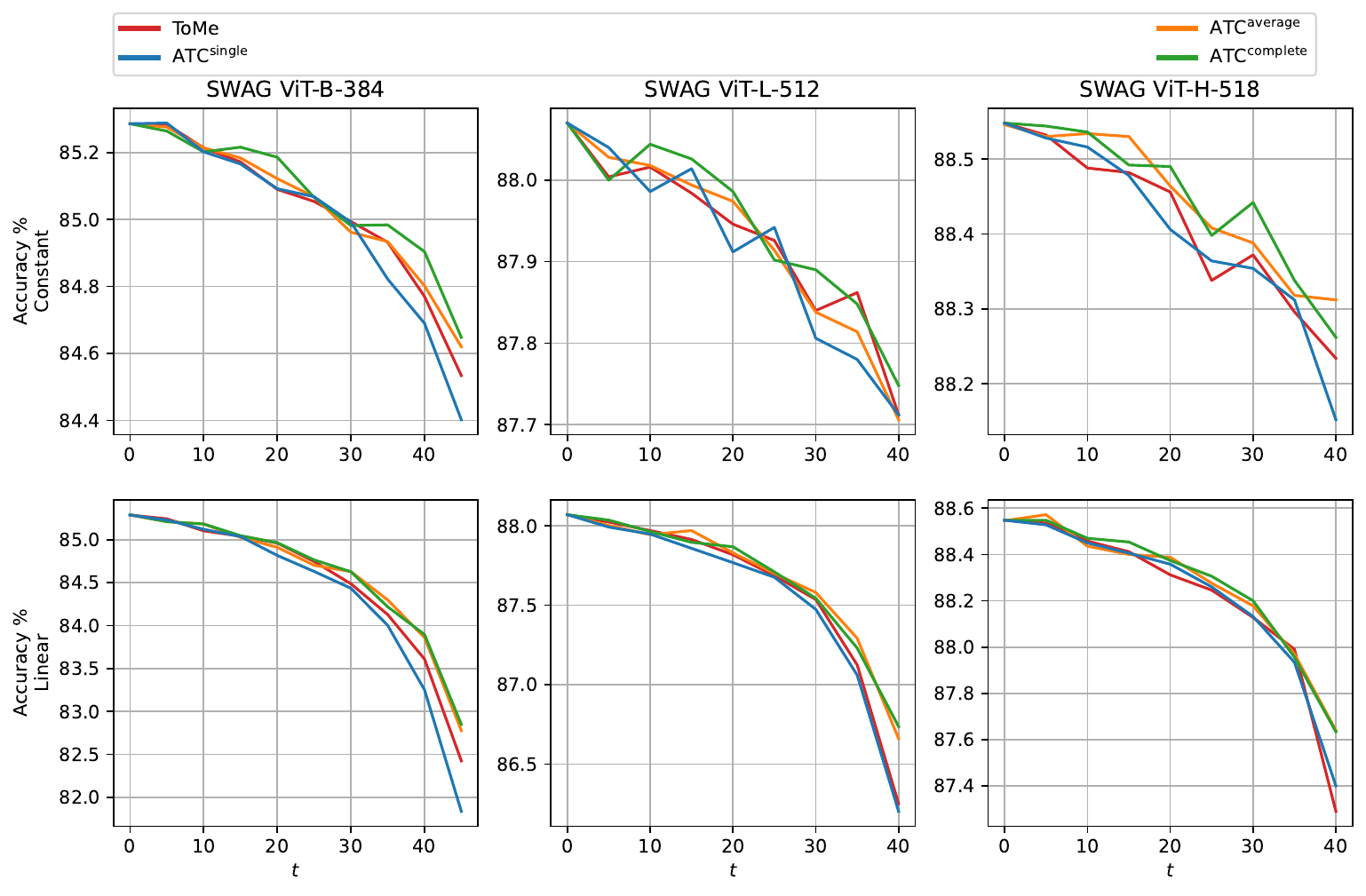}
    \caption{Off-the-Shelf Results -- SWAG.}
    \label{fig:ots_swag}
\end{figure*}

\begin{figure*}[!tp]
    \centering\includegraphics[width=\linewidth]{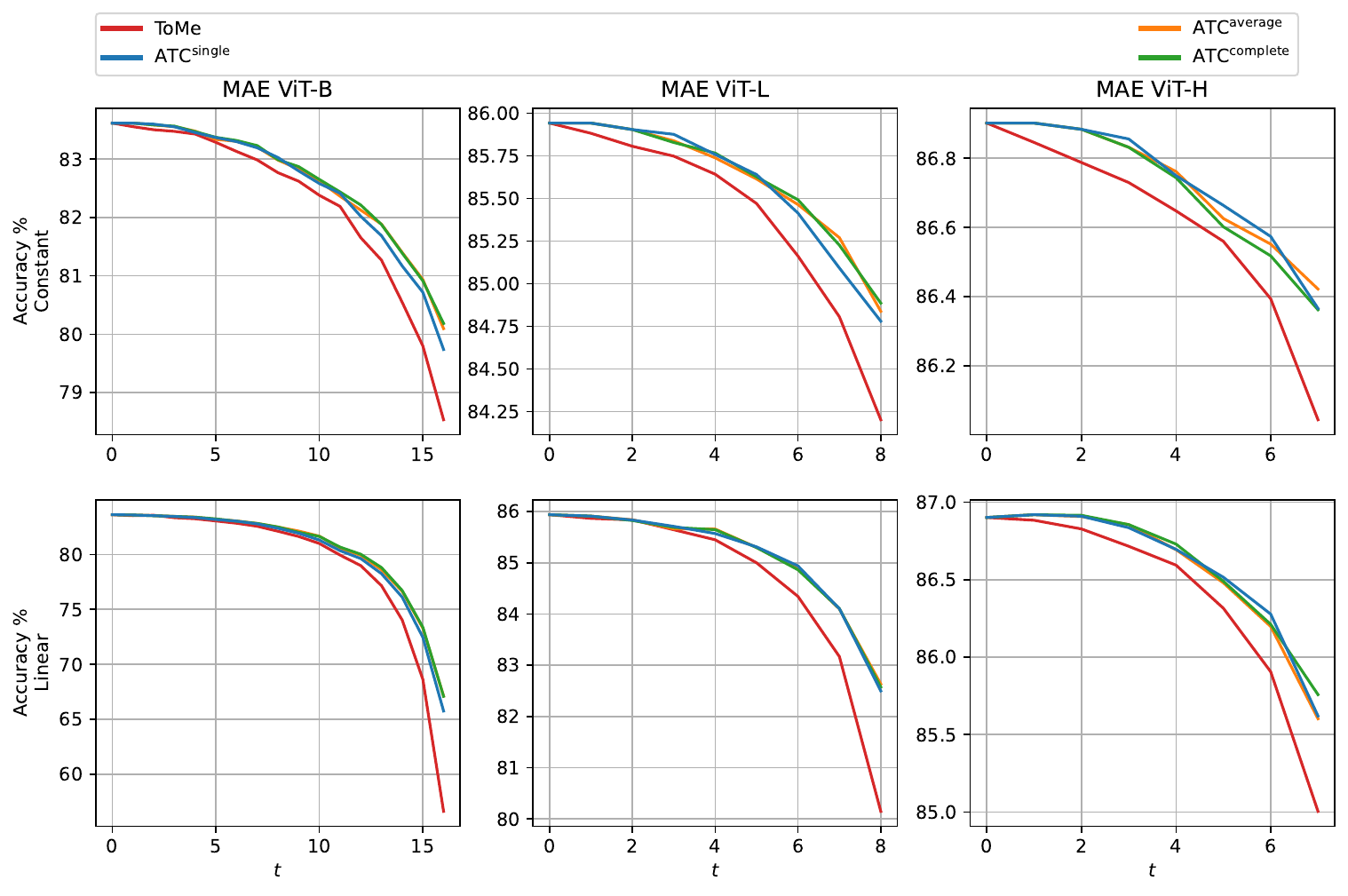}
    \caption{Off-the-Shelf Results -- MAE.}
    \label{fig:ots_mae}
\end{figure*}

Inspired by the work of Bolya \etal \cite{ToMe_2023}, we investigate the performance of ATC when applied off-the-shelf, \ie without fine-tuning, to different pre-trained Vision Transformers. Specifically, we consider models trained using the AugReg \cite{AugReg}, SWAG \cite{SWAG}, and MAE \cite{MAE} protocols. We match the evaluation setting of Bolya \etal, and vary the number of tokens removed in each ViT block, denoted $t$, using both the constant and linear token schedules. We find that as the model capacity is increased, the effect of token reduction is reduced significantly, see Figures~\ref{fig:ots_augreg}--\ref{fig:ots_mae}. We also find that in general ATC$^\text{average}$ and ATC$^\text{complete}$ outperform ATC$^\text{single}$ and ToMe when using AugReg and MAE weights. When using the constant schedule with the SWAG weights, performance is very similar across all methods.

\section{Performance of Token Reduction Methods on Object Detection and Segmentation}\label{sec:methodCompObj}

We extend our method comparison and analysis on the object detection \& segmentation task, by also comparing withteh pruning-based results from Liu \etal \cite{SViT}, see Table~\ref{tab:objdetsegFull}. It is worth noting that the pruning-based methods were only trained with a keep rate of $r=70\%$. We find that our proposed ATC model not only outperforms the prior best merging-based method, ToMe \cite{ToMe_2023}, but also outperforms all prior pruning-based methods, including the SViT \cite{SViT} method. This is especially observable with a ViT-Adapter-Tiny backbone, where the mAP$^\text{box}$ results is improve by 0.4 percentage points and mAP$^\text{mask}$ by 1.6 percentage points. We also find that the SViT performance can be matched with ATC at a keep rate of 50\%. These results demonstrate the clear benefit of using ATC over all prior pruning-based methods for the object detection and segmentation task.

\begin{table}[!t]
    \caption{\textbf{Object Detection \& Segmentation Results.} We compare the performance of ATC and ToMe on the dense object detection \& segmentation task using the COCO 2017 dataset, following Liu \etal \cite{SViT}. The pruning-based token reduction results are replicated from Table 5 in \cite{SViT}. The ViT-Adapter method is used, with DeiT-T and DeiT-S backbones. The best performing method per column is denoted in \textbf{bold}.  
    D denotes the model diverged during training. Note that ToMe is limited to $r\geq 50\%$.}
    \label{tab:objdetsegFull}
    \begin{subtable}[t]{0.5\linewidth}
    \caption{\textbf{ViT-Adapter-T Backbone}}
    \label{tab:vitadaptTSupp}
    \centering
    
    \resizebox{\linewidth}{!}{%
    \begin{tabular}{lccrcccrc}
    
    \toprule
     & \multicolumn{4}{c}{mAP$^{\text{box}}$}  & \multicolumn{4}{c}{mAP$^{\text{mask}}$}  \\ \cmidrule(lr){2-5} \cmidrule(lr){6-9}  
    ViT-Adapter-T & \multicolumn{4}{c}{45.8} & \multicolumn{4}{c}{40.9} \\ \midrule\midrule
    $r$  (\%) & 25 & 50 & \multicolumn{1}{c}{70} & 90 & 25 & 50 & \multicolumn{1}{c}{70} & 90 \\ \midrule
    EViT & - & - & 44.5 & - & - & - & 39.8 & - \\
    EvoViT & - & - & 44.8 & - & - & - & 39.9 & - \\
    ATS & - & - & 43.9 & - & - & - & 39.1 & - \\
    DynamicViT & - & - & 41.2 & - & - & - & 37.1 & - \\
    DynamicViT+prsv & - & - & 44.1 & - & - & - & 39.5 & - \\
    SViT & - & - & 45.5 & - & - & - & 39.4 & - \\ \midrule
ToMe& -& 43.7& 45.7& \textbf{46.0}& -& 39.3& 40.9& 41.0  \\ 
ATC$^{\text{single}}$ (ours)& 36.9& 43.5& 45.5& 45.9& 33.6& 39.1& 40.7& 40.9  \\ 
ATC$^{\text{average}}$ (ours)& 42.4& 45.2& 45.8& 45.9& 38.5& \textbf{40.5} & 40.8& \textbf{41.1}  \\ 
ATC$^{\text{complete}}$ (ours)& \textbf{42.6}& \textbf{45.3}& \textbf{45.9}& \textbf{46.0}& \textbf{38.7}& \textbf{40.5}& \textbf{41.0}& \textbf{41.1} \\
\bottomrule
    \end{tabular}
        }
    \end{subtable} \hfill    
    \begin{subtable}[t]{0.5\linewidth}
    \caption{\textbf{ViT-Adapter-S Backbone}}
    \label{tab:vitadaptSSupp}
    \centering
    
    \resizebox{\linewidth}{!}{%
    \begin{tabular}{lccrcccrc}
    \toprule
     & \multicolumn{4}{c}{mAP$^{\text{box}}$}  & \multicolumn{4}{c}{mAP$^{\text{mask}}$}  \\ \cmidrule(lr){2-5} \cmidrule(lr){6-9}  
    ViT-Adapter-S & \multicolumn{4}{c}{48.5} & \multicolumn{4}{c}{42.8} \\ \midrule\midrule
    $r$  (\%) & 25 & 50 & \multicolumn{1}{c}{70} & 90 & 25 & 50 & \multicolumn{1}{c}{70} & 90 \\ \midrule
    EViT & - & - & 47.1 & - & - & - & 41.6 & - \\
    EvoViT & - & - & 47.2 & - & - & - & 41.6 & - \\
    ATS & - & - & 46.7 & - & - & - & 41.1 & - \\
    DynamicViT & - & - & D & - & - & - & D & - \\
    DynamicViT+prsv & - & - & 47.2 & - & - & - & 41.6 & - \\
    SViT & - & - & 48.2 & - & - & - & 42.5 & -   \\ \midrule
ToMe& -& 46.4& 48.2& 48.3& -& 41.4& \textbf{42.6}& 42.7  \\ 
ATC$^{\text{single}}$ (ours)& 40.3& 46.4& 47.9& 48.3& 36.2& 41.1& 42.3& 42.6  \\ 
ATC$^{\text{average}}$ (ours)& 45.1& \textbf{47.8}& \textbf{48.3}& 48.5& 40.2& \textbf{42.4}& \textbf{42.6}& \textbf{42.9}  \\ 
ATC$^{\text{complete}}$ (ours)& \textbf{45.3}& \textbf{47.8}& 48.2& \textbf{48.6}& \textbf{40.5}& 42.3& \textbf{42.6}& 42.7 \\ 
\bottomrule
    \end{tabular}
    }
    \end{subtable}
\end{table}

\begin{figure*}[!tp]
\centering
\begin{subfigure}[b]{\textwidth}
   \includegraphics[width=1\linewidth]{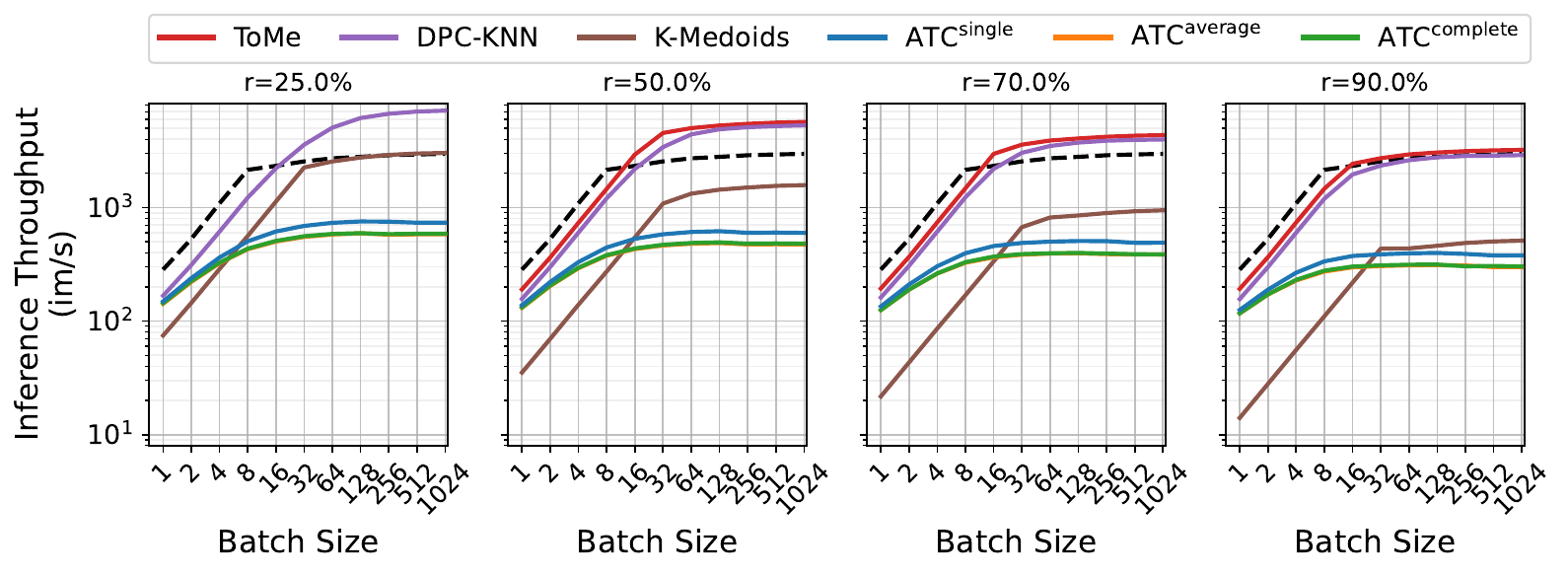}
   \caption{\textbf{Inference Throughput with DeiT-T backbone.}}
   \label{fig:infDeitTiny} 
\end{subfigure}

\begin{subfigure}[b]{\textwidth}
   \includegraphics[width=1\linewidth]{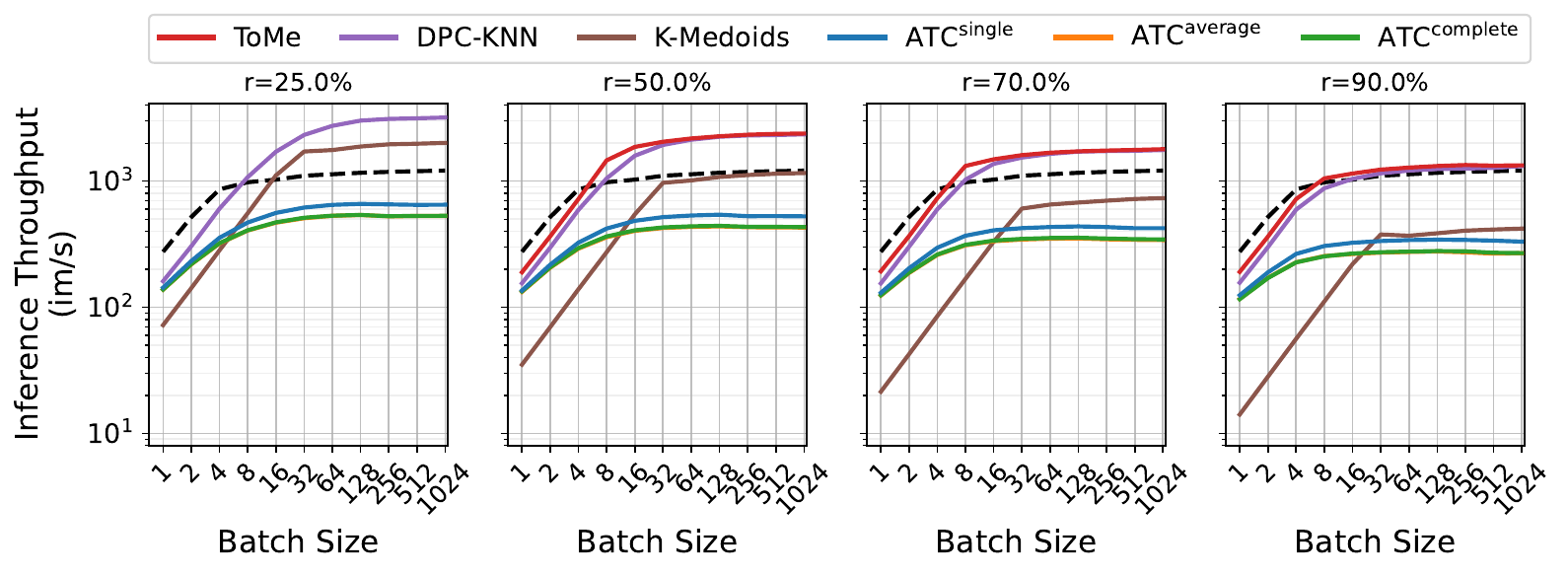}
   \caption{\textbf{Inference Throughput with DeiT-S backbone.}}
   \label{fig:infDeitSmall} 
\end{subfigure}
\begin{subfigure}[b]{\textwidth}
   \includegraphics[width=1\linewidth]{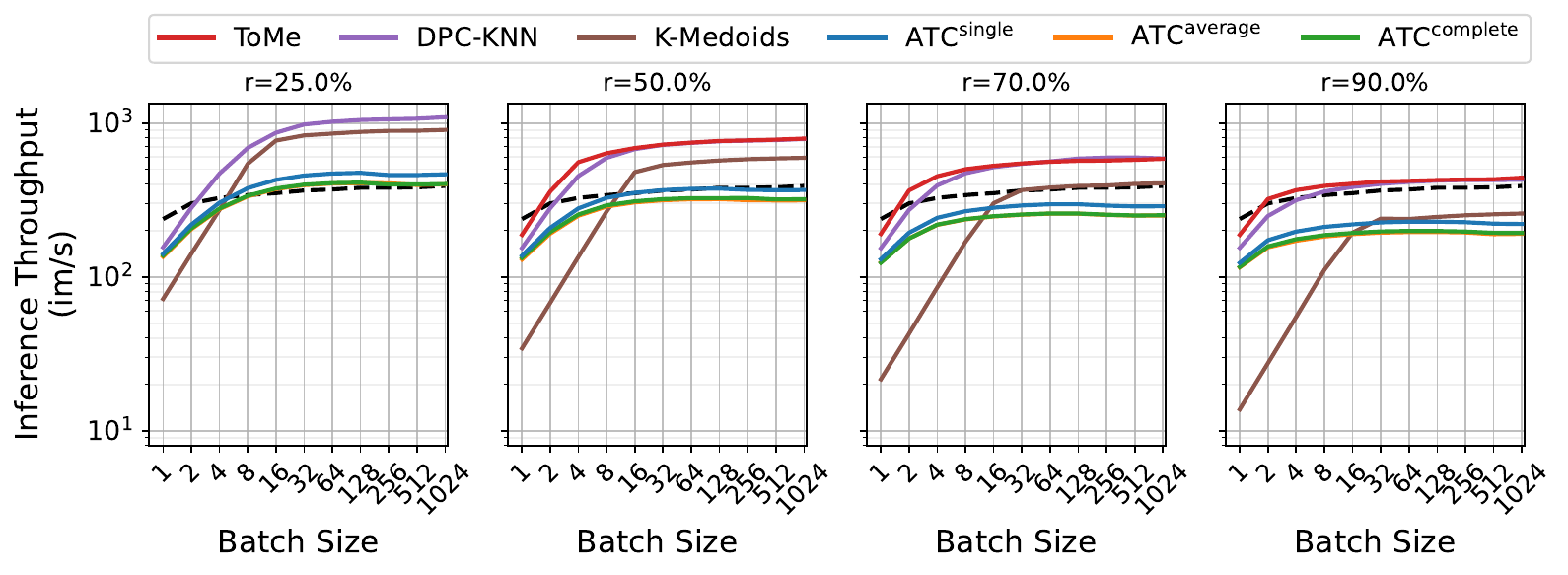}
   \caption{\textbf{Inference Throughput with DeiT-B backbone.}}
   \label{fig:infDeitBase}
\end{subfigure}
\caption{\textbf{Hard Token Merging Inference Throughput Comparison.} We compare the three hard-merging based token reduction methods from Haurum \etal as well as the three versions of ATC. The dashed black line denotes the corresponding DeiT baseline inference throughput. We find that as the batch size is increased, the ATC method becomes dramatically slower than the other hard-merging based methods. In comparison, when using smaller batch sizes, the ATC method achieves similar throughput rates as ToMe and DPC-KNN, and outperforms K-Medoids significantly. This is due to the implementation being CPU-bounded and non-batched. Note that the ATC$^{\text{average}}$ and ATC$^{\text{complete}}$ results overlap each other, and that ToMe does not work with $r=25\%$.}
\label{fig:backboneInference}
\end{figure*}

\section{Inference Throughput Comparisons}\label{sec:inference}

A key motivation for token reduction methods has been to improve the throughput of the model, while limiting the potential decrease in task performance. Therefore, we evaluate the inference throughput of ATC across the three considered linkage functions, as well as comparison to the ToMe, DPC-KNN, and K-Medoids hard-merging methods tested by Haurum \etal \cite{Haurum_2023_ICCV}. We evaluate the throughput on a computer consisting of an A40 GPU with 48GB VRAM, 12 Intel Xeon CPUs, CUDA 11.7 and PyTorch 2.0. We measure the average inference throughput as images per second (im/s) over 1000 batches, of which the first 25\% are used to warm-up the GPU, use 32-bit precision, and vary the batch size in steps of $2^k$, where $k\in\{0,1,2,3,4,5,6,7,8,9,10\}$.

\subsection{Backbone Comparison}
The results of the throughput comparison using the DeiT-T, Deit-S, and DeiT-B backbones can be found in Figure~\ref{fig:backboneInference}. 
We observe that the throughput of ATC is lower than the other methods when the batch sizes are increased, while at smaller batch sizes the throughput of ATC is comparable to ToMe and DPC-KNN. We also find that when using the DeiT-T backbone, only ToMe and DPC-KNN consistently outperform the baseline inference performance. In contrast, all token hard-merging methods outperform the baseline DeiT-B model when $r=25\%$, and becomes slightly slower when the keep rate is increased. 

The reason ATC has a lower throughput than other methods is due to the use of \textsc{scikit-learn}'s implementation of agglomerative clustering, which is CPU-bound and non-batched \cite{scikit-learn}. By investigating the underlying code, it appears that the average and complete linkage functions are built on a naive implementation with time complexity $\mathcal{O}(n^3)$, whereas the single linkage function use minimum spanning trees and has a time complexity  of $\mathcal{O}(n^2)$ \cite{minSpanTreeSLINK}. To see the effect of this we compare with two other implementations (\textsc{SciPy} \cite{SciPy} and \textsc{RAPIDS} \cite{RAPIDS}).

\begin{figure}[!tp]
    \centering
    \includegraphics[width=0.7\linewidth]{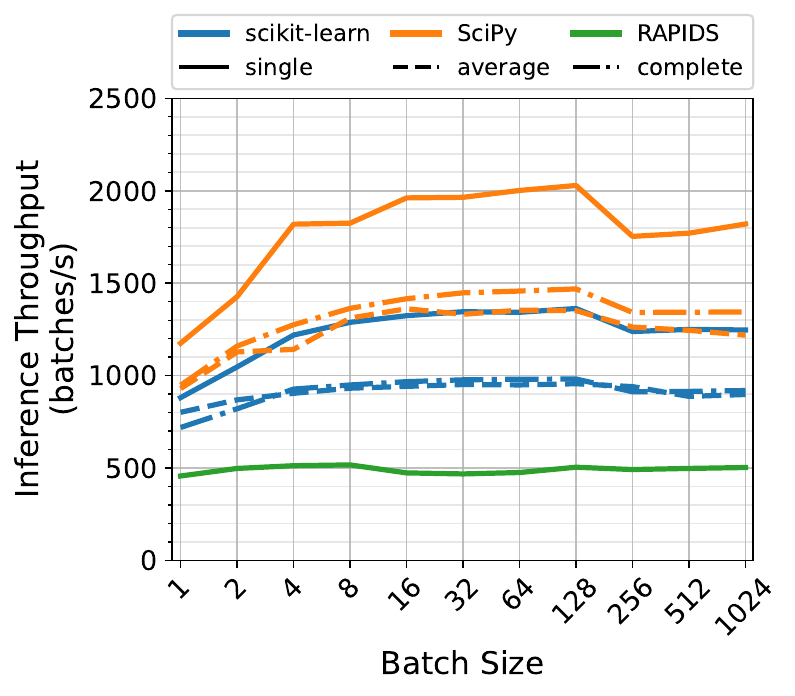}
    \caption{\textbf{Framework Inference Comparison -- Batch Size.} We compare the agglomerative clustering implementations of \textsc{scikit-learn}, \textsc{SciPy}, and \textsc{RAPIDS}. We find that the \textsc{SciPy} implementation is consistently the fastest, but has the downside that it can not guarantee that exactly $k$ clusters are returned at all times. The comparison is conducted with a keep rate $r=50\%$ and we note that the inference throughput is similar for other keep rates. Note  the average and complete linkage functions partially occlude each other.}
    \label{fig:framework_comp}
\end{figure}

\begin{figure*}[!tp]
    \centering
    \includegraphics[width=\linewidth]{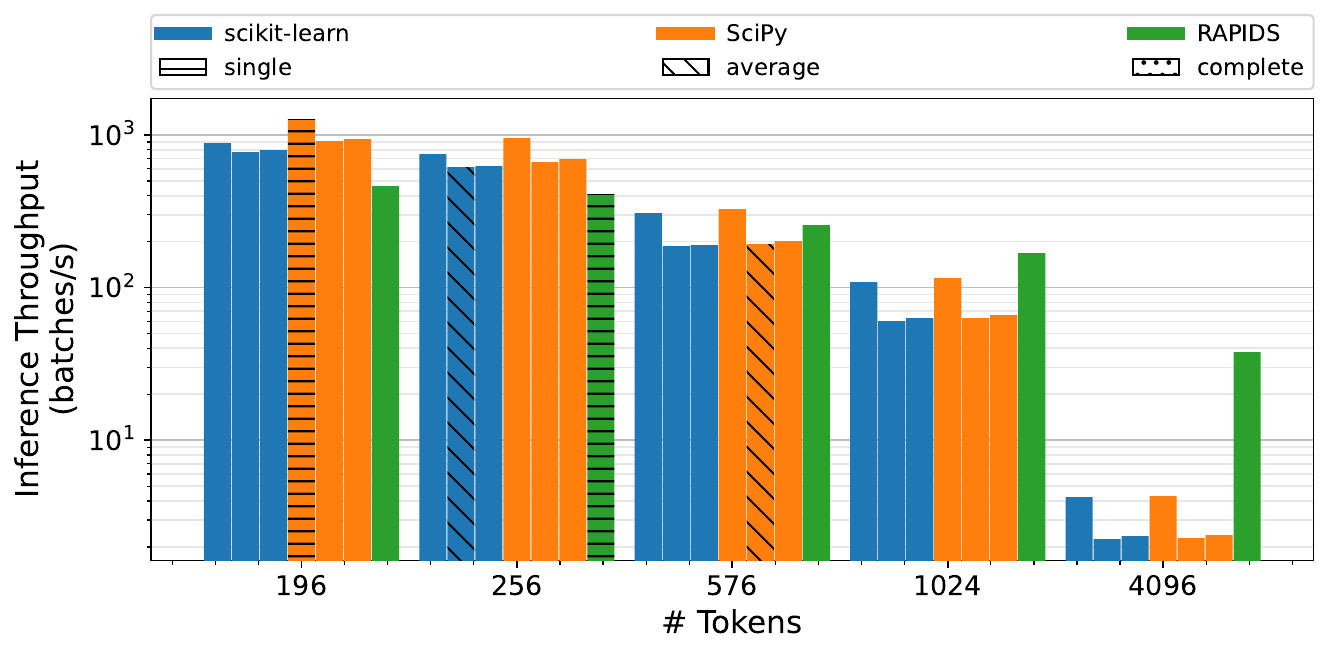}
    \caption{\textbf{Framework Inference Comparison -- Number of Tokens.} We compare the ATC implementations of \textsc{scikit-learn}, \textsc{SciPy}, and RAPIDS, while varying the number of tokens present. We find that as the number of tokens increases, the RAPIDS implementation becomes significantly faster, due to its GPU-based implementation.}
    \label{fig:inferenceIM}
\end{figure*}

\subsection{Framework Comparison}
\textsc{SciPy} implements agglomerative clustering with a $\mathcal{O}(n^2)$ time complexity for all three linkage functions, through the use of the nearest neighbour-chain \cite{SciPy, NNChainingClust}. This matches the time complexity of K-Medoids \cite{FastPAM} and DPC-KNN \cite{DPCKNN_2016}. Similar to the \textsc{scikit-learn} implementation \cite{scikit-learn}, these functions are also CPU-bound and non-batched. However, unlike the \textsc{scikit-learn} implementation, it is not possible to enforce that the \textsc{SciPy} clustering function always outputs exactly $k$ clusters. 
Lastly, the \textsc{RAPIDS} framework from Nvidia implements agglomerative clustering with a CUDA based version of the single linkage function \cite{RAPIDS, cuSLINK}, though still non-batched. For the throughput comparison of the three implementations, we follow the same evaluation protocol as before. The only modification is that instead of measuring the throughput over the entire network, we only measure it over the token reduction function.

We find that the \textsc{SciPy} implementation is significantly faster than both the \textsc{scikit-learn} and \textsc{RAPIDS} implementations, see Figure~\ref{fig:framework_comp}. Surprisingly, we also find that the GPU-based \textsc{RAPIDS} implementation is the slowest of the three frameworks. This is due to the \textsc{RAPIDS} implementation requiring the set of observations as input so that it can compute the distance matrix. In comparison, both \textsc{SciPy} and \textsc{scikit-learn} can take a precomputed distance matrix as input, which can be computed efficiently in a batched manner using PyTorch. We find that when the \textsc{SciPy} and \textsc{scikit-learn} implementations are similarly provided with the set of observations, the throughput can become 2.5-4 times slower depending on the linkage function, in which case the \textsc{RAPIDS} implementation would be faster.

\subsection{Image Size Comparison}
We measure the inference throughput of the aglgomerative clustering algorithm, using \textsc{scikit-learn}, \textsc{SciPy}, and RAPIDS, using a batch size of one, and consider a varying amount of tokens, $[196, 256, 576, 1024, 4096]$, corresponding to images of size $[224\times 224, 256\times 256, 384\times 384, 512\times 512, 1024\times 1024]$ with a patch size of $16\times 16$. We find that when a small number of tokens are considered, \textsc{SciPy} is the fastest, see Figure~\ref{fig:inferenceIM}. However, as the number of tokens is increased, the CPU-based framework achieves similar performance while the GPU-based RAPIDS implementation becomes significantly faster. Specifically, when 4096 tokens have to be clustered, the RAPIDS implementation is 8 times faster than the CPU-based \textsc{SciPy} and \textsc{scikit-learn}. This is a strong indication that GPU-based agglomerative clustering is of great value, as tasks such as image synthesis and object detection \& segmentation are applied on larger image sizes.

\section{Example Clustering Visualizations on ImageNet, NABirds, COCO, and NUS-Wide}\label{sec:reducViz}
\begin{figure*}[!tp]
\resizebox{\linewidth}{!}{%
\begin{tabular}{ccccccccc}
  
 ATC$^{\text{single}}$ & ATC$^{\text{average}}$ & ATC$^{\text{complete}}$ & ATC$^{\text{single}}$ & ATC$^{\text{average}}$ & ATC$^{\text{complete}}$ & ATC$^{\text{single}}$ & ATC$^{\text{average}}$ & ATC$^{\text{complete}}$ \\
\includegraphics[width=0.20\columnwidth]{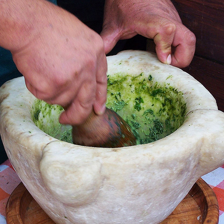}  & \includegraphics[width=0.20\columnwidth]{supp_figures_tables/IM_21007/original.png} & \includegraphics[width=0.20\columnwidth]{supp_figures_tables/IM_21007/original.png} & \includegraphics[width=0.20\columnwidth]{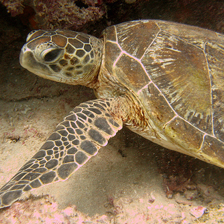}  & \includegraphics[width=0.20\columnwidth]{supp_figures_tables/IM_22469/original.png} & \includegraphics[width=0.20\columnwidth]{supp_figures_tables/IM_22469/original.png} & \includegraphics[width=0.20\columnwidth]{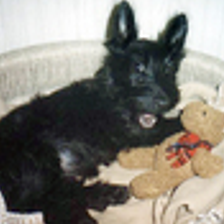}  & \includegraphics[width=0.20\columnwidth]{supp_figures_tables/IM_47649/original.png} & \includegraphics[width=0.20\columnwidth]{supp_figures_tables/IM_47649/original.png}\\
  \includegraphics[width=0.20\columnwidth]{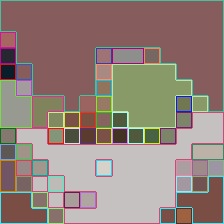} &   \includegraphics[width=0.20\columnwidth]{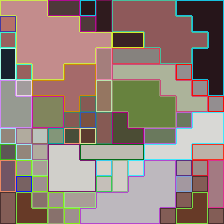} & \includegraphics[width=0.20\columnwidth]{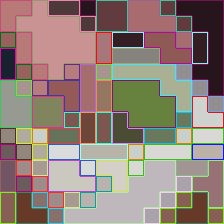} &   \includegraphics[width=0.20\columnwidth]{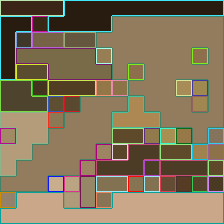} &   \includegraphics[width=0.20\columnwidth]{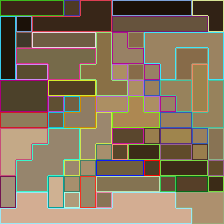} & \includegraphics[width=0.20\columnwidth]{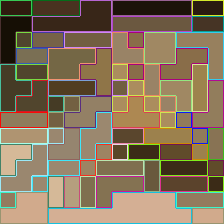} &   \includegraphics[width=0.20\columnwidth]{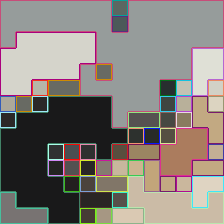} &   \includegraphics[width=0.20\columnwidth]{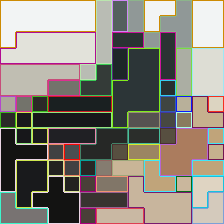} & \includegraphics[width=0.20\columnwidth]{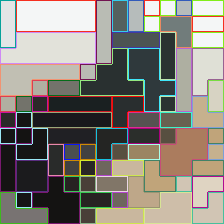} \\
    \includegraphics[width=0.20\columnwidth]{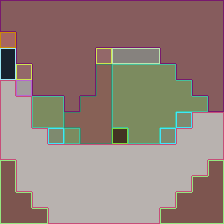} &   \includegraphics[width=0.20\columnwidth]{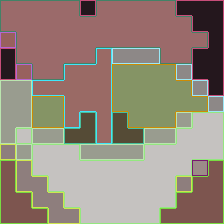} & \includegraphics[width=0.20\columnwidth]{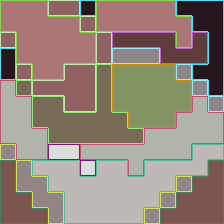} &     \includegraphics[width=0.20\columnwidth]{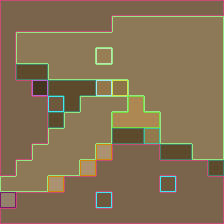} &   \includegraphics[width=0.20\columnwidth]{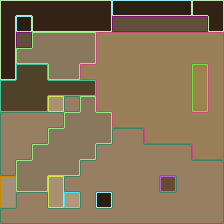} & \includegraphics[width=0.20\columnwidth]{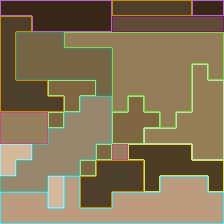} &     \includegraphics[width=0.20\columnwidth]{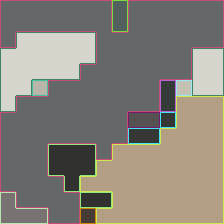} &   \includegraphics[width=0.20\columnwidth]{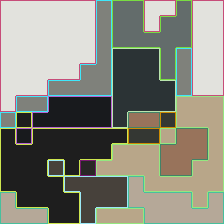} & \includegraphics[width=0.20\columnwidth]{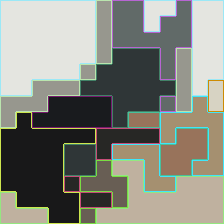} \\
      \includegraphics[width=0.20\columnwidth]{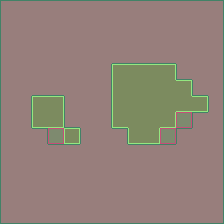} &   \includegraphics[width=0.20\columnwidth]{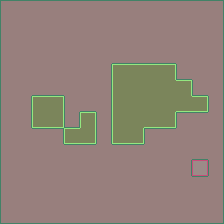} & \includegraphics[width=0.20\columnwidth]{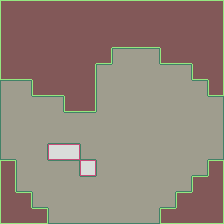} &       \includegraphics[width=0.20\columnwidth]{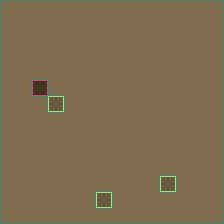} &   \includegraphics[width=0.20\columnwidth]{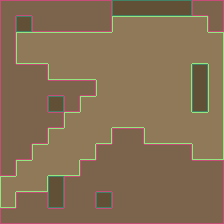} & \includegraphics[width=0.20\columnwidth]{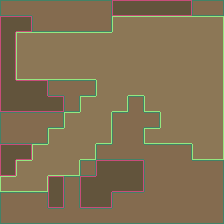}&       \includegraphics[width=0.20\columnwidth]{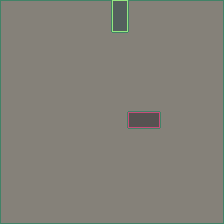} &   \includegraphics[width=0.20\columnwidth]{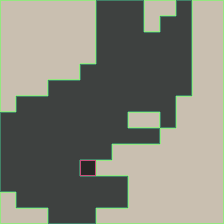} & \includegraphics[width=0.20\columnwidth]{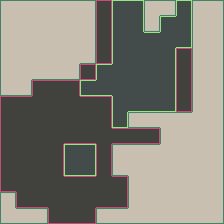} \\
\end{tabular}
}
\caption{\textbf{Token Merging Visualization with DeiT-S and $r=25$\% -- IM.} We visualize the three token merging steps using ATC with the three considered linkage functions. The first row is the input image, and each subsequent row is the constructed clusters after the first, second, and third reduction stage. }
\label{fig:IMViz}
\end{figure*}

\begin{figure*}[!tp]
\resizebox{\linewidth}{!}{%
\begin{tabular}{ccccccccc}
  
 ATC$^{\text{single}}$ & ATC$^{\text{average}}$ & ATC$^{\text{complete}}$ & ATC$^{\text{single}}$ & ATC$^{\text{average}}$ & ATC$^{\text{complete}}$ & ATC$^{\text{single}}$ & ATC$^{\text{average}}$ & ATC$^{\text{complete}}$ \\
\includegraphics[width=0.20\columnwidth]{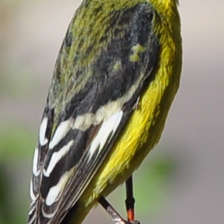}  & \includegraphics[width=0.20\columnwidth]{supp_figures_tables/NAB_b2f5c/original.png} & \includegraphics[width=0.20\columnwidth]{supp_figures_tables/NAB_b2f5c/original.png} & \includegraphics[width=0.20\columnwidth]{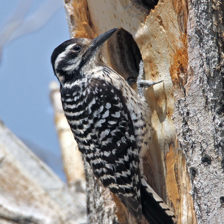}  & \includegraphics[width=0.20\columnwidth]{supp_figures_tables/NAB_7f475/original.png} & \includegraphics[width=0.20\columnwidth]{supp_figures_tables/NAB_7f475/original.png} & \includegraphics[width=0.20\columnwidth]{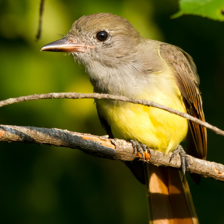}  & \includegraphics[width=0.20\columnwidth]{supp_figures_tables/NAB_3b4ac/original.png} & \includegraphics[width=0.20\columnwidth]{supp_figures_tables/NAB_3b4ac/original.png}\\
  \includegraphics[width=0.20\columnwidth]{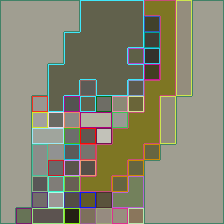} &   \includegraphics[width=0.20\columnwidth]{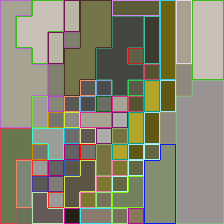} & \includegraphics[width=0.20\columnwidth]{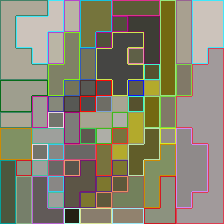} &   \includegraphics[width=0.20\columnwidth]{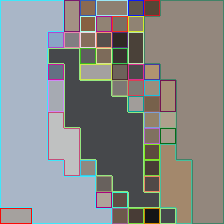} &   \includegraphics[width=0.20\columnwidth]{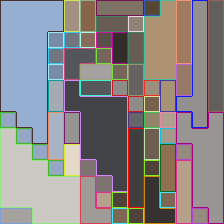} & \includegraphics[width=0.20\columnwidth]{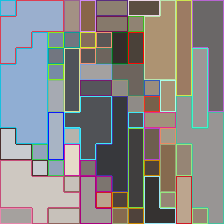} &   \includegraphics[width=0.20\columnwidth]{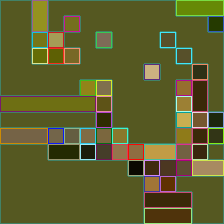} &   \includegraphics[width=0.20\columnwidth]{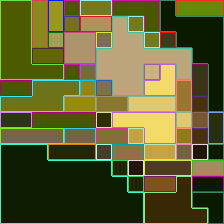} & \includegraphics[width=0.20\columnwidth]{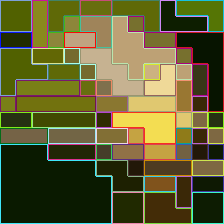} \\
    \includegraphics[width=0.20\columnwidth]{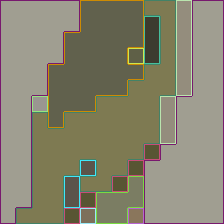} &   \includegraphics[width=0.20\columnwidth]{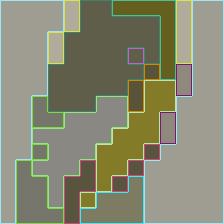} & \includegraphics[width=0.20\columnwidth]{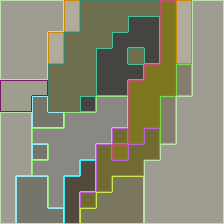} &     \includegraphics[width=0.20\columnwidth]{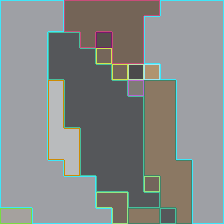} &   \includegraphics[width=0.20\columnwidth]{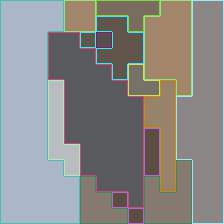} & \includegraphics[width=0.20\columnwidth]{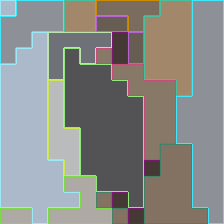} &     \includegraphics[width=0.20\columnwidth]{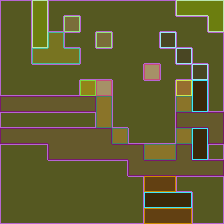} &   \includegraphics[width=0.20\columnwidth]{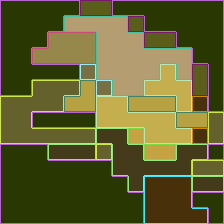} & \includegraphics[width=0.20\columnwidth]{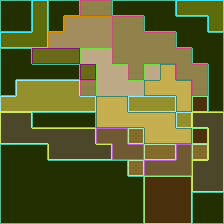} \\
      \includegraphics[width=0.20\columnwidth]{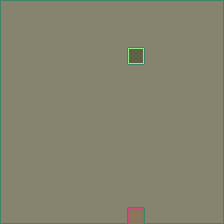} &   \includegraphics[width=0.20\columnwidth]{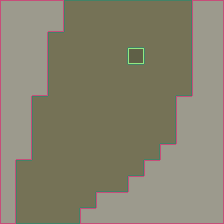} & \includegraphics[width=0.20\columnwidth]{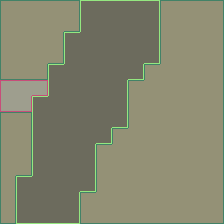} &       \includegraphics[width=0.20\columnwidth]{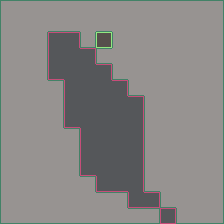} &   \includegraphics[width=0.20\columnwidth]{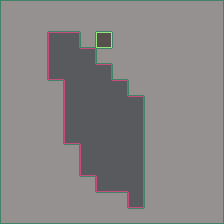} & \includegraphics[width=0.20\columnwidth]{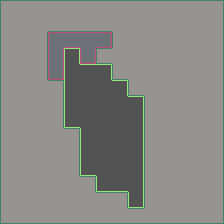}&       \includegraphics[width=0.20\columnwidth]{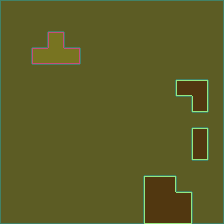} &   \includegraphics[width=0.20\columnwidth]{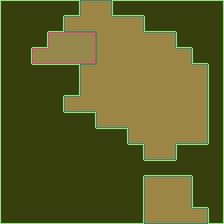} & \includegraphics[width=0.20\columnwidth]{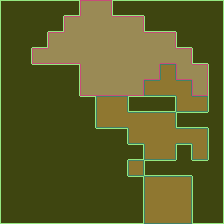} \\
\end{tabular}
}
\caption{\textbf{Token Merging Visualization with DeiT-S and $r=25$\% -- NAB.} We visualize the three token merging steps using ATC with the three considered linkage functions. The first row is the input image, and each subsequent row is the constructed clusters after the first, second, and third reduction stage. }
\label{fig:NABViz}
\end{figure*}

\begin{figure*}[!tp]
\resizebox{\linewidth}{!}{%
\begin{tabular}{ccccccccc}
  
 ATC$^{\text{single}}$ & ATC$^{\text{average}}$ & ATC$^{\text{complete}}$ & ATC$^{\text{single}}$ & ATC$^{\text{average}}$ & ATC$^{\text{complete}}$ & ATC$^{\text{single}}$ & ATC$^{\text{average}}$ & ATC$^{\text{complete}}$ \\
\includegraphics[width=0.20\columnwidth]{supp_figures_tables/COCO_58111/original.png}  & \includegraphics[width=0.20\columnwidth]{supp_figures_tables/COCO_58111/original.png} & \includegraphics[width=0.20\columnwidth]{supp_figures_tables/COCO_58111/original.png} & \includegraphics[width=0.20\columnwidth]{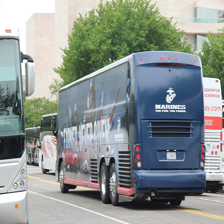}  & \includegraphics[width=0.20\columnwidth]{supp_figures_tables/COCO_111619/original.png} & \includegraphics[width=0.20\columnwidth]{supp_figures_tables/COCO_111619/original.png} & \includegraphics[width=0.20\columnwidth]{supp_figures_tables/COCO_569893/original.png}  & \includegraphics[width=0.20\columnwidth]{supp_figures_tables/COCO_569893/original.png} & \includegraphics[width=0.20\columnwidth]{supp_figures_tables/COCO_569893/original.png}\\
  \includegraphics[width=0.20\columnwidth]{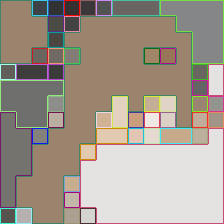} &   \includegraphics[width=0.20\columnwidth]{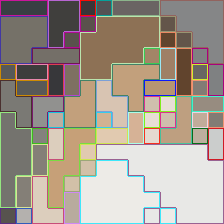} & \includegraphics[width=0.20\columnwidth]{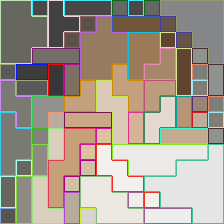} &   \includegraphics[width=0.20\columnwidth]{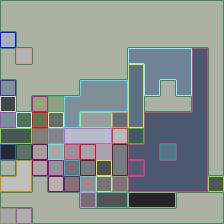} &   \includegraphics[width=0.20\columnwidth]{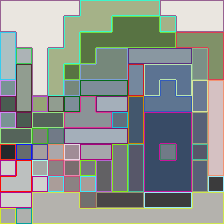} & \includegraphics[width=0.20\columnwidth]{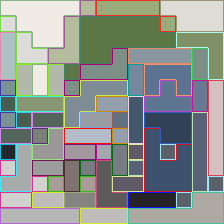} &   \includegraphics[width=0.20\columnwidth]{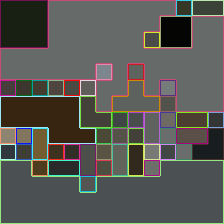} &   \includegraphics[width=0.20\columnwidth]{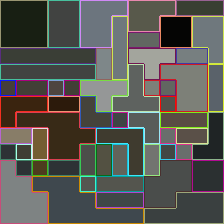} & \includegraphics[width=0.20\columnwidth]{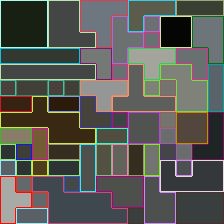} \\
    \includegraphics[width=0.20\columnwidth]{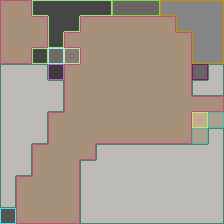} &   \includegraphics[width=0.20\columnwidth]{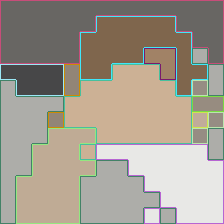} & \includegraphics[width=0.20\columnwidth]{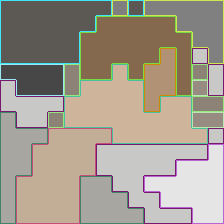} &     \includegraphics[width=0.20\columnwidth]{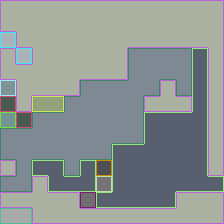} &   \includegraphics[width=0.20\columnwidth]{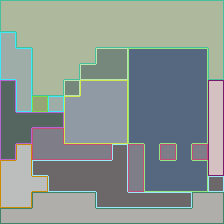} & \includegraphics[width=0.20\columnwidth]{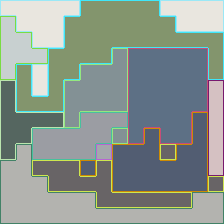} &     \includegraphics[width=0.20\columnwidth]{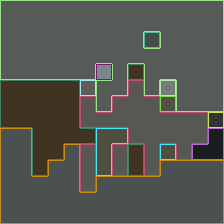} &   \includegraphics[width=0.20\columnwidth]{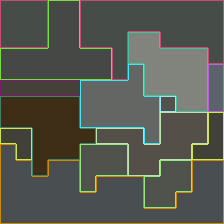} & \includegraphics[width=0.20\columnwidth]{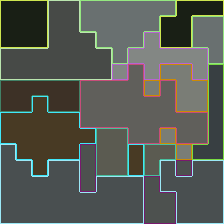} \\
      \includegraphics[width=0.20\columnwidth]{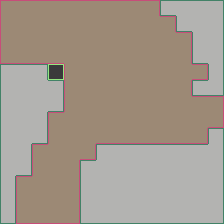} &   \includegraphics[width=0.20\columnwidth]{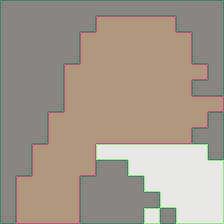} & \includegraphics[width=0.20\columnwidth]{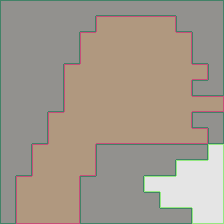} &       \includegraphics[width=0.20\columnwidth]{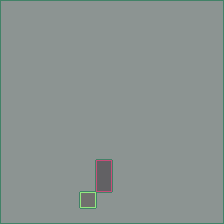} &   \includegraphics[width=0.20\columnwidth]{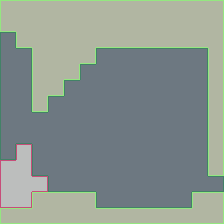} & \includegraphics[width=0.20\columnwidth]{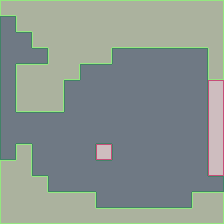}&       \includegraphics[width=0.20\columnwidth]{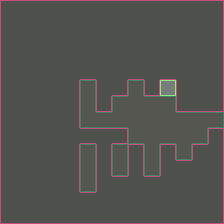} &   \includegraphics[width=0.20\columnwidth]{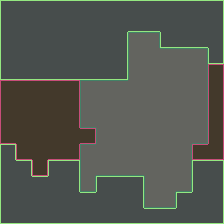} & \includegraphics[width=0.20\columnwidth]{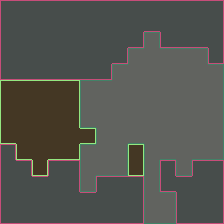} \\
\end{tabular}
}
\caption{\textbf{Token Merging Visualization with DeiT-S and $r=25$\% -- COCO.} We visualize the three token merging steps using ATC with the three considered linkage functions. The first row is the input image, and each subsequent row is the constructed clusters after the first, second, and third reduction stage.}
\label{fig:COCOViz}
\end{figure*}

\begin{figure*}[!tp]
\resizebox{\linewidth}{!}{%
\begin{tabular}{ccccccccc}
  
 ATC$^{\text{single}}$ & ATC$^{\text{average}}$ & ATC$^{\text{complete}}$ & ATC$^{\text{single}}$ & ATC$^{\text{average}}$ & ATC$^{\text{complete}}$ & ATC$^{\text{single}}$ & ATC$^{\text{average}}$ & ATC$^{\text{complete}}$ \\
\includegraphics[width=0.20\columnwidth]{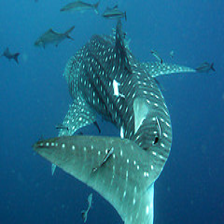}  & \includegraphics[width=0.20\columnwidth]{supp_figures_tables/NUS_85530/original.png} & \includegraphics[width=0.20\columnwidth]{supp_figures_tables/NUS_85530/original.png} & \includegraphics[width=0.20\columnwidth]{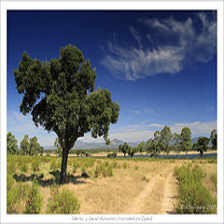}  & \includegraphics[width=0.20\columnwidth]{supp_figures_tables/NUS_125808/original.png} & \includegraphics[width=0.20\columnwidth]{supp_figures_tables/NUS_125808/original.png} & \includegraphics[width=0.20\columnwidth]{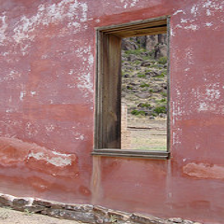}  & \includegraphics[width=0.20\columnwidth]{supp_figures_tables/NUS_165144/original.png} & \includegraphics[width=0.20\columnwidth]{supp_figures_tables/NUS_165144/original.png}\\
  \includegraphics[width=0.20\columnwidth]{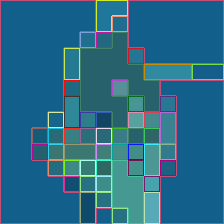} &   \includegraphics[width=0.20\columnwidth]{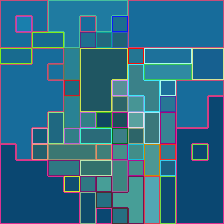} & \includegraphics[width=0.20\columnwidth]{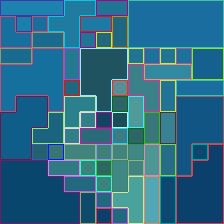} &   \includegraphics[width=0.20\columnwidth]{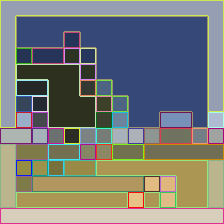} &   \includegraphics[width=0.20\columnwidth]{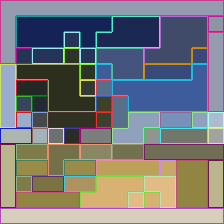} & \includegraphics[width=0.20\columnwidth]{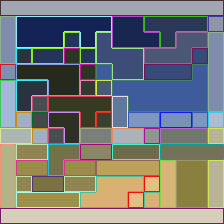} &   \includegraphics[width=0.20\columnwidth]{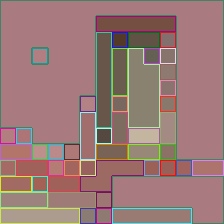} &   \includegraphics[width=0.20\columnwidth]{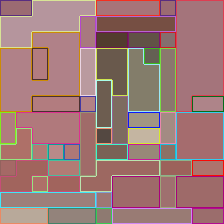} & \includegraphics[width=0.20\columnwidth]{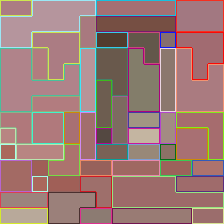} \\
    \includegraphics[width=0.20\columnwidth]{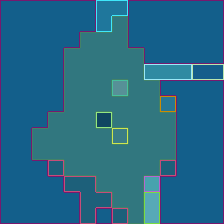} &   \includegraphics[width=0.20\columnwidth]{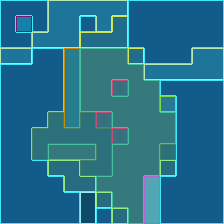} & \includegraphics[width=0.20\columnwidth]{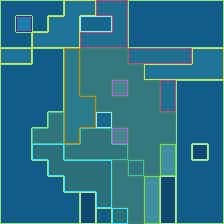} &     \includegraphics[width=0.20\columnwidth]{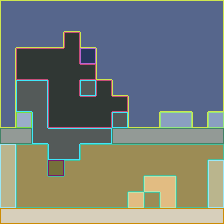} &   \includegraphics[width=0.20\columnwidth]{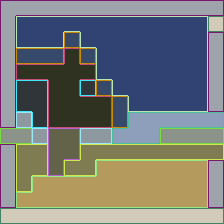} & \includegraphics[width=0.20\columnwidth]{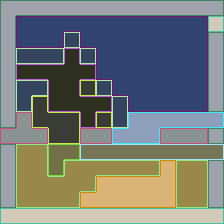} &     \includegraphics[width=0.20\columnwidth]{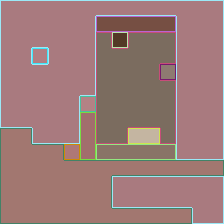} &   \includegraphics[width=0.20\columnwidth]{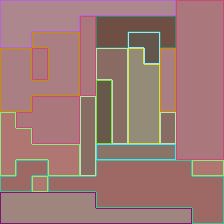} & \includegraphics[width=0.20\columnwidth]{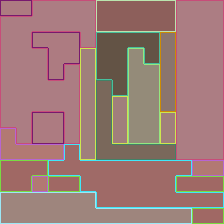} \\
      \includegraphics[width=0.20\columnwidth]{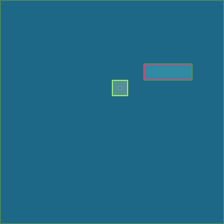} &   \includegraphics[width=0.20\columnwidth]{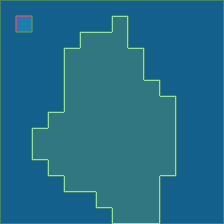} & \includegraphics[width=0.20\columnwidth]{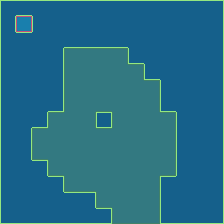} &       \includegraphics[width=0.20\columnwidth]{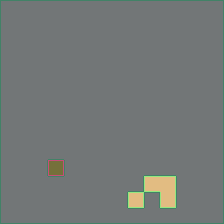} &   \includegraphics[width=0.20\columnwidth]{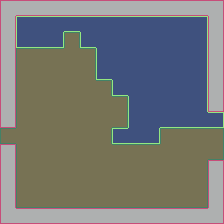} & \includegraphics[width=0.20\columnwidth]{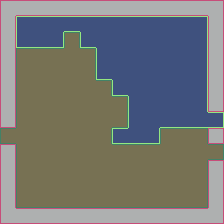}&       \includegraphics[width=0.20\columnwidth]{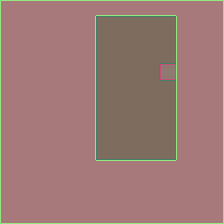} &   \includegraphics[width=0.20\columnwidth]{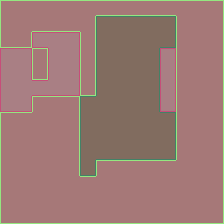} & \includegraphics[width=0.20\columnwidth]{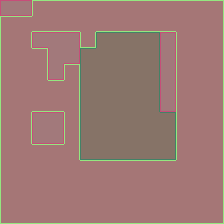} \\
\end{tabular}
}
\caption{\textbf{Token Merging Visualization with DeiT-S and $r=25$\% -- NUS.} We visualize the three token merging steps using ATC with the three considered linkage functions. The first row is the input image, and each subsequent row is the constructed clusters after the first, second, and third reduction stage. }
\label{fig:NUSViz}
\end{figure*}

We provide examples of the token clustering assignments with ATC using all three linkage functions. We present three images from ImageNet (Figure~\ref{fig:IMViz}), NABirds (Figure~\ref{fig:NABViz}), COCO (Figure~\ref{fig:COCOViz}), and NUS-WIDE (Figure~\ref{fig:NUSViz}), all extracted with the DeiT-Small backbone and a keep rate of $r=25\%$.


\section{Example Image Synthesis Results}\label{sec:synthesisViz}
\begin{figure}[!tp]
\centering
\begin{tabular}{cccc}

ToMe  & ATC$^\text{single}$  & ATC$^\text{average}$  & ATC$^\text{complete}$\\
\includegraphics[width=0.20\linewidth]{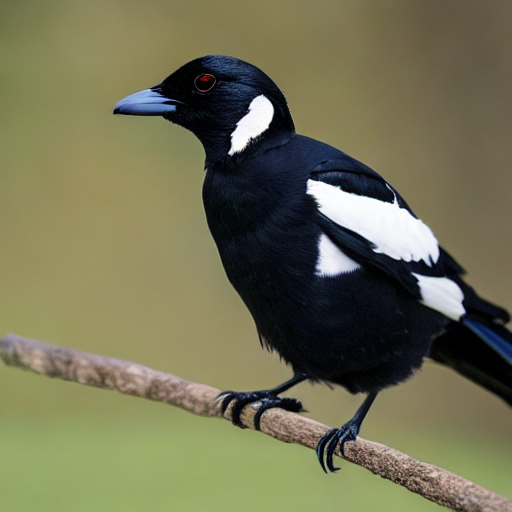} & \includegraphics[width=0.20\linewidth]{supp_figures_tables/sd_18_1/sd.png} & \includegraphics[width=0.20\linewidth]{supp_figures_tables/sd_18_1/sd.png} & \includegraphics[width=0.20\linewidth]{supp_figures_tables/sd_18_1/sd.png} \\
\includegraphics[width=0.20\linewidth]{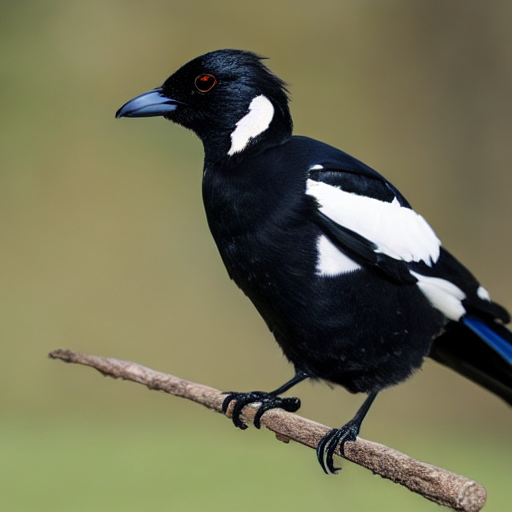} & \includegraphics[width=0.20\linewidth]{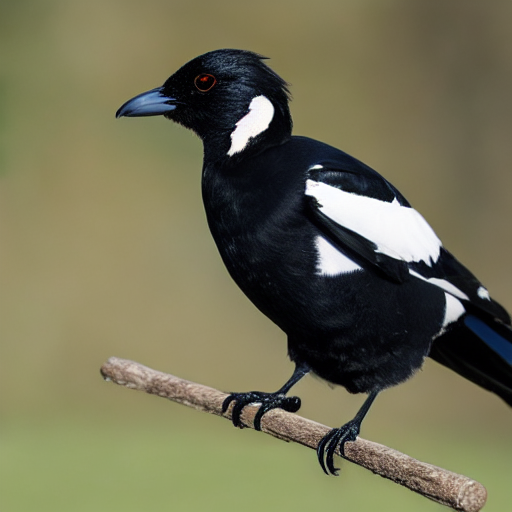} & \includegraphics[width=0.20\linewidth]{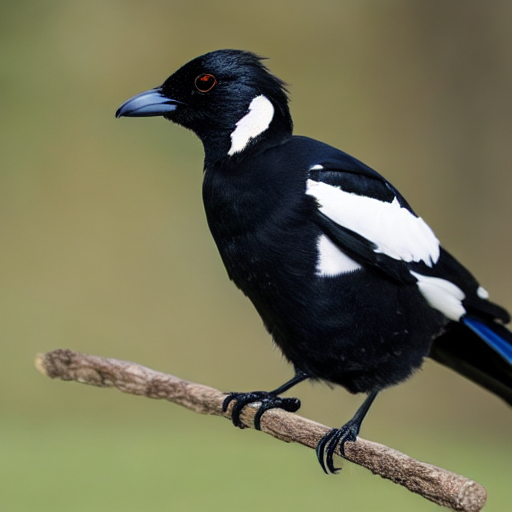} & \includegraphics[width=0.20\linewidth]{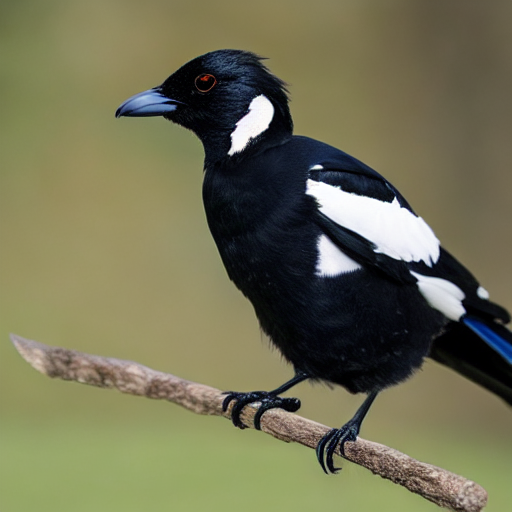} \\
\includegraphics[width=0.20\linewidth]{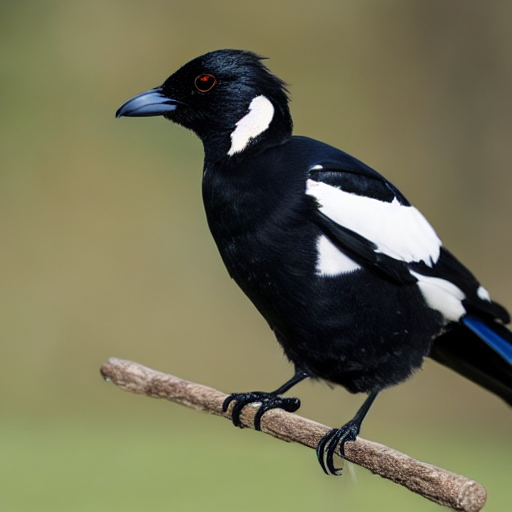} & \includegraphics[width=0.20\linewidth]{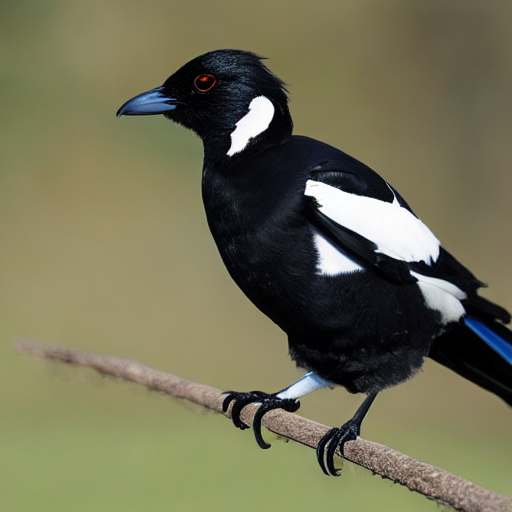} & \includegraphics[width=0.20\linewidth]{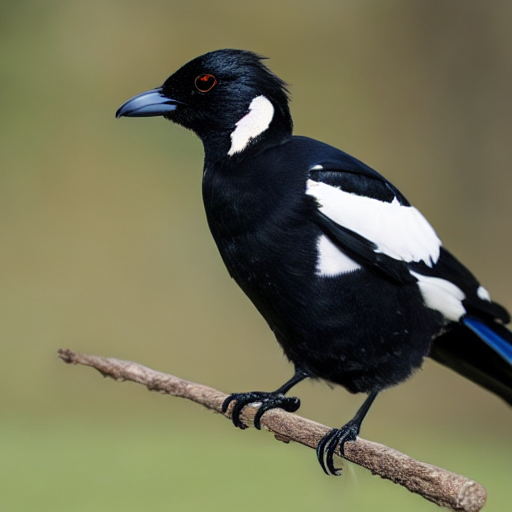} & \includegraphics[width=0.20\linewidth]{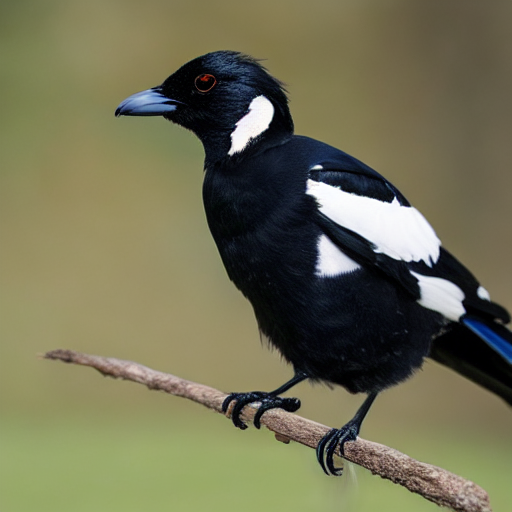} \\
\includegraphics[width=0.20\linewidth]{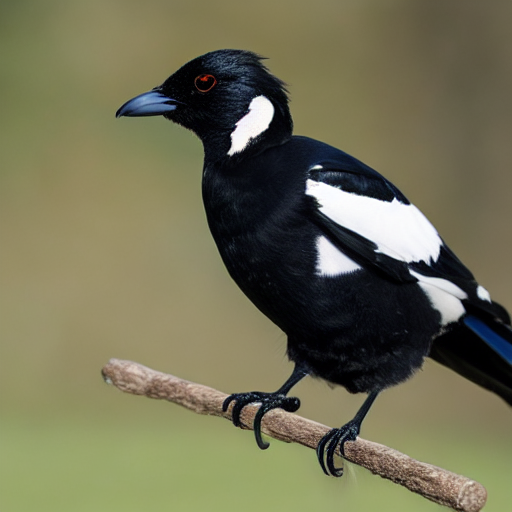} & \includegraphics[width=0.20\linewidth]{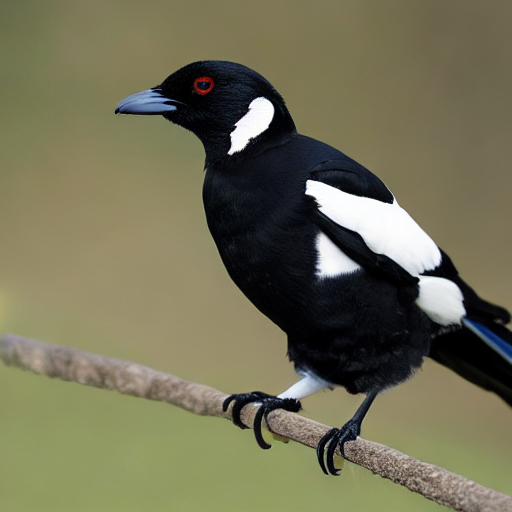} & \includegraphics[width=0.20\linewidth]{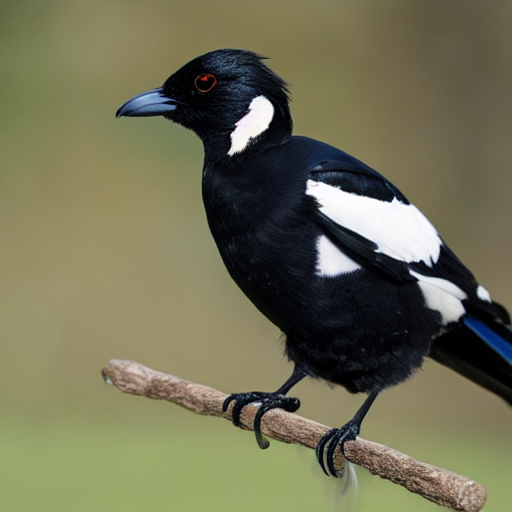} & \includegraphics[width=0.20\linewidth]{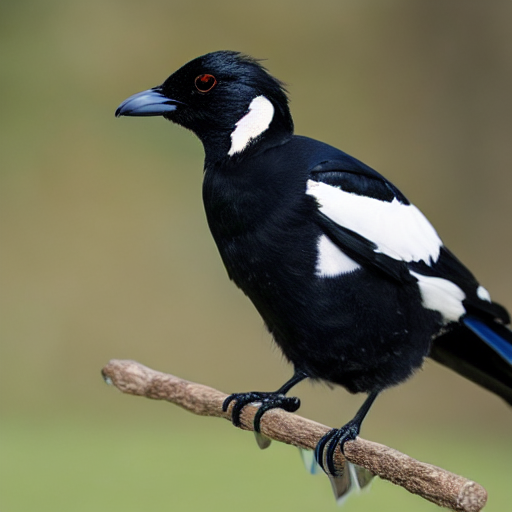} \\
\includegraphics[width=0.20\linewidth]{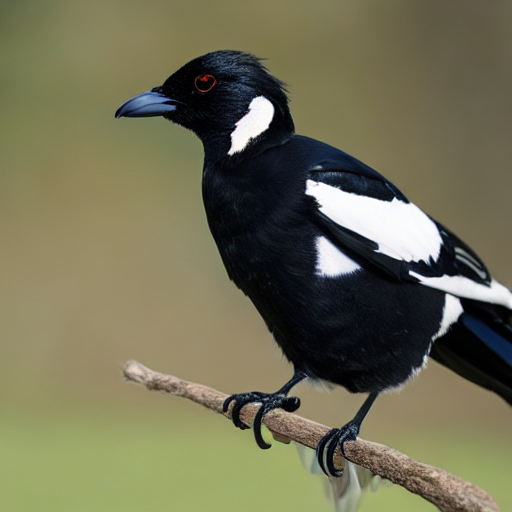} & \includegraphics[width=0.20\linewidth]{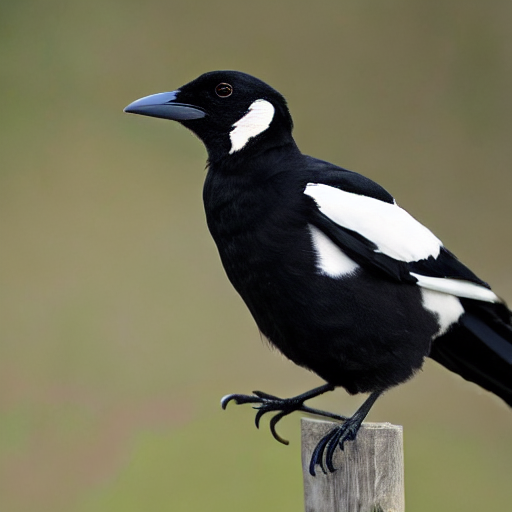} & \includegraphics[width=0.20\linewidth]{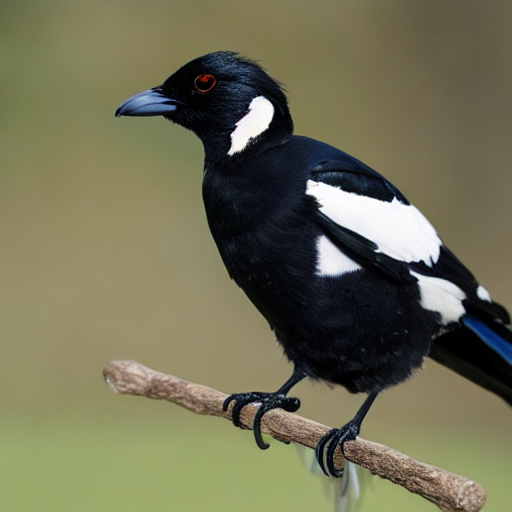} & \includegraphics[width=0.20\linewidth]{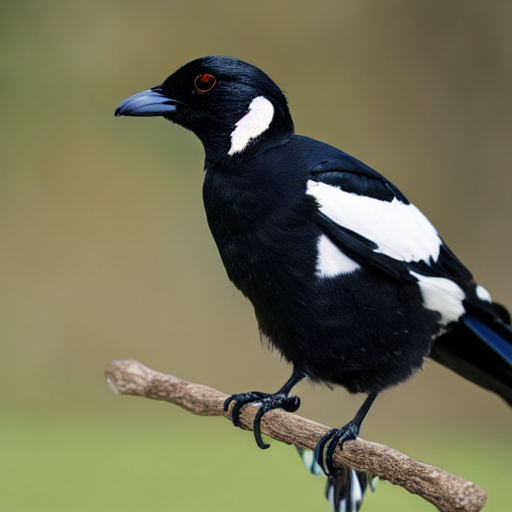}  \\
\includegraphics[width=0.20\linewidth]{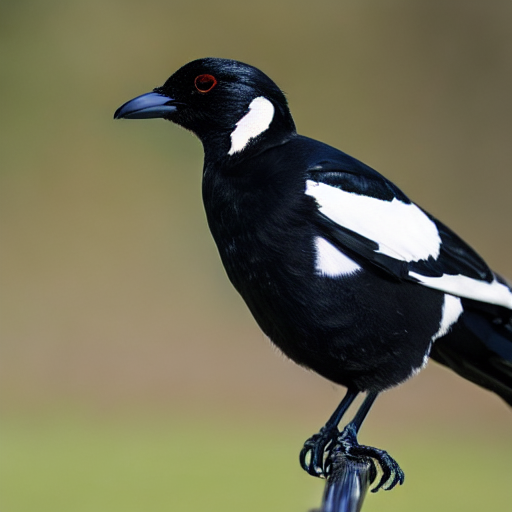} & \includegraphics[width=0.20\linewidth]{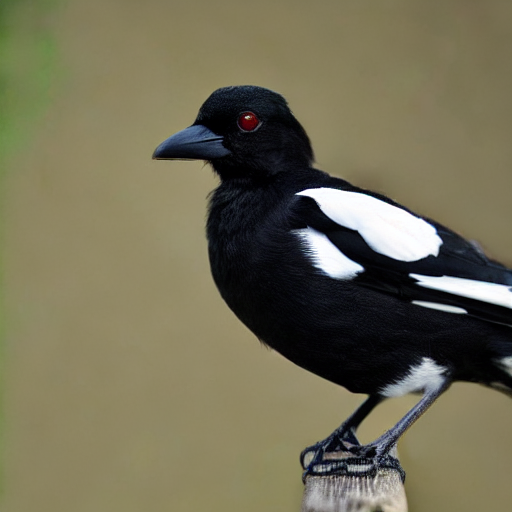} & \includegraphics[width=0.20\linewidth]{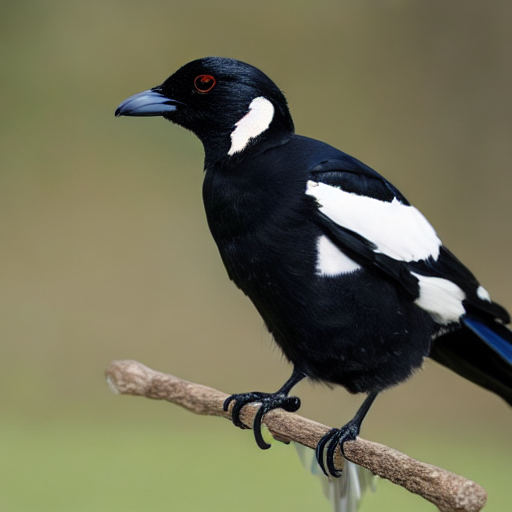} & \includegraphics[width=0.20\linewidth]{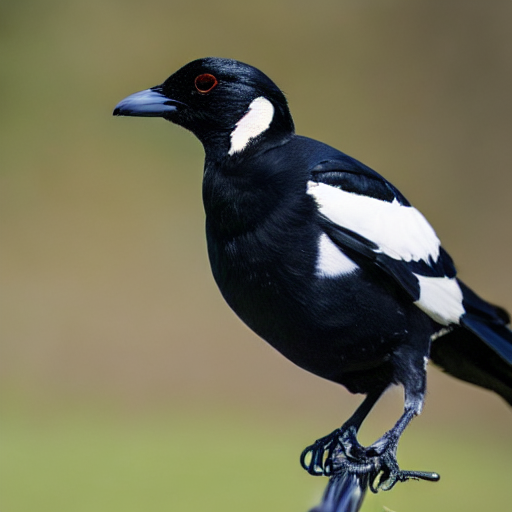} \\
 \includegraphics[width=0.20\linewidth]{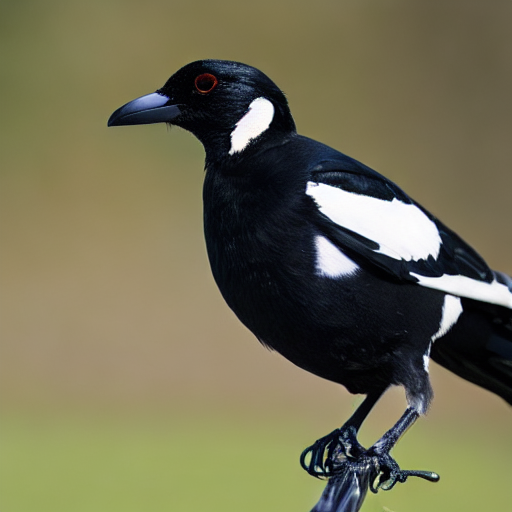} & \includegraphics[width=0.20\linewidth]{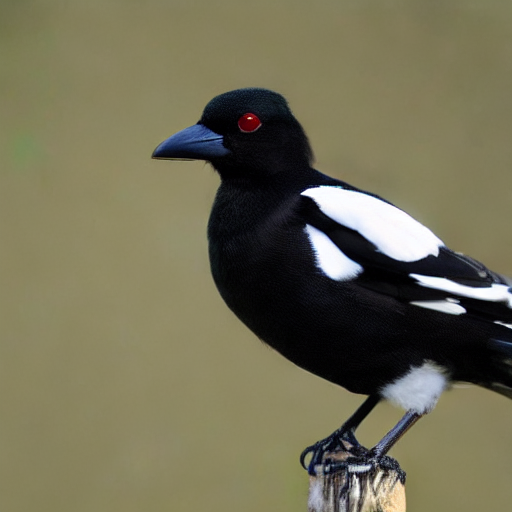} & \includegraphics[width=0.20\linewidth]{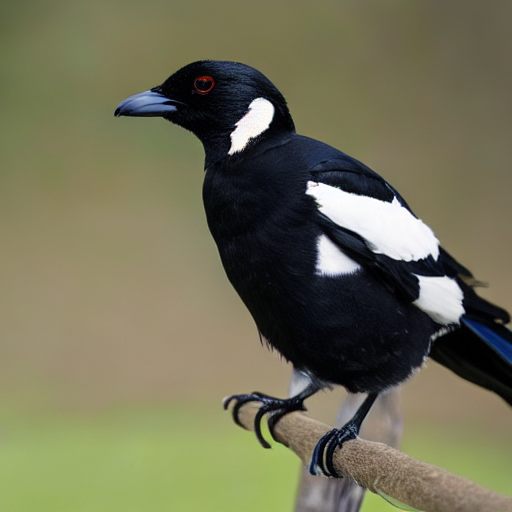} & \includegraphics[width=0.20\linewidth]{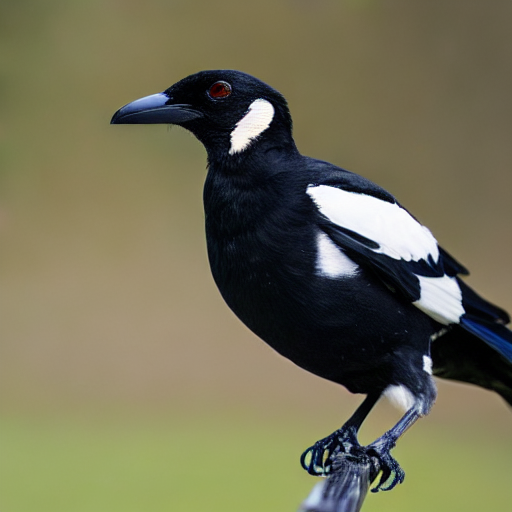} \\

\end{tabular}
\caption{\textbf{Image Synthesis Results -- Magpie.} The first row is the standard Stable Diffusion result, with each subsequent row having token merging applied with $r\in\{90,80,70,60,50,40\}\%$.}
\label{fig:sdVizSupp1}
\end{figure}

\begin{figure}[!tp]
\centering
\begin{tabular}{cccc}

ToMe  & ATC$^\text{single}$  & ATC$^\text{average}$  & ATC$^\text{complete}$\\
\includegraphics[width=0.20\linewidth]{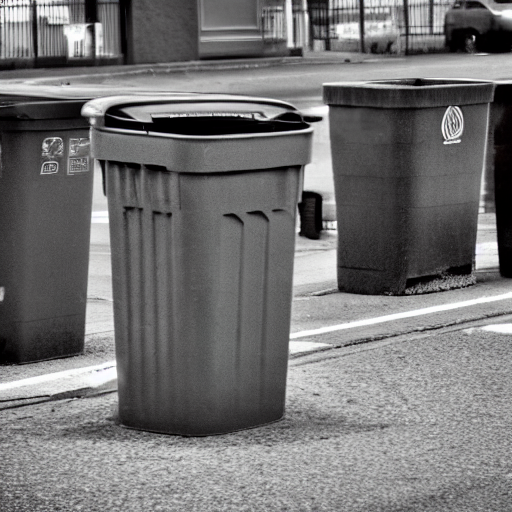} & \includegraphics[width=0.20\linewidth]{supp_figures_tables/sd_412_1/sd.png} & \includegraphics[width=0.20\linewidth]{supp_figures_tables/sd_412_1/sd.png} & \includegraphics[width=0.20\linewidth]{supp_figures_tables/sd_412_1/sd.png} \\
\includegraphics[width=0.20\linewidth]{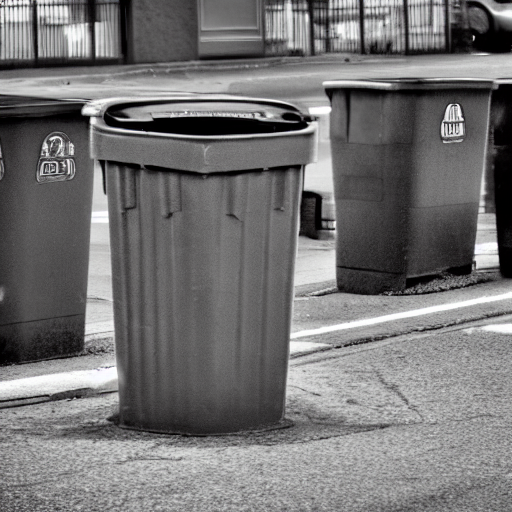} & \includegraphics[width=0.20\linewidth]{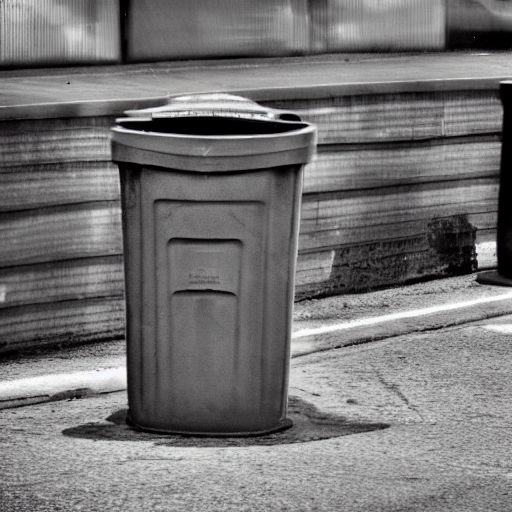} & \includegraphics[width=0.20\linewidth]{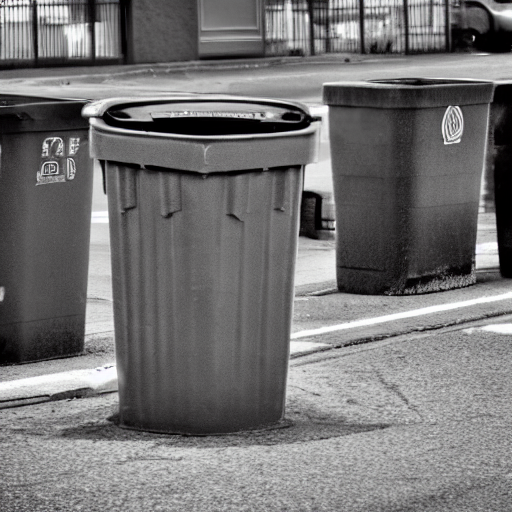} & \includegraphics[width=0.20\linewidth]{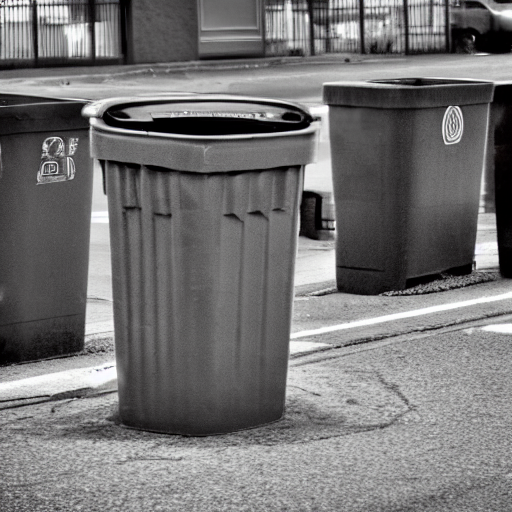} \\
\includegraphics[width=0.20\linewidth]{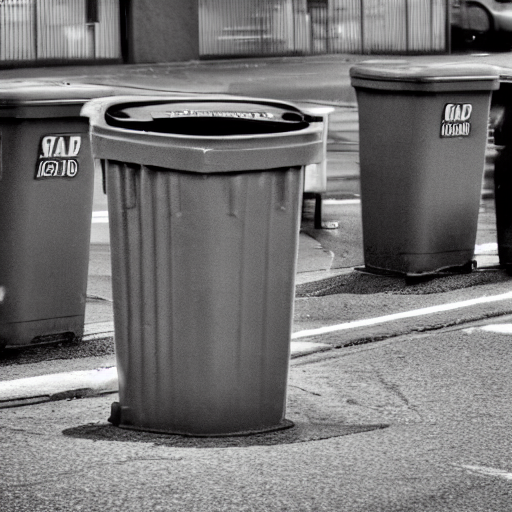} & \includegraphics[width=0.20\linewidth]{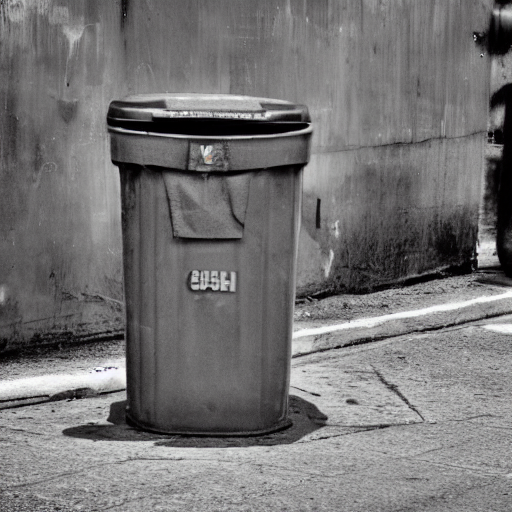} & \includegraphics[width=0.20\linewidth]{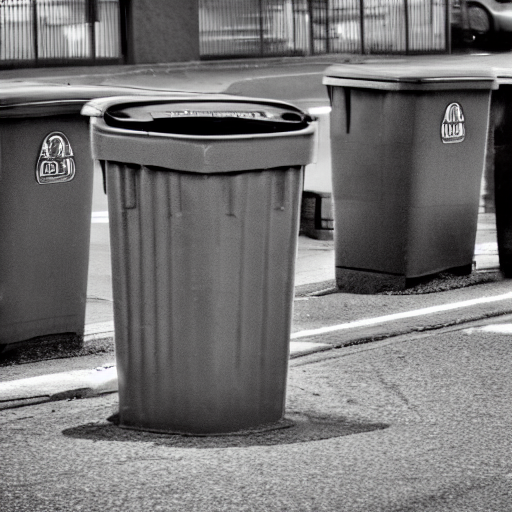} & \includegraphics[width=0.20\linewidth]{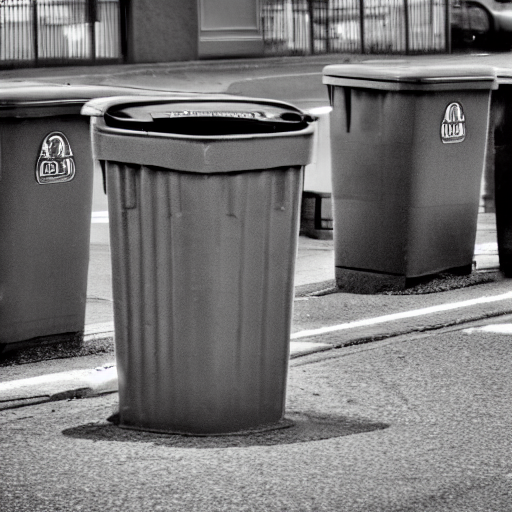} \\
\includegraphics[width=0.20\linewidth]{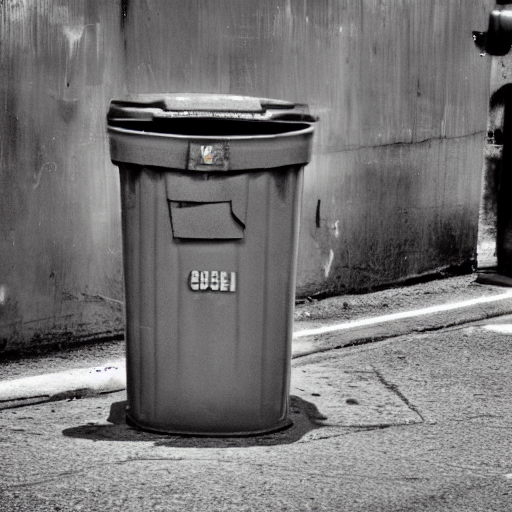} & \includegraphics[width=0.20\linewidth]{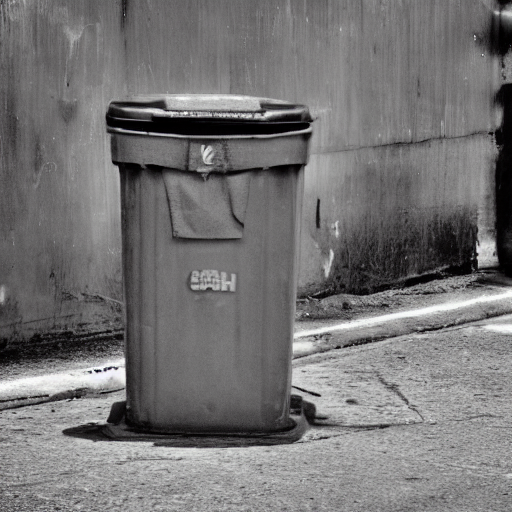} & \includegraphics[width=0.20\linewidth]{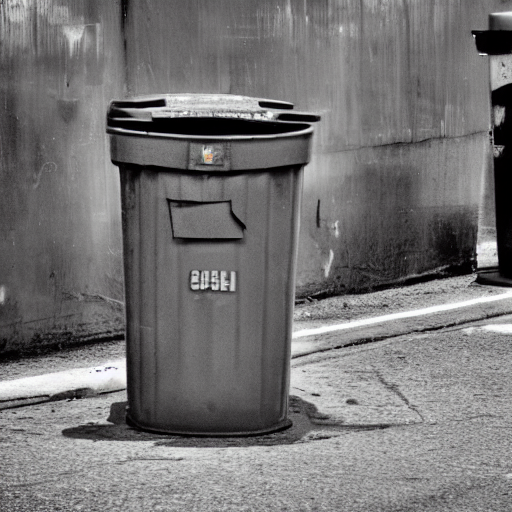} & \includegraphics[width=0.20\linewidth]{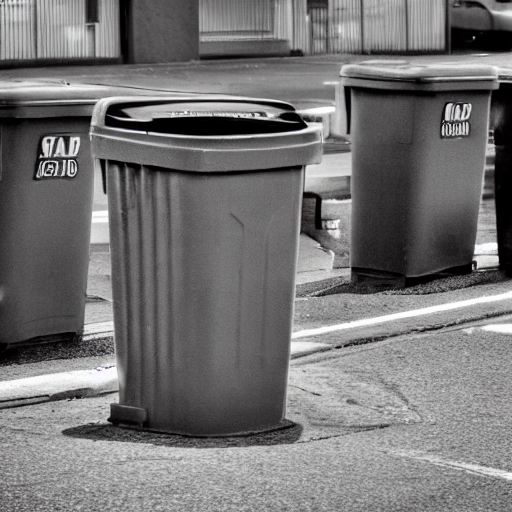} \\
\includegraphics[width=0.20\linewidth]{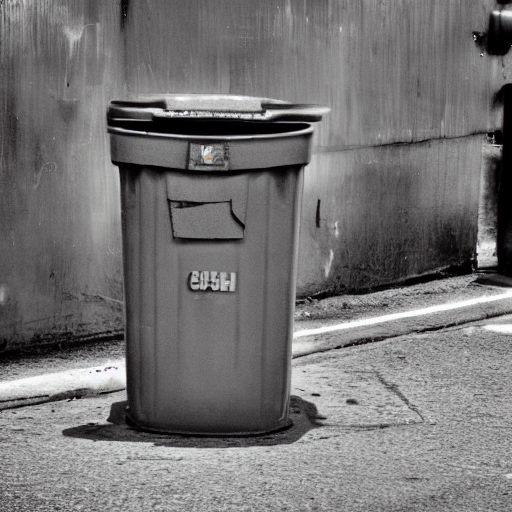} & \includegraphics[width=0.20\linewidth]{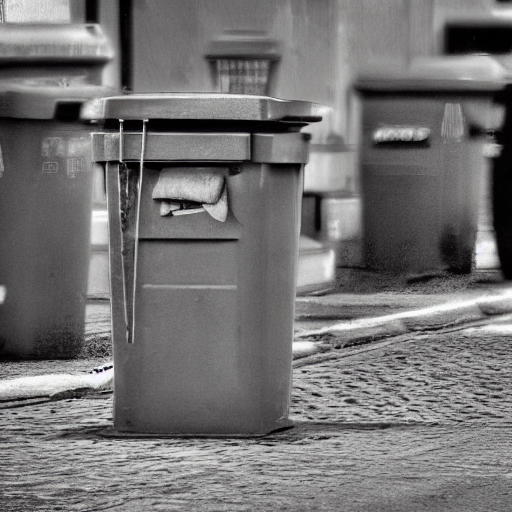} & \includegraphics[width=0.20\linewidth]{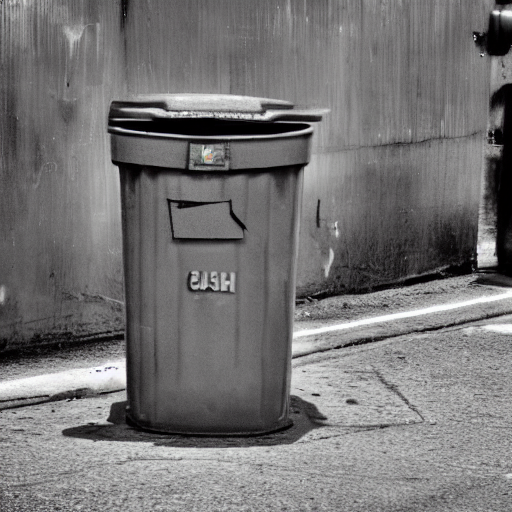} & \includegraphics[width=0.20\linewidth]{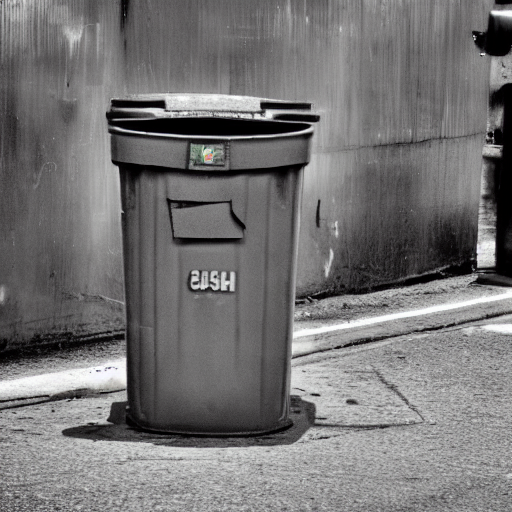}  \\
\includegraphics[width=0.20\linewidth]{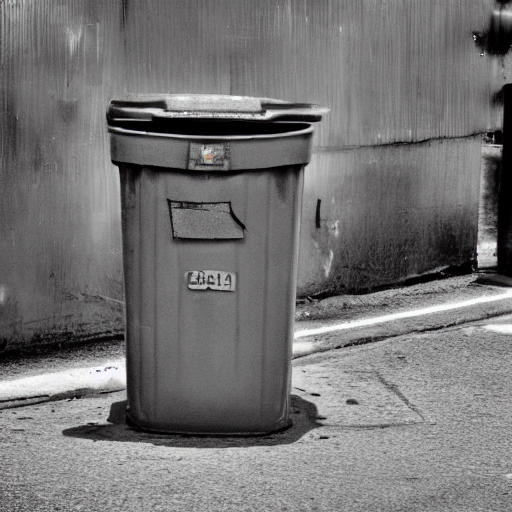} & \includegraphics[width=0.20\linewidth]{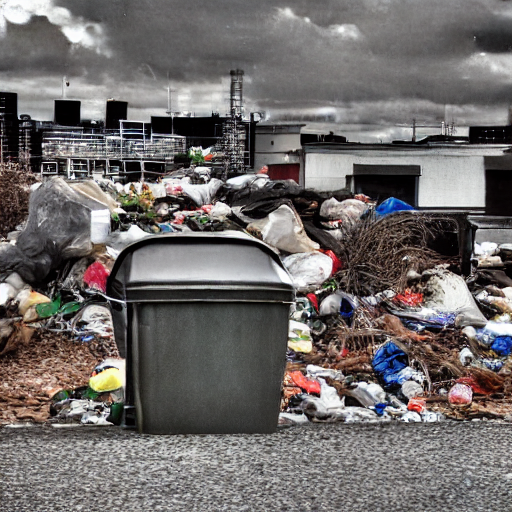} & \includegraphics[width=0.20\linewidth]{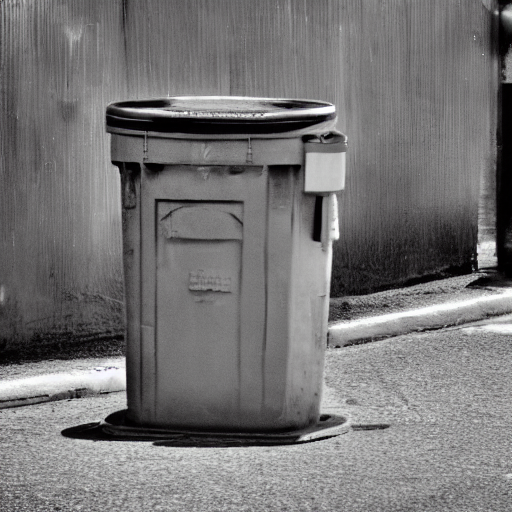} & \includegraphics[width=0.20\linewidth]{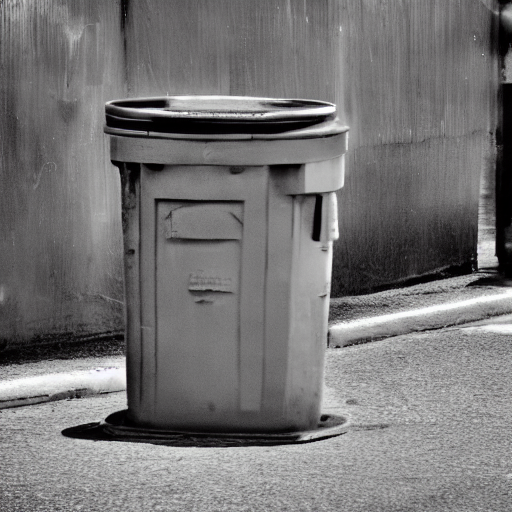} \\
 \includegraphics[width=0.20\linewidth]{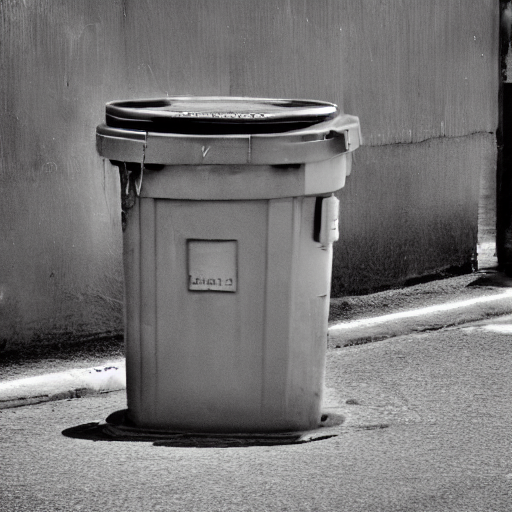} & \includegraphics[width=0.20\linewidth]{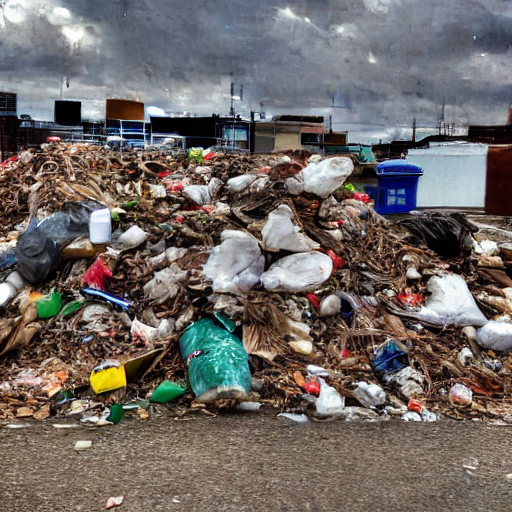} & \includegraphics[width=0.20\linewidth]{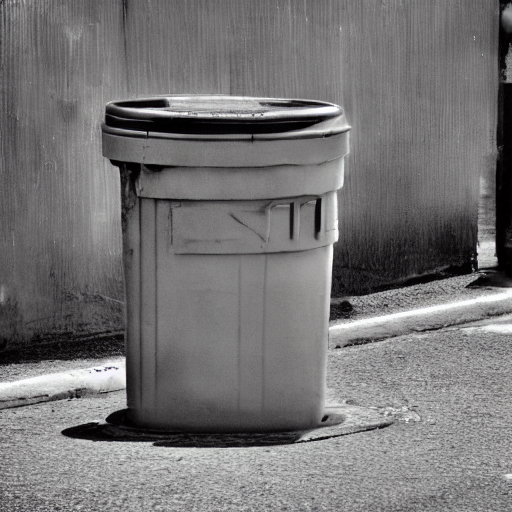} & \includegraphics[width=0.20\linewidth]{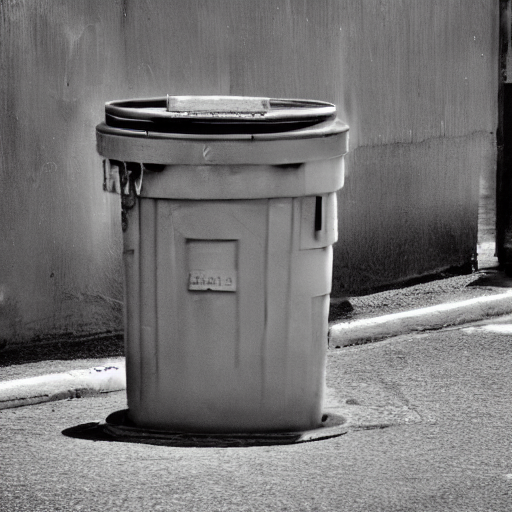} \\

\end{tabular}
\caption{\textbf{Image Synthesis Results -- Trashcan.} The first row is the standard Stable Diffusion result, with each subsequent row having token merging applied with $r\in\{90,80,70,60,50,40\}\%$.}
\label{fig:sdVizSupp2}
\end{figure}

\begin{figure}[!tp]
\centering
\begin{tabular}{cccc}

ToMe  & ATC$^\text{single}$  & ATC$^\text{average}$  & ATC$^\text{complete}$\\
\includegraphics[width=0.20\linewidth]{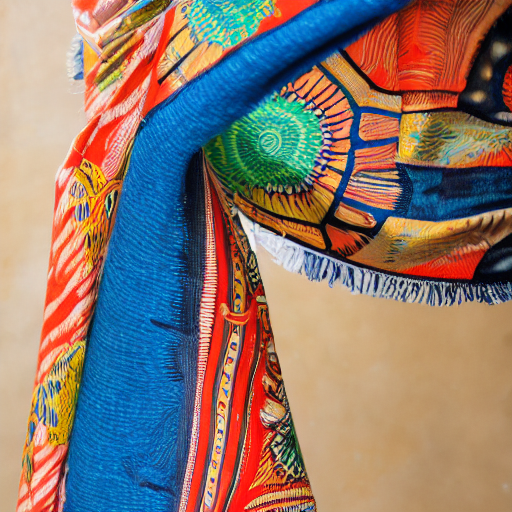} & \includegraphics[width=0.20\linewidth]{supp_figures_tables/sd_775_1/sd.png} & \includegraphics[width=0.20\linewidth]{supp_figures_tables/sd_775_1/sd.png} & \includegraphics[width=0.20\linewidth]{supp_figures_tables/sd_775_1/sd.png} \\
\includegraphics[width=0.20\linewidth]{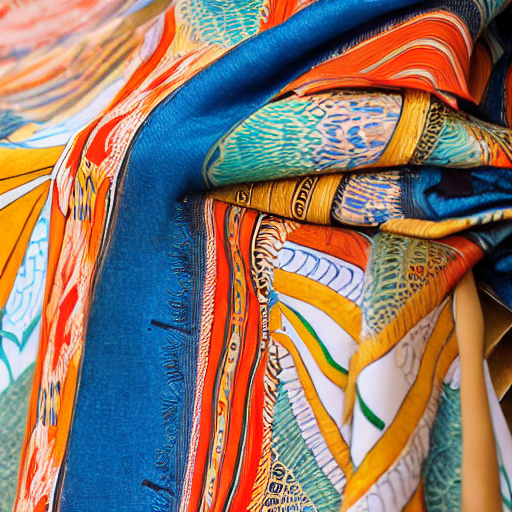} & \includegraphics[width=0.20\linewidth]{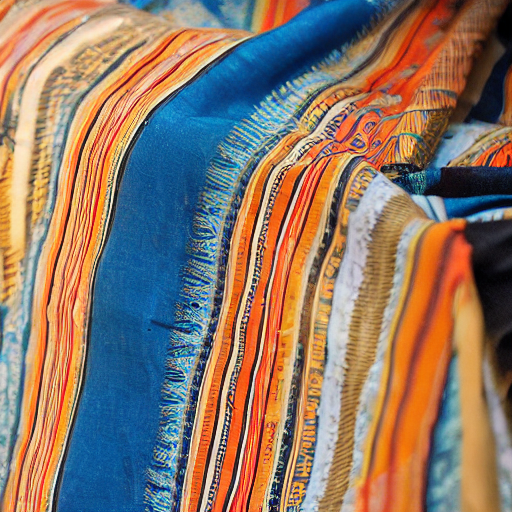} & \includegraphics[width=0.20\linewidth]{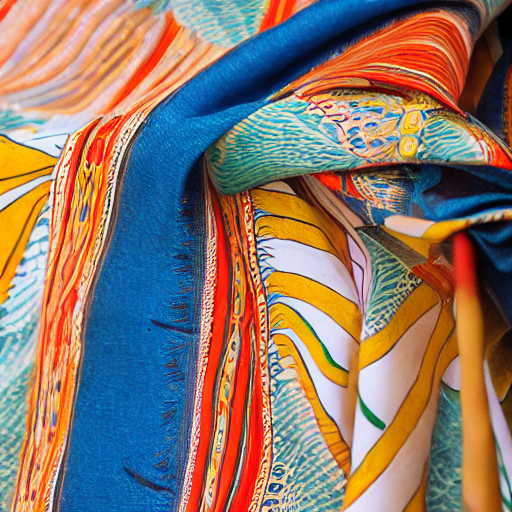} & \includegraphics[width=0.20\linewidth]{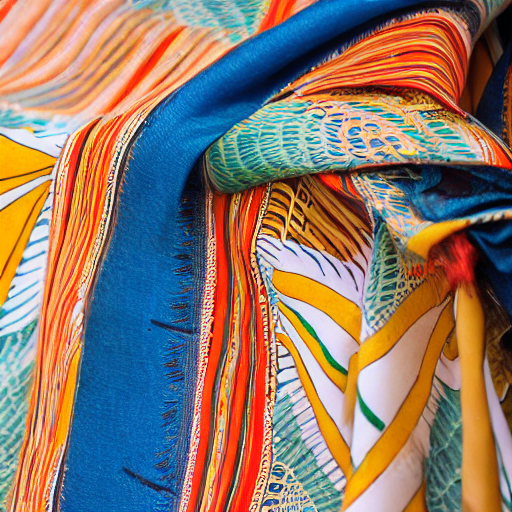} \\
\includegraphics[width=0.20\linewidth]{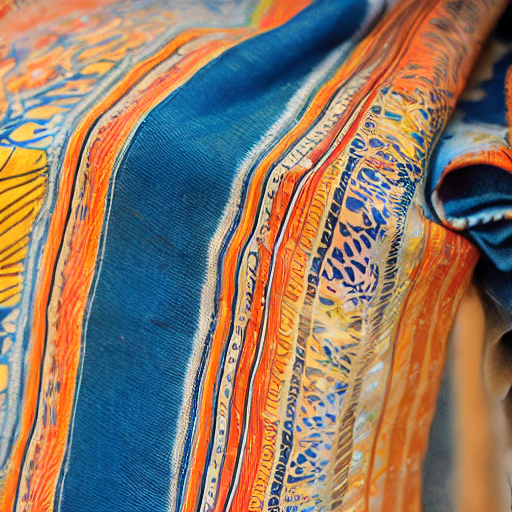} & \includegraphics[width=0.20\linewidth]{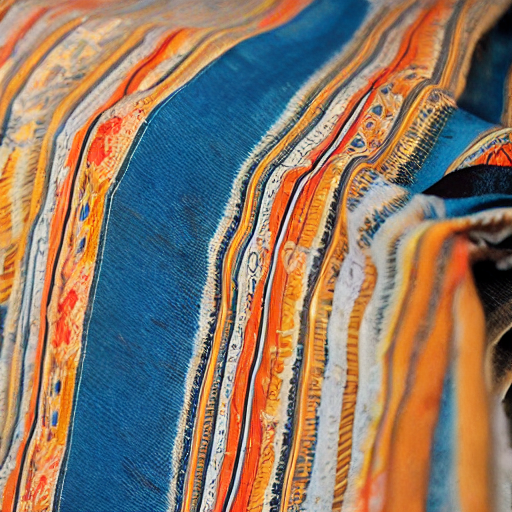} & \includegraphics[width=0.20\linewidth]{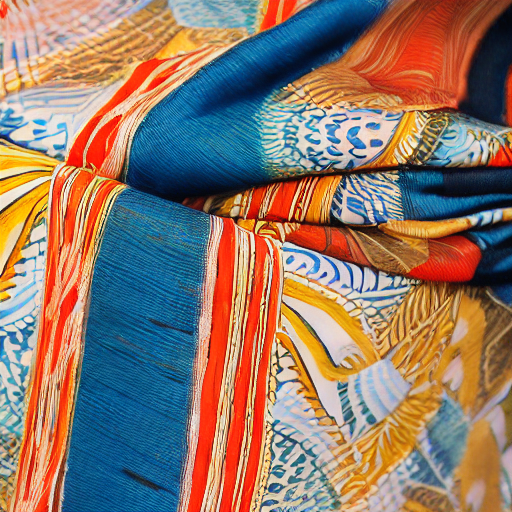} & \includegraphics[width=0.20\linewidth]{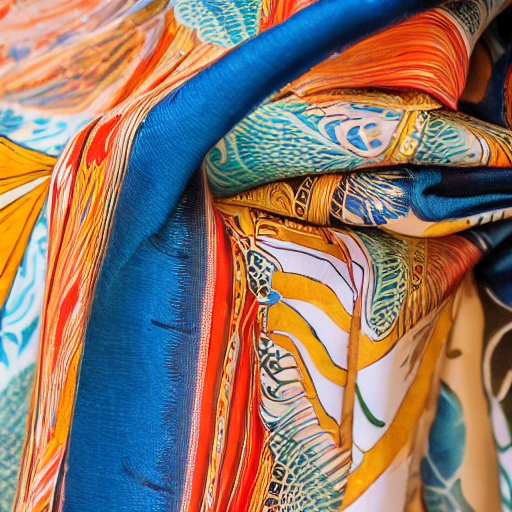} \\
\includegraphics[width=0.20\linewidth]{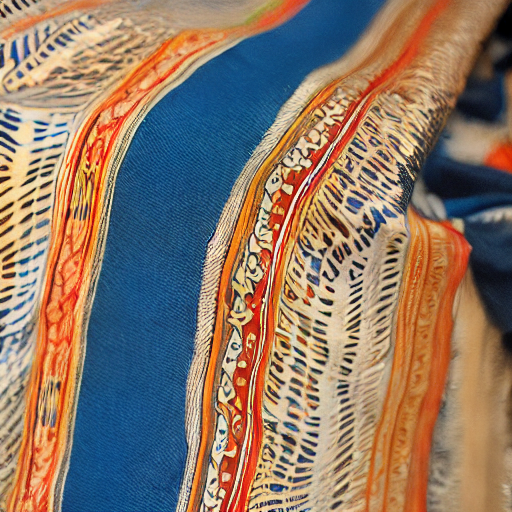} & \includegraphics[width=0.20\linewidth]{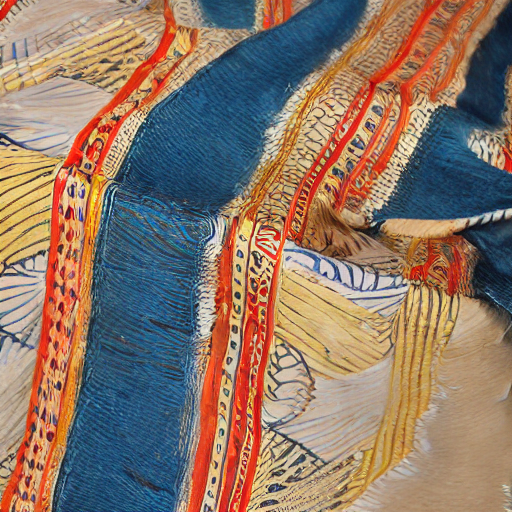} & \includegraphics[width=0.20\linewidth]{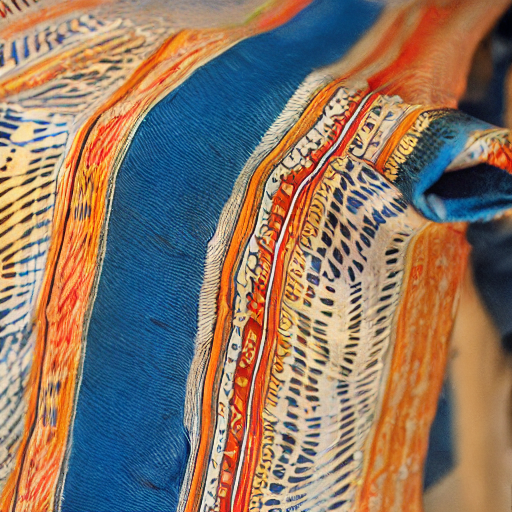} & \includegraphics[width=0.20\linewidth]{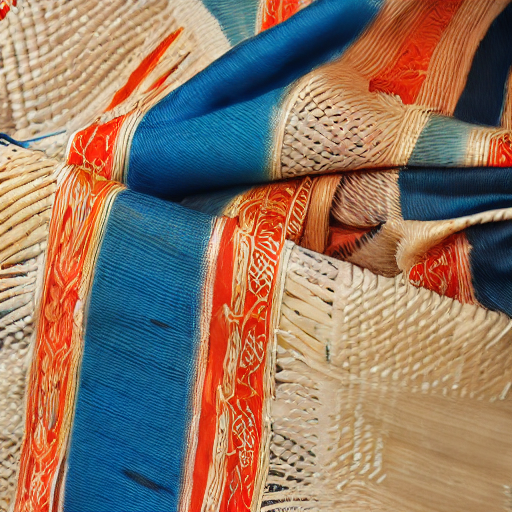} \\
\includegraphics[width=0.20\linewidth]{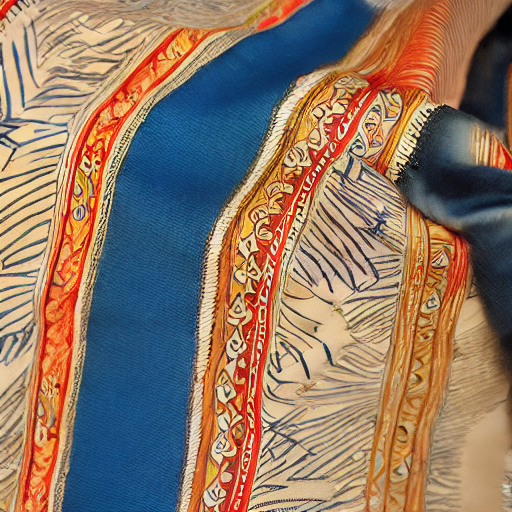} & \includegraphics[width=0.20\linewidth]{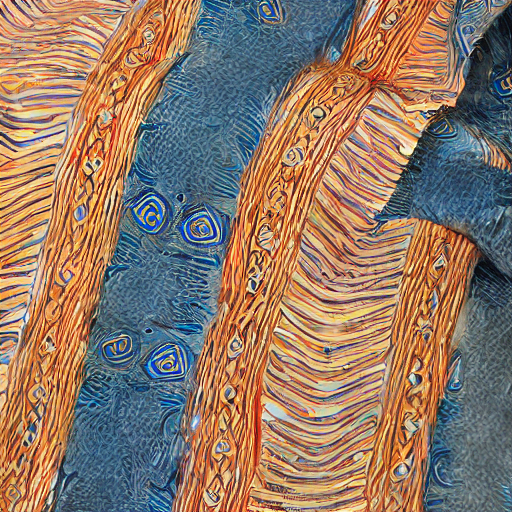} & \includegraphics[width=0.20\linewidth]{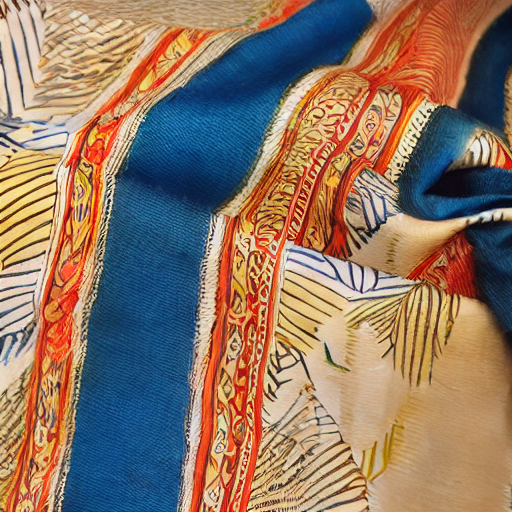} & \includegraphics[width=0.20\linewidth]{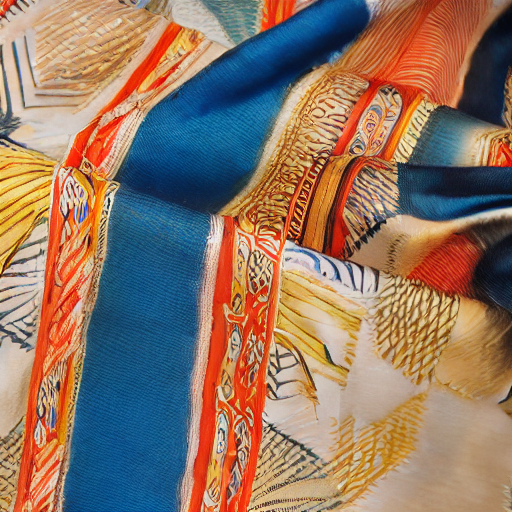}  \\
\includegraphics[width=0.20\linewidth]{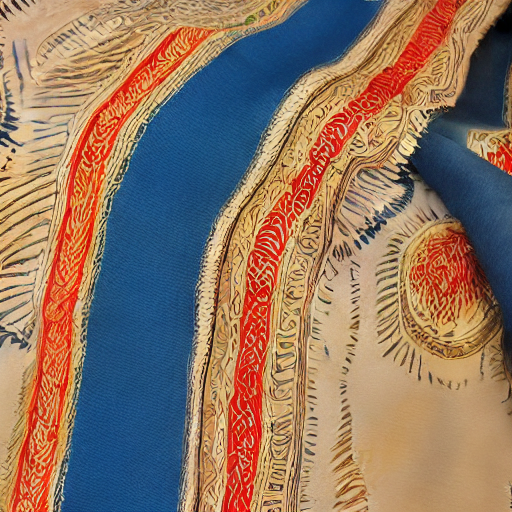} & \includegraphics[width=0.20\linewidth]{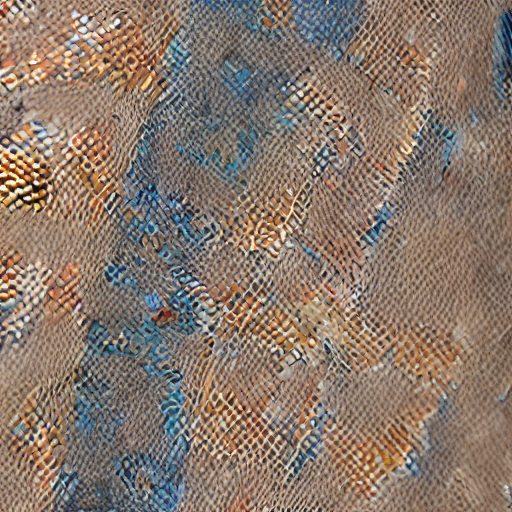} & \includegraphics[width=0.20\linewidth]{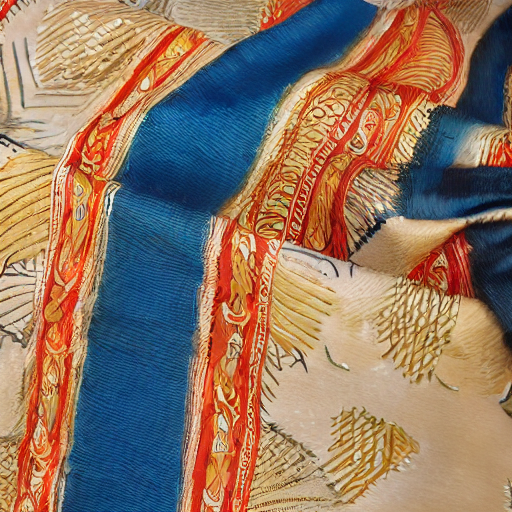} & \includegraphics[width=0.20\linewidth]{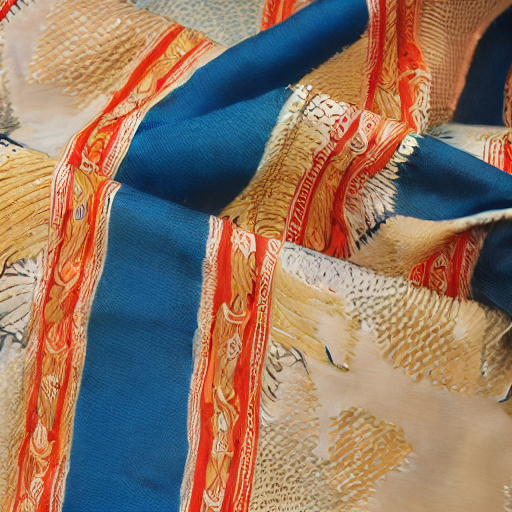} \\
 \includegraphics[width=0.20\linewidth]{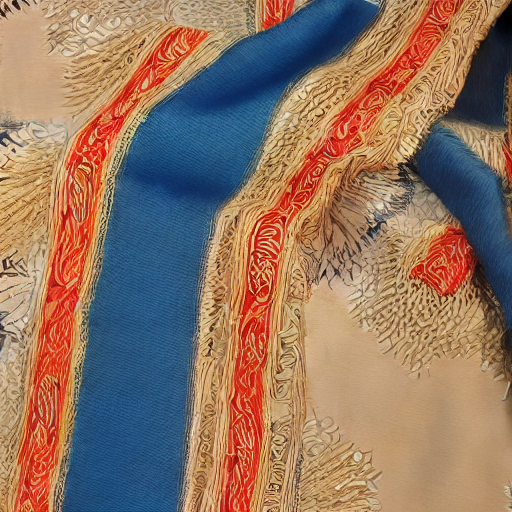} & \includegraphics[width=0.20\linewidth]{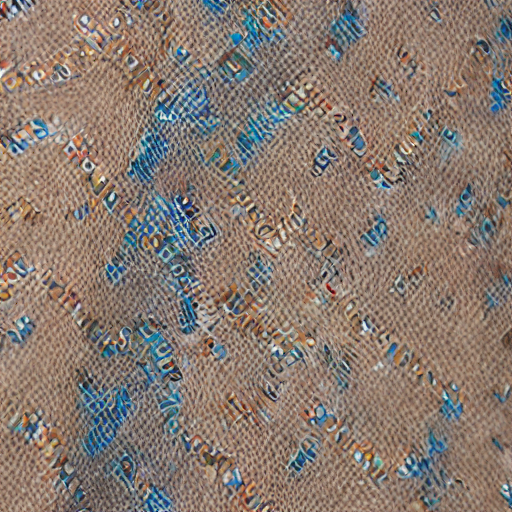} & \includegraphics[width=0.20\linewidth]{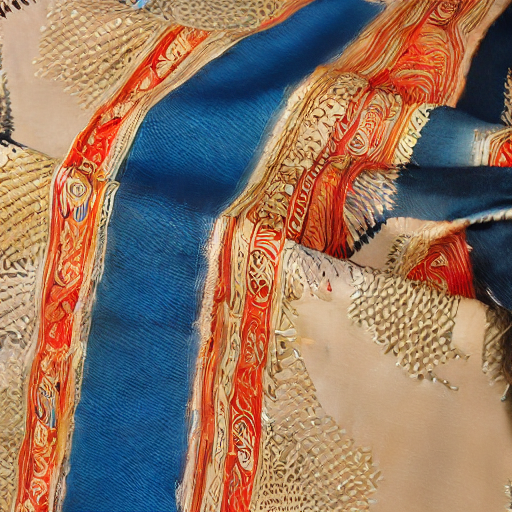} & \includegraphics[width=0.20\linewidth]{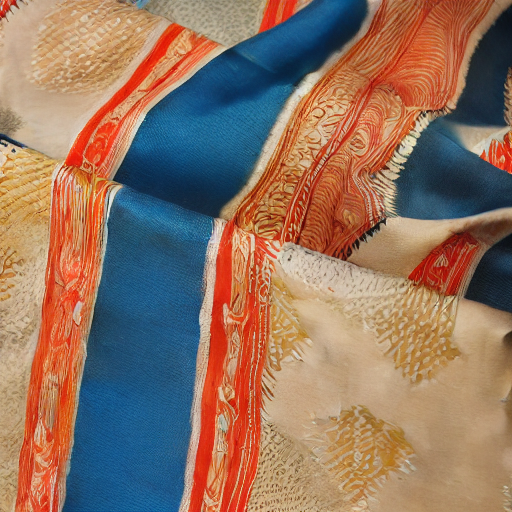} \\

\end{tabular}
\caption{\textbf{Image Synthesis Results -- Sarong.} The first row is the standard Stable Diffusion result, with each subsequent row having token merging applied with $r\in\{90,80,70,60,50,40\}\%$.}
\label{fig:sdVizSupp3}
\end{figure}

\begin{figure}[!tp]
\centering
\begin{tabular}{cccc}

ToMe  & ATC$^\text{single}$  & ATC$^\text{average}$  & ATC$^\text{complete}$\\
\includegraphics[width=0.20\linewidth]{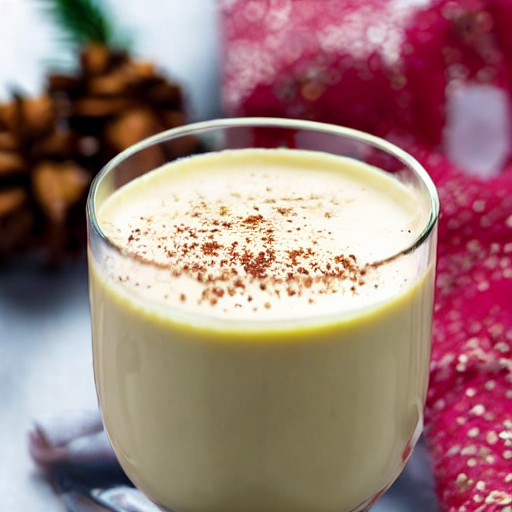} & \includegraphics[width=0.20\linewidth]{supp_figures_tables/sd_969_0/sd.png} & \includegraphics[width=0.20\linewidth]{supp_figures_tables/sd_969_0/sd.png} & \includegraphics[width=0.20\linewidth]{supp_figures_tables/sd_969_0/sd.png} \\
\includegraphics[width=0.20\linewidth]{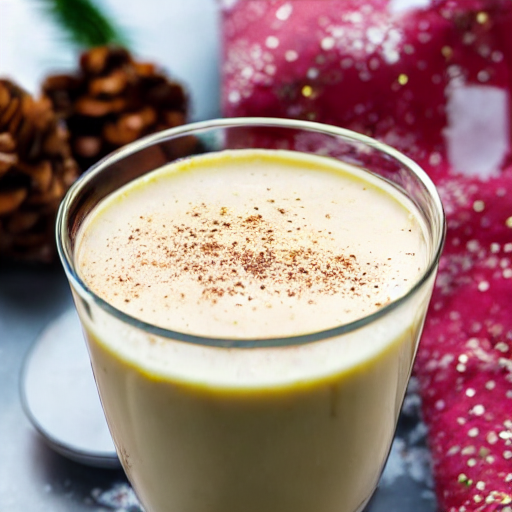} & \includegraphics[width=0.20\linewidth]{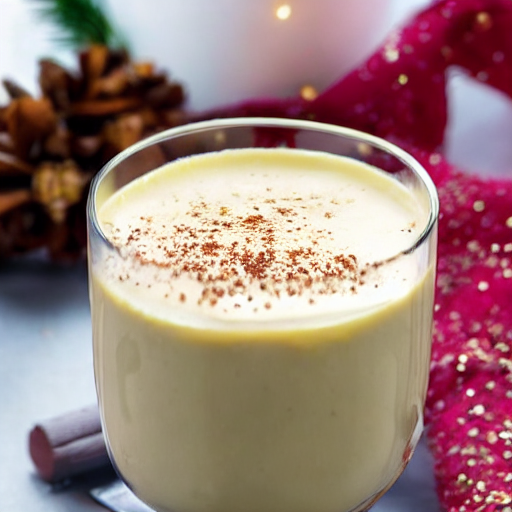} & \includegraphics[width=0.20\linewidth]{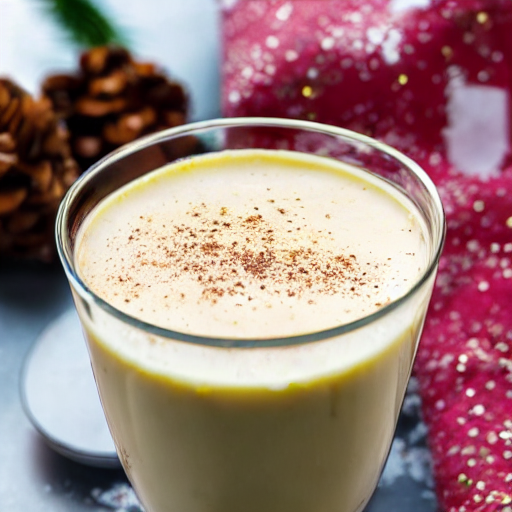} & \includegraphics[width=0.20\linewidth]{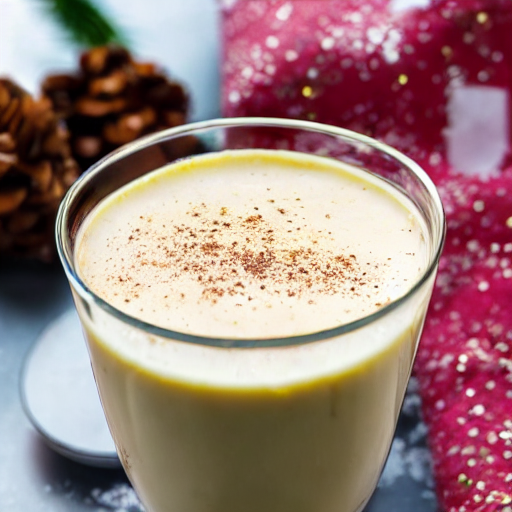} \\
\includegraphics[width=0.20\linewidth]{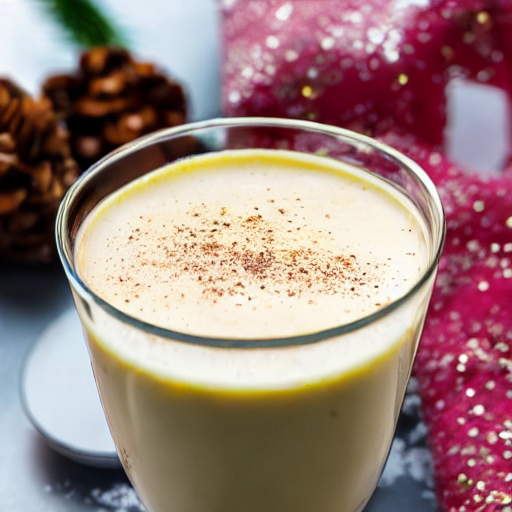} & \includegraphics[width=0.20\linewidth]{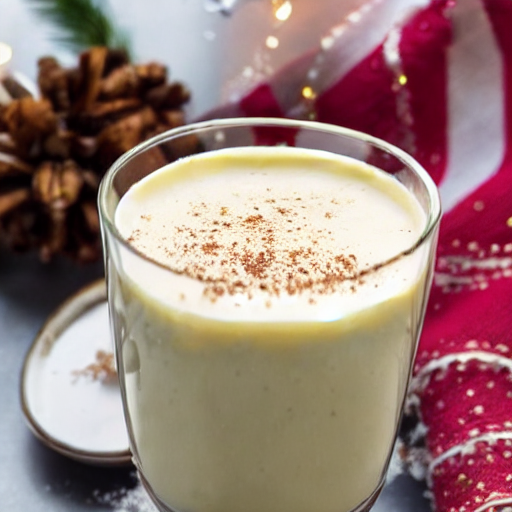} & \includegraphics[width=0.20\linewidth]{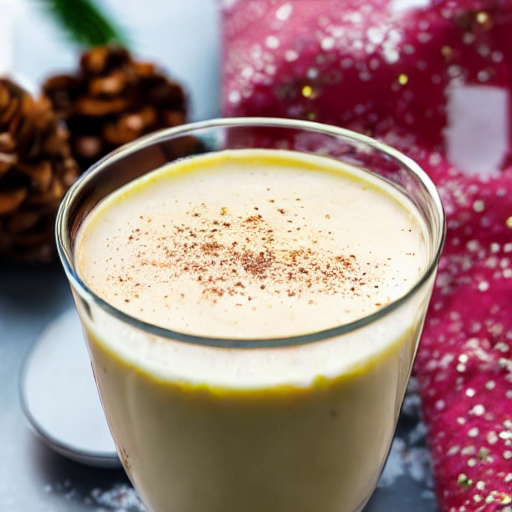} & \includegraphics[width=0.20\linewidth]{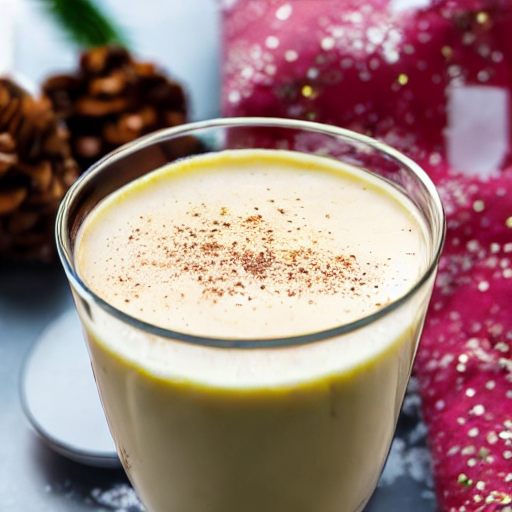} \\
\includegraphics[width=0.20\linewidth]{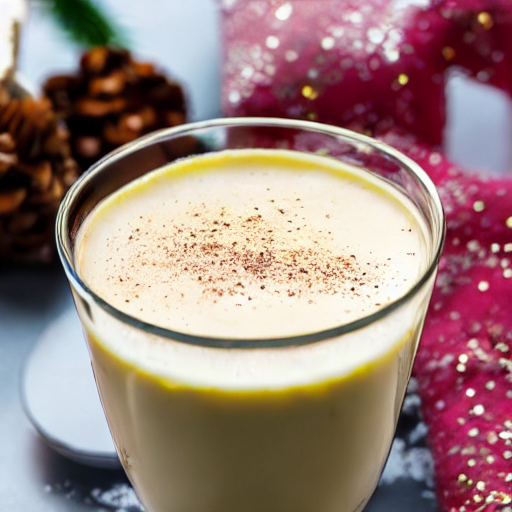} & \includegraphics[width=0.20\linewidth]{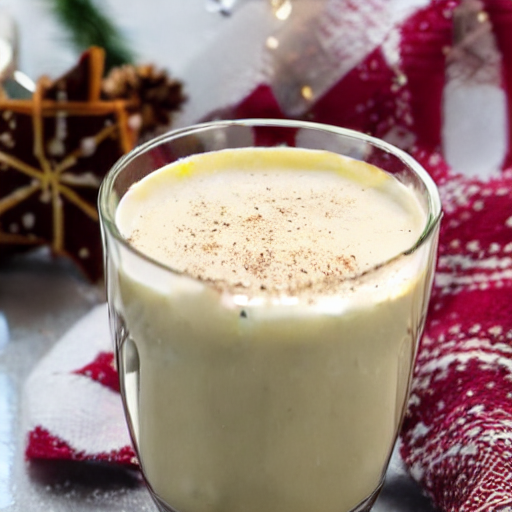} & \includegraphics[width=0.20\linewidth]{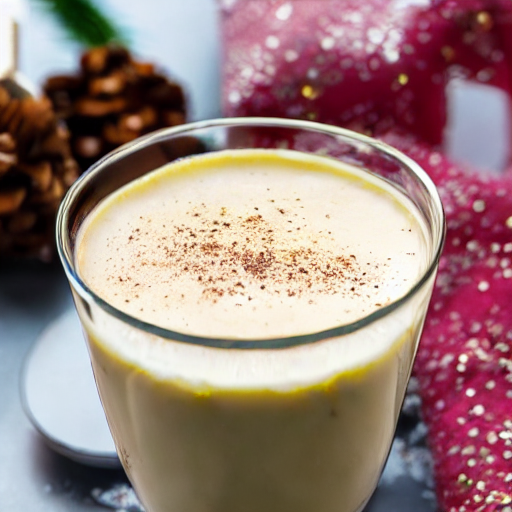} & \includegraphics[width=0.20\linewidth]{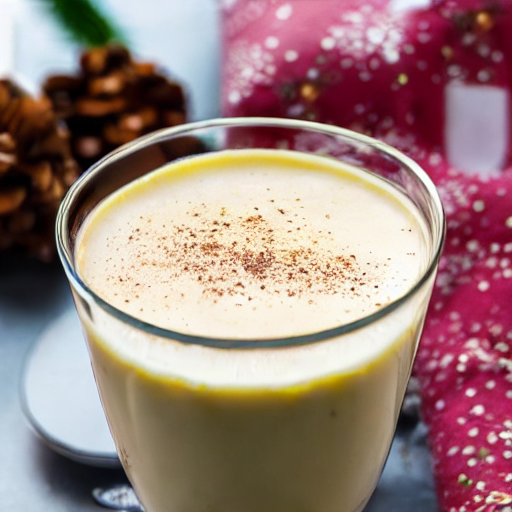} \\
\includegraphics[width=0.20\linewidth]{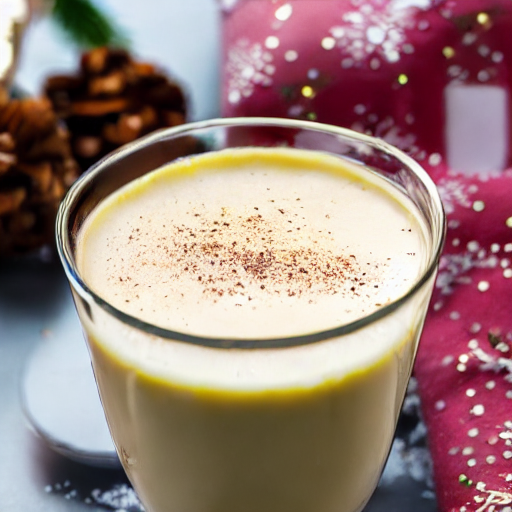} & \includegraphics[width=0.20\linewidth]{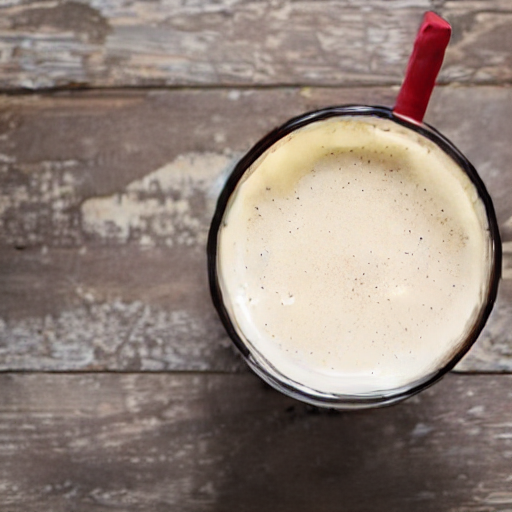} & \includegraphics[width=0.20\linewidth]{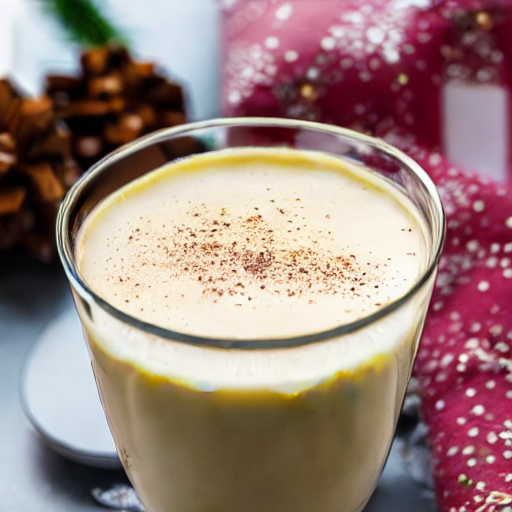} & \includegraphics[width=0.20\linewidth]{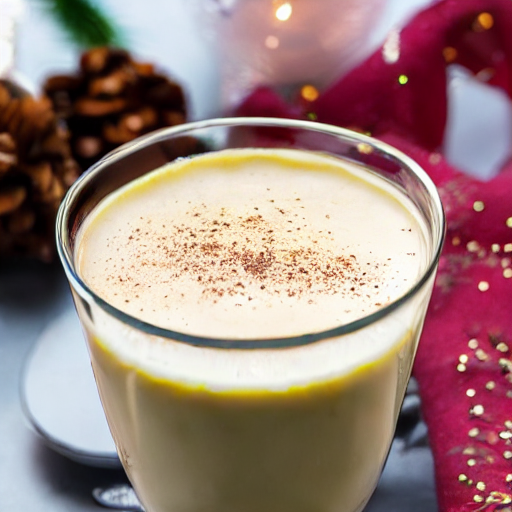}  \\
\includegraphics[width=0.20\linewidth]{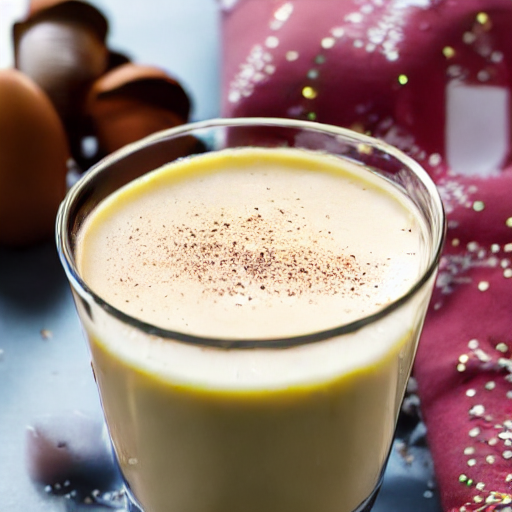} & \includegraphics[width=0.20\linewidth]{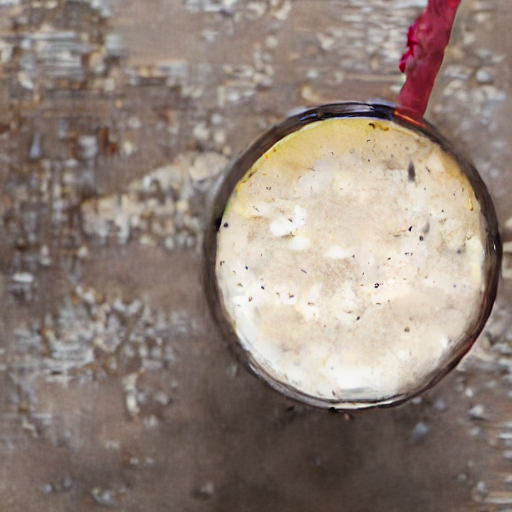} & \includegraphics[width=0.20\linewidth]{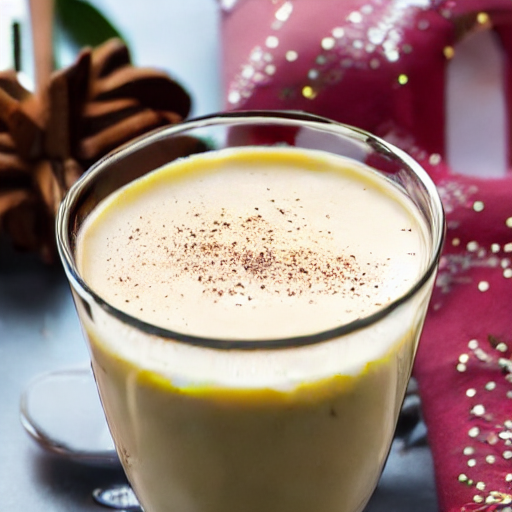} & \includegraphics[width=0.20\linewidth]{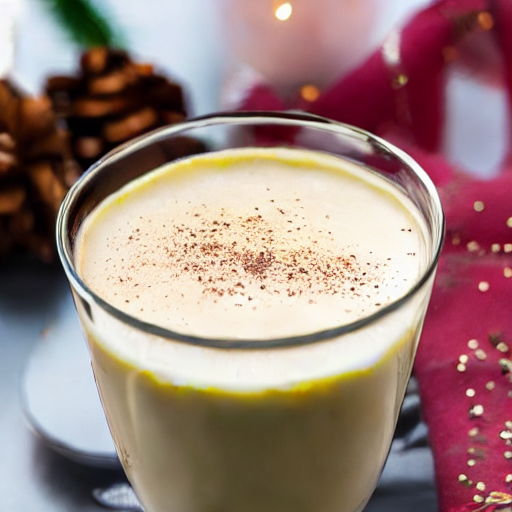} \\
 \includegraphics[width=0.20\linewidth]{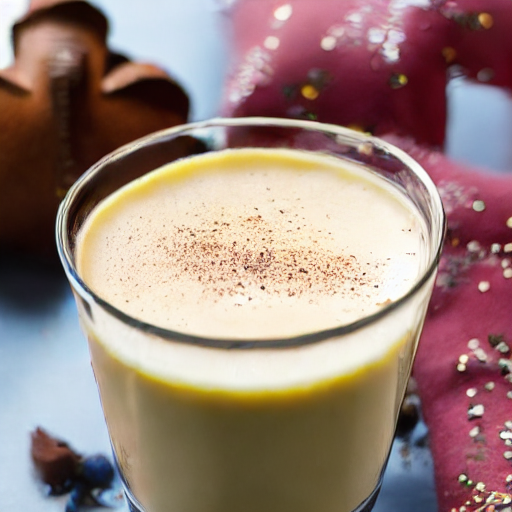} & \includegraphics[width=0.20\linewidth]{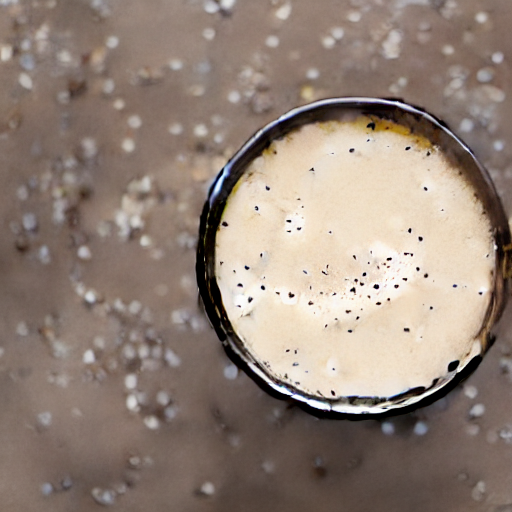} & \includegraphics[width=0.20\linewidth]{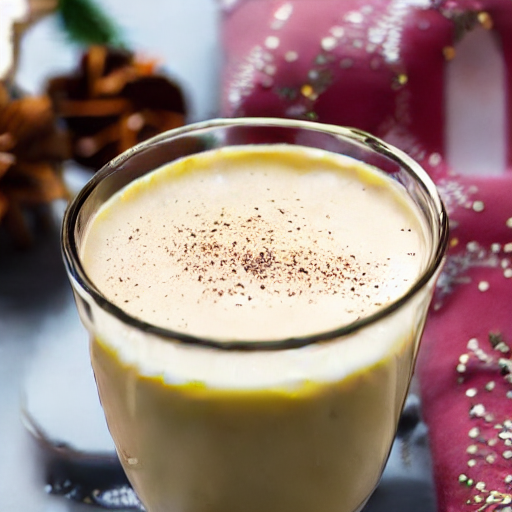} & \includegraphics[width=0.20\linewidth]{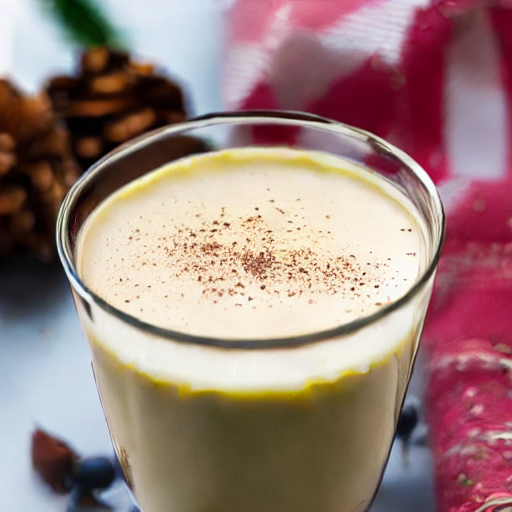} \\

\end{tabular}
\caption{\textbf{Image Synthesis Results -- Eggnog.} The first row is the standard Stable Diffusion result, with each subsequent row having token merging applied with $r\in\{90,80,70,60,50,40\}\%$.}
\label{fig:sdVizSupp4}
\end{figure}

We provide examples of the image synthesized using Stable Diffusion \cite{stableDiffusion} when applying the ToMe and ATC. We consider keep rates $r\in\{90,80,70,60,50,40\}\%$, and generate all images using fixed seeds and with the following prompt:
\begin{quote}
    \textit{A high quality photograph of a $\{\text{CLASS}\}$.}
\end{quote}

\noindent where {CLASS} is the corresponding ImageNet class name. The results are shown in Figures~\ref{fig:sdVizSupp1}-\ref{fig:sdVizSupp4}.

%
%
\bibliographystyle{splncs04}
\bibliography{main}

\end{document}